%% file: main.tex
\documentclass[final,12pt,3p]{elsarticle}

\usepackage[colorlinks=true]{hyperref}

\usepackage{amssymb}

\input{macros.tex}

\renewcommand{\mathbf}[1]{\bm{#1}}

\graphicspath{{./tikzpics/}{./figures/}}

\usepackage{lineno}

\journal{Computer Methods in Applied Mechanics and Engineering}

\begin{document}
	
\begin{frontmatter}
		

		
\title{Unifying and extending Diffusion Models through PDEs for solving Inverse Problems}

		
\author[label1]{Agnimitra Dasgupta}
\author[label2]{Alexsander Marciano da Cunha}
\author[label1]{Ali Fardisi}
\author[label1]{Mehrnegar Aminy}
\author[label1]{Brianna Binder}
\author[label1]{Bryan Shaddy}
\author[label1]{Assad A Oberai}

\affiliation[label1]{organization={{Department of Aerospace \& Mechanical Engineering, University of Southern California}},
city={Los Angeles},
postcode={90089}, 
state={California},
country={USA}}

\affiliation[label2]{organization={{Institute of Computing, Universidade Federal Fluminense}},
city={Rio de Janeiro},
postcode={RJ 24210-346}, 
state={Rio de Janeiro},
country={Brazil}}

\begin{abstract}

Diffusion models have emerged as powerful generative tools with applications in computer vision and scientific machine learning (SciML), where they have been used to solve large-scale probabilistic inverse problems. Traditionally, these models have been derived using principles of variational inference, denoising, statistical signal processing, and stochastic differential equations. In contrast to the conventional presentation, in this study we derive diffusion models using ideas from linear partial differential equations and demonstrate that this approach has several benefits that include a constructive derivation of the forward and reverse processes, a unified derivation of multiple formulations and sampling strategies, and the discovery of a new class of variance preserving models. We also apply the conditional version of these models to solve canonical conditional density estimation problems and challenging inverse problems. These problems help establish benchmarks for systematically quantifying the performance of different formulations and sampling strategies in this study and for future studies. Finally, we identify and implement a mechanism through which a single diffusion model can be applied to measurements obtained from multiple measurement operators. Taken together, the contents of this manuscript provide a new understanding of and several new directions in the application of diffusion models to solving physics-based inverse problems. 


\end{abstract}
		
		
		
		
	\end{frontmatter}

\section{Introduction}\label{sec:introduction}

\subsection{Diffusion models}

Diffusion models have emerged as one of the most popular generative tools. They have found applications in computer vision, where they are used in popular text-to-image and text-to-video software like Dall-E-3\footnote{https://openai.com/index/dall-e-3/} and Sora\footnote{https://openai.com/sora/}. They have also been used in applications of Scientific Machine Learning (SciML), where they have been used to quantify and propagate uncertainty in physics-based forward and inverse problems~\cite{bastek2024physics,jacobsen2025cocogen,dasgupta2025conditional,shu2024zero,zhuang2025spatially,feng2023score,sun2024provable,tashiro2021csdi,chung2022diffusion,chung2022improving,graikos2022diffusion,mardani2023variational}. 

These models come in two broad forms - unconditional and conditional. Unconditional diffusion models use independent and identically distributed (iid) samples from an underlying probability distribution to generate more samples from the same distribution~\cite{song2019generative,song2020score,ho2020denoising}. On the other hand, conditional diffusion models use iid samples drawn from a joint probability distribution to create a tool to generate samples from a conditional distribution~\cite{batzolis2021conditional}. 

Diffusion models work by first generating samples from a simple Gaussian probability distribution and treating each sample as the initial state of a stochastic differential equation (SDE) or an ordinary differential equation (ODE). The SDE/ODE is designed such that the samples obtained by integrating these equations to their final state are iid samples of the target distribution. For unconditional diffusion models, the target distribution is the original data distribution, whereas for conditional diffusion models, it is the conditional probability distribution. In the design of the SDE/ODE, the score function (defined as the gradient of the log of a probability density function) associated with time-dependent  probability density plays an important role. For this reason, the models are often referred to as score-based diffusion models~\cite{hyvarinen2005estimation,song2019generative}. 

A version of diffusion models was first derived from ideas that originated in variational inference applied to a denoising problem~\cite{ho2020denoising}. Another version was independently derived using ideas based on the Langevin Monte-Carlo method and score-matching~\cite{song2019generative,song2020improved}. Both versions were then shown to originate from a common framework that used forward and reverse SDEs to derive the underlying algorithms. These versions were labeled as the variance exploding and variance preserving formulations~\cite{song2020score}. In the variance-exploding formulation, the initial Gaussian distribution has zero mean and a very large variance, whereas in the variance-preserving formulation, it is the standard Normal distribution.

\paragraph{Contributions of this work} In this study we present an alternative derivation of diffusion models from a perspective that operates at the level of probability density functions (pdfs) rather than the corresponding samples and SDEs. It relies on elementary concepts from partial differential equations (PDEs), especially those related to the scalar drift-diffusion equation. This point of view has several advantages. First, it yields a simple and constructive derivation of the reverse process that transforms a high-dimensional Gaussian pdf to the data pdf. Second, similar to the derivation in \cite{song2020score}, it provides a unified exposition of variance exploding and variance preserving diffusion models. Third, it leads to a class of novel variance-preserving formulations that are identified for the first time in this study. Finally, it naturally leads to a family of sampling methods of which the probability flow ODE and the SDEs described in \cite{song2021solving} are special cases. 

\subsection{Diffusion models for solving inverse problems}

Recently, both unconditional and conditional diffusion models have been applied to solving probabilistic inverse problems in science and engineering~\cite{dasgupta2025conditional,jacobsen2025cocogen,bastek2024physics,song2021solving,song2022pseudoinverse,chung2022diffusion}. In the approach that uses the unconditional diffusion model, following the application of Bayes rule, the posterior density for the inferred vector is written as the product of the likelihood and prior densities~\cite{daras2024survey}. This allows the score of the posterior density to be expressed as the sum of the scores of the likelihood and prior densities. The former is obtained by approximating the forward problem, while the latter is learned by an unconditional diffusion model that uses samples from the prior distribution. In this approach, the diffusion model is used to approximate the prior density, and, therefore, once learned, the same diffusion model can be used to solve multiple inverse problems with different likelihood terms. 

In the approach that uses the conditional diffusion model to solve the inverse problem, samples from the prior distribution of the inferred vector are used in the forward model to generate corresponding samples of the measurement~\cite{batzolis2021conditional,dasgupta2025conditional}. Thereafter, paired samples of the inferred and measured vectors (viewed as iid samples from their joint distribution) are used to train a conditional diffusion model. Once trained, this model is used to generate samples of the inferred vector conditioned on a given measurement. In this approach,  the trained diffusion model can be used for multiple instances of the measurement. However, if the forward model or the measurement function is altered, the diffusion model has to be retrained. This is a disadvantage when compared with the approach that uses an unconditional diffusion model to learn the score of the prior density (described in the previous paragraph). However, the advantages of this approach are that it can work with complex models for measurement noise, and it does not require the explicit knowledge of the forward model and can engage with it as a black box. In contrast, the approach that utilizes the unconditional diffusion model requires the evaluation of the gradient of the likelihood term, which in turn necessitates a simple model for measurement noise and the ability to compute the derivative of the forward model with respect to the inferred vector. 

\paragraph{Contributions of this work} In this study, in addition to the alternative derivation of diffusion models, we present several novel developments related to the application of conditional diffusion models to solving inverse problems. First, we apply these models to a low-dimensional conditional density estimation problem and quantify the error in their approximation, thereby establishing a benchmark in the use of these models as a conditional estimation tool. Second, we apply them to solve a challenging inverse problem of moderate dimensions, where we determine the boundary flux of a transported species given sparse and noisy measurements of its concentration in the domain. Third, in the context of this problem, by conditioning the model on a vector that parameterizes the measurement operator, in addition to the measurement, we demonstrate how a single diffusion model can be used to solve inverse problems corresponding to multiple measurement operators. This enables the trained diffusion model to solve a larger class of inverse problems. Finally, for both examples considered in this study, we examine the effect of using different diffusion model formulations (variance exploding and preserving) and sampling strategies (stochastic and deterministic). 

\subsection{Organization of the paper}

The format of the remainder of this manuscript is as follows. In \Cref{sec:uncond-gen}, we describe a new PDE-based approach to deriving unconditional diffusion models. This includes deriving the forward and reverse processes, their particle counterparts, and their loss functions for variance exploding and variance preserving formulations. In \Cref{sec:inverse-problem} we introduce the probabilistic inverse problem and demonstrate how it may be recast as a conditional generative problem. In \Cref{sec:cond-gen}, we derive variance exploding and variance preserving conditional diffusion models. Thereafter, in \Cref{sec:results}, we apply these models to solve several canonical conditional density estimation problems and challenging inverse problems motivated by advection-diffusion and advection-diffusion-reaction equations. Here we compare the performance of different formulations of diffusion models and sampling strategies. We end with conclusions in \Cref{sec:conclusion}.

\section{Unconditional diffusion models}\label{sec:uncond-gen}
	
We begin by describing the unconditional generation problem. Given $N$ independent realizations sampled from an underlying, unknown distribution with density $\prob{\mathrm{data}}$, the objective of the unconditional generative problem is to sample new realizations of $\prob{\mathrm{data}}$. Therefore, we assume that we have available independent realizations of the random variable $\X \in \Omega_{\mathcal{X}} \subseteq \mathbb{R}^{\Nx}$ with density $\prob{\mathrm{data}}$. 
		
Like other generative models, diffusion models also transform realizations from a tractable distribution (\eg multivariate Gaussian distribution) into new realizations from the unknown data distribution $\prob{\mathrm{data}}$. To do this, $\prob{\mathrm{data}}$ is first transformed into a tractable distribution. Then, samples are drawn from this distribution, and the corresponding inverse transform is applied to these samples to obtain samples from $\prob{\mathrm{data}}$. For diffusion models, the forward transform is not a function but a stochastic process. Similarly, the inverse transform is obtained from the reverse process. In the paragraphs below, we will derive the forward and reverse stochastic processes. 

\subsection{Variance exploding forward process}
	
We start with $\prob{t}(\x)$ to denote a time-dependent probability distribution of a stochastic process with the initial condition $\prob{0}(\x) = \prob{\mathrm{data}}(\x)$. Herein, we will drop the argument $\x$ unless necessary to simplify our notation. The stochastic process models the evolution of the random vector $\X_t$, such that at any time instant $t$ we have $\X_t \sim  \prob{t}$ and $\X_0 \sim \prob{0}=\prob{\mathrm{data}}$. Now, let $\prob{t}$ satisfy the diffusion equation
	\begin{equation}\label{eq:diffusion}
		\frac{\partial \prob{t}(\x)}{\partial t} = \frac{\gamma(t)}{2} \Delta \prob{t}(\x).
	\end{equation}
    The solution of this equation is written in terms of the Green's function, $\prob{t}{ (\x \vert \x^\prime) }$ which satisfies the same PDE with the initial condition $\prob{0}{ (\x \vert \x^\prime) } = \delta (\x-\x^\prime)$. The solution is given by 
	\begin{equation}\label{eq:forward-diffusion}
		\prob{t}(\x) = \int_{\Omega_{\mathcal{X}}} \prob{t}{ (\x \vert \x^\prime) } \prob{\mathrm{data}}{(\x^\prime)} \mathrm{d}\x^\prime,
	\end{equation}
and the Green's function is 
	\begin{equation}\label{eq:diff-kernel}
		\prob{t}{ (\x \vert \x^\prime) } = (2\pi \sigma^2(t))^{-\Nx/2} \exp \left( - \frac{ \lvert \x - \x^\prime  \rvert_2^2 }{2\sigma^2(t)} \right),
	\end{equation}
	where
	\begin{equation}\label{eq:sigma-t}
		\sigma^2(t) = \int_{0}^{t} \gamma(t^\prime) \mathrm{d}t^\prime .
	\end{equation}
	\Cref{eq:diff-kernel} is the zero-mean multivariate normal distribution with an isotropic covariance matrix equal to $\sigma^2(t)\mathbb{I}$, which we will denote as $\mathcal{N}(\bm{0}, \sigma^2(t)\mathbb{I})$.  For some large time $T$, when $\sigma(T)$ assumes a large value, $\prob{T}(\x)$ can be approximated as $\mathcal{N}(\bm{0}, \sigma^2(T)\mathbb{I})$, from which we can easily sample. This completes the derivation of the forward process that diffuses $\prob{0}$ into $\prob{T} = \mathcal{N}(\bm{0}, \sigma^2(T)\mathbb{I})$. 
    
The process described above will lead to the so-called variance exploding formulation of the diffusion model. An undesirable characteristic of this formulation is that the variance ``blows up'' for large times. This issue is addressed in the variance-preserving formulation, which is derived next. 

\subsection{Variance preserving forward process}
    
We begin by noting that the probability density for the variance exploding formulation can be transformed to a density for which the variance remains bounded through a transformation of coordinates, 
\begin{equation}
    \q = \frac{\x}{\xi(t)},
\end{equation}
and working with a transformed Green's function
\begin{equation} \label{eq:defy}
    \probh{t}(\q| \x') = \xi(t) \prob{t}(\x| \x').
\end{equation}
We impose two conditions on $\xi(t)$. These are $\xi(0) = 1$, which ensures $\q = \x$ at $t=0$. The second is
\begin{equation} \label{eq:cond1}
    \lim_{t \to \infty} \frac{\xi(t)}{\sigma(t)} = 1, 
\end{equation}
which ensures that the variance remains bounded at large times. 

Substituting \Cref{eq:defy} in \Cref{eq:diff-kernel}, and recognizing that at $ t = 0$, $\q = \x$, we arrive at an expression for the transformed kernel,
\begin{eqnarray} \label{eq:varpres-green}
\probh{t}(\q| \x') &\propto&    \exp \left( - \frac{ \lvert \xi(t) \q - \x^\prime  \rvert_2^2 }{2\sigma^2(t)} \right)   \nonumber \\
    &=& \exp \left( - \frac{ \lvert \q - m(t) \x^\prime  \rvert_2^2 }{2 \hat{\sigma}^2(t)} \right) ,
\end{eqnarray}
where $m(t) = \xi^{-1}(t)$ and $\hat{\sigma}(t) = m(t) \sigma(t)$. 

From \Cref{eq:cond1}, we conclude that $\lim_{t \to \infty} m(t) = 0$, and $\lim_{t \to \infty} m(t) \sigma(t) = 1$. Thus, independent of $\x^\prime$, $\probh{t}(\q| \x')$ tends to the standard normal Gaussian density, which in turn implies that the density $\probh{t}(\q)$ given by 
\begin{equation}\label{eq:sol_var_pres}
    \probh{t}(\q) = \int_{\Omega_{\mathcal{X}}} \probh{t}{ (\q \vert \x^\prime) } \prob{\mathrm{data}}{(\x^\prime)} \mathrm{d}\x^\prime,
\end{equation}
also tends to the standard normal distribution.

Next, we determine the PDE satisfied by $\probh{t}(\q| \x')$. To do this, we recognize that 
\begin{equation}
    \frac{\partial \prob{t}(\x| \x') }{ \partial t} = - \frac{\dot{\xi}(t)}{\xi^2(t)} \nabla_{\q} \cdot(\q \probh{t}(\q| \x')) + \frac{1}{\xi(t)} \frac{\partial \probh{t}(\q| \x') }{\partial t},
\end{equation}
and
\begin{equation}
    \Delta \prob{t}(\x| \x')  =  \frac{1}{\xi^3(t)} \Delta_{\q}  \probh{t}(\q| \x').
\end{equation}
Using these in \Cref{eq:diffusion}, we arrive at 
\begin{equation}
\label{eq:pde-var-pres01}
    \frac{\partial \probh{t}(\q| \x') }{\partial t} -\frac{\dot{\xi}(t)}{ \xi(t)}  \nabla_{\q} \cdot \big( \q \probh{t}(\q| \x')\big) -\frac{\gamma(t)}{2 \xi^2(t)} \Delta_y  \probh{t}(\q| \x')  = 0. 
\end{equation}
The density $\probh{t}(\q)$ obtained by convolving the initial distribution with this kernel also satisfies this PDE.

A specific choice of $\xi$ that leads to a family of variance preserving versions of diffusion models is given by 
\begin{equation} \label{eq:xi_def}
    \xi^\mu(t) = 1 + \sigma^\mu(t),
\end{equation}
where the parameter $\mu > 0$ defines the members of this family. As shown in  \ref{sec:var_pres}, this family of models is conveniently described in terms of a user-defined function $\beta(t) \equiv 2\dot{\xi}(t) /\xi(t)$. Written in terms of this function, the transformed kernel is given by \Cref{eq:varpres-green}, where 
\begin{equation} \label{eq:defm}
    m(t) =  \exp \left( - \frac{1}{2} \int_0^t \beta(s) \,\mathrm{d}s \right),
\end{equation}
and 
\begin{equation}
    \hat{\sigma}(t) = \left( 1 - m(t)^{\mu} \right)^{\frac{1}{\mu}}.
\end{equation}
Further, the PDE satisfied by this kernel is given by 
\begin{equation}
    \frac{\partial \probh{t}(\q| \x') }{\partial t} -\frac{\beta(t)}{2}  \nabla_{\q} \cdot \left(\q \probh{t}(\q| \x')\right) -\frac{\beta(t)}{2 } \left(1 - \exp\left(- \frac{\mu}{2} \int_0^t \beta(s) \, \mathrm{d}s\right) \right)^{1 - \frac{2}{\mu}} \Delta_{\q}  \probh{t}(\q| \x')  = 0. 
\end{equation}

\begin{figure}
    \centering
    \includegraphics[width=0.5\linewidth]{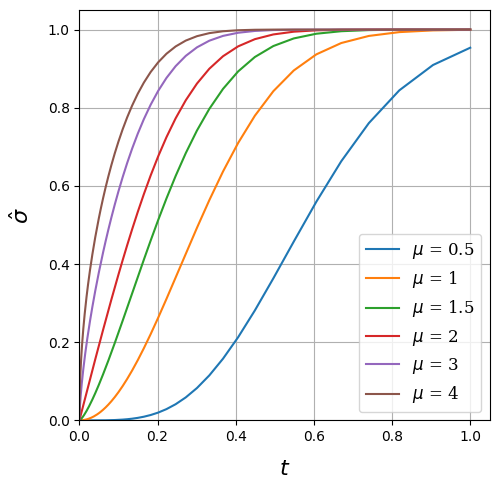}
    \caption{Plot of $\hat{\sigma}(t)$ for $\beta(t) = Ct$, where $C= 30$, and different values of $\mu$ for the family of variance preserving methods}
    \label{fig:stdHat_t_mu}
\end{figure}

In \cite{song2020score}, the authors consider the specific case $\mu = 2$ and label it as the variance preserving formulation. The extension described here generalizes this formulation for any value of $\mu >0$. One way to interpret this generalization is that gives the practitioner independent control on $m(t)$ and $\hat{\sigma}(t)$. To appreciate this, consider the special case of $T = 1$ and $\beta(t) = C t$, where $C \gg 1$, which ensures that $m(t) \to 0$ as $t \to 1$. In \Cref{fig:stdHat_t_mu}, for $C=30$, we have plotted $\hat{\sigma}(t)$ as a function of $t$ for different values of $\mu$. In this figure, it is clearly seen that varying $\mu$ leads to different behavior for $\hat{\sigma}$ as a function of $t$. 

\subsection{Unified description}

It is instructive and useful to write a unified description of the variance exploding and variance preserving versions of the diffusion model. In both cases, the forward process satisfies the PDE
\begin{equation}
\label{eq:pde-forward-combined}
    \frac{\partial \prob{t}(\x) }{\partial t} - \frac{b(t)}{2}  \nabla \cdot \left(\x \prob{t}(\x)\right) -\frac{g(t)}{2} \Delta  \prob{t}(\x)  = 0,
\end{equation}
with the initial condition $\prob{0}(\x) = \prob{\mathrm{data}}(\x)$. For the variance exploding version, $b(t) = 0$, and the user specifies $g(t) = \gamma(t)$. Whereas, for the variance preserving version the user specifies $b(t) = \beta(t)$ and 
\begin{equation}
    g(t) = \beta(t) \left(1 - \exp \left(- \frac{\mu}{2} \int_0^t \beta(s) \,\mathrm{d}s\right) \right)^{1 - \frac{2}{\mu}}.
\end{equation}

The solution to the PDE above is given by 
\begin{equation}
    \prob{t}(\x) = \int_{\Omega_{\mathcal{X}}} \prob{t}{ (\x \vert \x^\prime) } \prob{\mathrm{data}}{(\x^\prime)} \mathrm{d}\x^\prime,
\end{equation}
where $\prob{t}{ (\x \vert \x^\prime) }$ is a Gaussian kernel with mean $m(t) \x^\prime$ and variance $\sigma^2(t) \mathbb{I}$. For the variance exploding version $m(t) = 1$ and $\sigma^2(t) = \int_0^t \gamma(s) \mathrm{d}s$; whereas for the variance preserving version $m(t) = \exp\left(-\frac{1}{2} \int_0^t \beta(s) \mathrm{d}s \right)$, and $\sigma(t) = (1 - m^\mu(t))^{\frac{1}{\mu}}$. \Cref{tab:diff-formulations} summarizes the various properties --- $\beta(t), g(t), m(t)$ and $\sigma(t)$ --- for the two well known formulations for diffusion models along with the necessary user inputs.
\begin{table}[t]
    \centering
    
    \caption{Different formulations for diffusion models, their time-dependent properties and necessary user inputs}
    \label{tab:diff-formulations}
    \begin{tabular}{ccc}
    \toprule
    Property/Input & Variance exploding & Variance preserving \\
    \midrule
      $b(t)$ & 0 & $\beta(t)$ \\
      $g(t)$ & $\gamma(t)$ & $\beta(t) \left(1 - \exp \left(- \frac{\mu}{2} \int_0^t \beta(s) \,\mathrm{d}s\right) \right)^{1 - \frac{2}{\mu}}$ \\
      $m(t)$ & 1 &  $\exp\left(-\frac{1}{2} \int_0^t \beta(s) \mathrm{d}s \right)$\\
      $\sigma^2(t)$ & $\int_0^t \gamma(s) \mathrm{d}s$ & $(1 - m^\mu(t))^{\frac{2}{\mu}}$ \\
      User inputs & $\gamma(t)$ & $\beta(t)$ \\
    \bottomrule
    \end{tabular}
\end{table}

The score function associated with $\prob{t}{ (\x \vert \x^\prime) }$	is \begin{equation}\label{eq:diff-kernel-score}
		\nabla  \log  \prob{t}{ (\x \vert \x^\prime) } = \frac{ m(t) \x^\prime  - \x }{\sigma^2(t)}.
	\end{equation}
We will use \Cref{eq:diff-kernel-score} later.

Finally, since $\prob{t}{ (\x \vert \x^\prime) }$, which is the conditional density of $\x_t$ for a given $\x^\prime$, is Gaussian with mean $m(t) \x^\prime$ and variance $\sigma^2(t)\mathbb{I}$, given $\x^\prime \sim \prob{\mathrm{data}}{(\x^\prime)}$ we may obtain $\x_t \sim \prob{t}(\x)$ through 
\begin{equation}
\label{eq:forward_sample}
    \x_t = m(t) \x^\prime + \sigma(t) \bm{z}, 
\end{equation}
where $\bm{z} \sim \mathcal{N}(\bm{0}, \mathbb{I})$.

\subsection{The reverse process}    
	
We start by introducing the variable transformation $\tau = T - t$  such that marching back in $t$ is equivalent to moving forward in $\tau$. Also, let $\probt{\tau}(\x) = \prob{t}(\x)$ so the densities match at every point in time. Now
\begin{eqnarray}\label{eq:pde-temp1}
\frac{\partial \probt{\tau}(\x)}{\partial \tau}  &=&  - \frac{\partial \prob{t}(\x)}{\partial t} \nonumber \\
    &=& -\frac{b(t)}{2}  \nabla \cdot(\x \prob{t}(\x)) -\frac{g(t)}{2} \Delta  \prob{t}(\x) \nonumber \\
    &=& -\frac{b(t)}{2}  \nabla \cdot(\x \prob{t}(\x)) -\frac{(1+\alpha)g(t)}{2} \Delta  \prob{t}(\x) + \frac{\alpha g(t)}{2} \Delta  \prob{t}(\x) \nonumber \\
    &=& -\frac{b(t)}{2}  \nabla \cdot(\x \probt{\tau}(\x)) -\frac{(1+\alpha)g(t)}{2} \Delta  \prob{t}(\x) + \frac{\alpha g(t)}{2} \Delta  \probt{\tau}(\x)
\end{eqnarray}
where the second equality is obtained by using \Cref{eq:pde-forward-combined}, the third equality is obtained by adding and subtracting $(\alpha g(t) / 2) \Delta  \prob{t}(\x)$ (where $\alpha \ge 0$), and the fourth equality is obtained by recognizing $\prob{t}(\x) = \probt{\tau}(\x)$. We can rewrite 
	\begin{equation}\label{eq:score-transform}
		\nabla \prob{t}(\x) = \frac{\nabla \prob{t}(\x)}{\prob{t}(\x)} \prob{t}(\x)   =  \underbrace{\nabla \log \prob{t}(\x)}_{\text{score function}} \;\; \probt{\tau}(\x) 
	\end{equation}
	where the score function $\bm{s}_t(\x) = \nabla \log \prob{t}(\x)$ makes an appearance. Substituting \Cref{eq:score-transform} into \Cref{eq:pde-temp1} results in the following drift-diffusion equation
    \begin{equation}\label{eq:reverse-drift-diffusion}
		\frac{\partial \probt{\tau}(\x)}{\partial \tau} = - \nabla \cdot \left(  \left( \frac{b(t)}{2} \x + \frac{(1+ \alpha)g(t)}{2}  \bm{s}_t(\x) \right) \probt{\tau}(\x) \right) + \frac{\alpha g(t)}{2} \Delta  \probt{\tau}(\x)
	\end{equation}
In this equation $\bm{v}_t (\x) \equiv \left( b(t)/2 \right) \x + \left( (1+ \alpha)g(t) / 2 \right)  \bm{s}_t(\x) $ is the velocity field and $\alpha g(t) / 2$ is the diffusion coefficient. 

By construction, $\probt{\tau}(\x) = \prob{t}(\x)$, where $t = T-\tau$, is a solution to this equation. Further, from the uniqueness of the solutions to the drift-diffusion equation, we conclude that this is the only solution. Therefore, if we set $\probt{0} = \mathcal{N}(\bm{0}, \sigma^2(T)\mathbb{I})$ as the initial condition, and then solve \Cref{eq:reverse-drift-diffusion}, then we are guaranteed $\probt{T} = \prob{\mathrm{data}}$. 

\subsection{Particle counterpart of the reverse process}

The drift-diffusion equation above may be interpreted as the Fokker-Planck equation for the evolution of the probability density of a stochastic process governed by an It\^{o} SDE. This SDE is given by 
\begin{equation}\label{eq:reverse_SDE}
		\mathrm{d}\x_\tau =  \left( \frac{b(t)}{2} \x + \frac{(1+ \alpha) g(t)}{2}  \bm{s}_t(\x) \right) \mathrm{d}\tau + \sqrt{g(t) \alpha} \mathrm{d}\w_\tau,
\end{equation}
where $\w_\tau$ denotes the $\Nx$-dimensional Wiener process. Therefore, if $\x_0 \sim \mathcal{N}(\bm{0}, \sigma^2(T)\mathbb{I})$ and evolved according to this SDE, then at $\tau = T$, $\x_T \sim \prob{\mathrm{data}}$, the desired data density. For the variance exploding formulation, $\sigma(T)$ is a large positive number, whereas for the variance preserving formulation $\sigma(T) = 1$. 

This SDE may be integrated using a numerical method, such as the Euler-Maruyama method, which provides the update for $\x_{\tau + \delta \tau} $, given $\x_\tau$, 
\begin{equation}\label{eq:euler_maruyama}
		\x_{\tau + \Delta \tau} = \x_{\tau } + \left( \frac{b(t)}{2} \x + \frac{(1+ \alpha) g(t)}{2}  \bm{s}_t(\x) \right) \Delta \tau + \sqrt{ \alpha g(t) \Delta \tau} \z,
\end{equation}
where $\z$ are sampled independently from the $\Nx$-dimensional standard normal distribution. 

We note that in the development above, $\alpha \ge 0$ is a parameter selected by the user that determines the form of the specific sampler used to generate new samples. In \cite{song2020score}, the authors propose $\alpha = 1$. Another interesting choice, which is also discussed in \cite{song2020score}, is $\alpha = 0$. With this choice, the reverse drift-diffusion process reduces to 
\begin{equation}\label{eq:reverse-drift}
		\frac{\partial \probt{\tau}(\x)}{\partial \tau} = - \nabla \cdot \left(  \left( \frac{b(t)}{2} \x + \frac{g(t)}{2}  \bm{s}_t(\x) \right) \probt{\tau}(\x) \right),
\end{equation}
which is the continuity equation for particles or samples being advected by the velocity $\frac{b(t)}{2} \x + \frac{g(t)}{2}  \bm{s}_t(\x)$. Thus, if initially one samples $\x_0 \sim \mathcal{N}(\bm{0}, \sigma^2(T)\mathbb{I})$ and evolves them according to the ODE
\begin{equation}\label{eq:reverse_ODE}
		\frac{\mathrm{d}\x_\tau}{\mathrm{d} \tau} =  \frac{b(t)}{2} \x + \frac{g(t)}{2}  \bm{s}_t(\x) , 
\end{equation}
then at $\tau = T$, $\x_T \sim \prob{\mathrm{data}}(\x)$, the desired data density. This ODE is often referred to as the probability flow ODE~\cite{song2020score}, and once a means to determine the right hand side is available, it may be integrated using any explicit time integration scheme.

\subsection{Score matching}

Integrating the SDE in \Cref{eq:reverse_SDE} or the ODE in \Cref{eq:reverse_ODE}, requires the evaluation of the right hand side at any time $t$, which in turn requires the evaluation of the score function of $\prob{t}(\x)$ for any value of $\x$. Diffusion models approximate this time-dependent score function using a neural network, the so-called score network. We denote the score network using $s_{\thetaa}(\x, t)$, where the subscript $\thetaa$ denotes the learnable parameters (weights and biases) of the score network. We can learn these parameters by minimizing an objective function, say $\mathcal{L}$, \ie
\begin{equation}
\thetaa^\ast = \arg\min_{\thetaa} \mathcal{L}(\thetaa).
\end{equation}

The trained score network using the parameters $\thetaa^\ast$ is used to generate samples from $\prob{\mathrm{data}}$ upon replacing $\bm{s}_t(\x)$ with $s_{\thetaa^\ast}(\x, t)$ in \Cref{eq:euler_maruyama} or \Cref{eq:reverse_ODE}. The objective function $\mathcal{L}$ should measure the quality of the approximation from using the score network, and \citet{hyvarinen2005estimation} propose using the Fisher divergence
\begin{equation}\label{eq:fisher-divergence}
\mathcal{L}(\thetaa) = \int_{0}^{T} \int_{\Omega_{\mathcal{X}}} \lvert  s_{\thetaa}(\x, t) - \nabla \log \prob{t}(\x) \rvert_2^2  \; \prob{t}(\x)   \mathrm{d}\x dt.
\end{equation}
We will now simplify \Cref{eq:fisher-divergence} and show how \Cref{eq:fisher-divergence} can be estimated using realizations $\left\{\x^{(i)}_{\mathrm{data}}\right\}_{i=1}^{N}$. We begin by expanding \Cref{eq:fisher-divergence} and collecting terms that do not depend on $\thetaa$ in the ``constants'' $K_1, K_2$,
\begin{equation}\label{eq:score-matching-loss}
\begin{split}
\mathcal{L}(\thetaa) &= \int_{0}^{T} \int_{\Omega_{\mathcal{X}}} \Big[  \lvert s_{\thetaa}(\x, t) \rvert_2^2 \prob{t}(\x) - 2 s_{\thetaa}(\x, t) \cdot \nabla \prob{t}(\x) \Big]   \; \mathrm{d}\x \mathrm{d}t +K_1 \\
	&= \int_{0}^{T} \int_{\Omega_{\mathcal{X}}} \Bigg[  \lvert s_{\thetaa}(\x, t) \rvert_2^2 \int_{\Omega_{\mathcal{X}}} \prob{t}{ (\x \vert \x^\prime) } \prob{\mathrm{data}}{(\x^\prime)} \mathrm{d}\x^\prime \\
			&\hphantom{= \int_{0}^{T} \int_{\Omega_{\mathcal{X}}} \Bigg[  \lvert s_{\thetaa}(\x, t) \rvert_2^2} - 2 s_{\thetaa}(\x, t) \cdot \int_{\Omega_{\mathcal{X}}} \nabla \prob{t}{ (\x \vert \x^\prime) } \prob{\mathrm{data}}{(\x^\prime)} \mathrm{d}\x^\prime \Bigg]   \; \mathrm{d}\x \mathrm{d}t +K_1 \\
			&= \int_{0}^{T} \int_{\Omega_{\mathcal{X}}} \int_{\Omega_{\mathcal{X}}} \Bigg[  \lvert s_{\thetaa}(\x, t) \rvert_2^2 - 2 s_{\thetaa}(\x, t) \cdot  \nabla \log \prob{t}{ (\x \vert \x^\prime) } \Bigg]  \\
			&\hphantom{\int_{0}^{T} \int_{\Omega_{\mathcal{X}}} \int_{\Omega_{\mathcal{X}}} \Bigg[  \lvert s_{\thetaa}(\x, t) \rvert_2^2 - 2 s_{\thetaa}(\x, t) \nabla \log} \prob{t}{ (\x \vert \x^\prime) } \prob{\mathrm{data}}{(\x^\prime)} \mathrm{d}\x^\prime  \; \mathrm{d}\x \mathrm{d}t +K_1 \\
			&= \int_{0}^{T} \int_{\Omega_{\mathcal{X}}} \int_{\Omega_{\mathcal{X}}} \Bigg[  \lvert s_{\thetaa}(\x, t) - \nabla \log \prob{t}{ (\x \vert \x^\prime) } \rvert_2^2 \Bigg] \prob{t}{ (\x \vert \x^\prime) } \prob{\mathrm{data}}{(\x^\prime)} \mathrm{d}\x^\prime  \; \mathrm{d}\x \mathrm{d}t +K_1 + K_2 \\
			&= \int_{0}^{T} \int_{\Omega_{\mathcal{X}}} \int_{\Omega_{\mathcal{X}}} \Bigg[  \lvert s_{\thetaa}(\x, t) -  \frac{ m(t) \x^\prime  - \x }{\sigma^2(t)} \rvert_2^2 \Bigg] \prob{t}{ (\x \vert \x^\prime) } \prob{\mathrm{data}}{(\x^\prime)} \mathrm{d}\x^\prime  \; \mathrm{d}\x \mathrm{d}t +K_1 +K_2.
		\end{split}
	\end{equation}
In the first step above we have expanded the square in \Cref{eq:fisher-divergence}, recognized that $\nabla \log \prob{t} = \nabla \prob{t} / \prob{t}$, and absorbed the term that does not depend on $\thetaa$ into $K_1$. In the second step, we have used \Cref{eq:forward-diffusion} to rewrite $\prob{t}(\x)$. In the fourth step, we have completed the square with a term that does not depend on $\thetaa$, and denoted it by $-K_2$. Finally, in the last step, we have used the expression for the score function of the diffusion kernel from \Cref{eq:diff-kernel-score}. We note that we wish to minimize this expression for the loss function, and since the terms $K_1$ and $K_2$ do not depend on $\thetaa$ they may be dropped from the expression above. 

In practice, the loss function \Cref{eq:score-matching-loss} is approximated by its Monte Carlo approximation given by
\begin{equation}\label{eq:score-matching-loss2}
\mathcal{L}(\thetaa) = \frac{1}{N} \sum_{i=1}^{N} \Bigg\lvert s_{\thetaa}(\x^{(i)}, t^{(i)}) + \frac{\z^{(i)}}{\sigma(t^{(i)})} \Bigg\rvert_2^2 ,
\end{equation}
where $t^{(i)} \sim \mathcal{U}(0,T)$, $\x^{\prime (i)} \sim \prob{\mathrm{data}}{(\x^\prime)}$, $\bm{z}^{(i)} \sim \mathcal{N}(\bm{0}, \mathbb{I})$, and from \Cref{eq:forward_sample} we have $\x^{(i)} = m(t^{(i)}) \x^{\prime (i)} + \sigma(t^{(i)}) \z^{(i)}$. Also, in practice, $\mathcal{L}(\thetaa)$ is scaled by $\sigma^2(t^{(i)})$ to ensure numerical stability for small values of $\sigma(t^{(i)})$. This leads to the \emph{denoising score matching loss}~\cite{song2019generative,song2020improved}
\begin{equation}\label{eq:denoise-score-matching-loss}
		\mathcal{L}(\thetaa) = \frac{1}{N} \sum_{i=1}^{N} \Big\lvert \sigma(t^{(i)}) s_{\thetaa}(\x^{(i)}, t^{(i)}) + \z^{(i)} \Big\rvert_2^2. 
\end{equation}

\section{Probabilistic inverse problem} \label{sec:inverse-problem}
	
We have seen how diffusion models can be used to solve the generative problem. Next, we demonstrate how the conditional version of diffusion models can be used to solve inverse problems. We start with the preliminaries, first defining the probabilistic inverse problem, then formulating it as a conditional generative problem, and then developing conditional diffusion models to solve it. 

Let $\X$ and $\Y$ be random vectors with joint distribution $\prob{\X\Y}$. We treat $\X$ as the vector of quantities we wish to infer and $\Y$ as the vector of measurements. In what follows, we often refer to $\X$ as the inferred vector and $\Y$ as the measurement vector. For example, in the problem considered in this study, $\X$ represents the distribution of flux of a chemical species over the boundary of a fluid domain. Similarly, $\Y$ represents the concentration of the species measured at a few select measurement points. The goal of the inverse problem is to determine possible values of $\X$ corresponding to an observation $\Y = \hat{\y}$.

Without loss in generality, we will assume that $\x \in \Omega_\mathcal{X} \subseteq \mathbb{R}^{\Nx}$, $\y \in \Omega_\mathcal{Y} \subseteq \mathbb{R}^{\Ny}$, and the forward model and the measurement operator relating the two vectors to be encoded in the conditional distribution $\prob{\Y\vert\X}\!\left(\y \vert \x \right)$. For a given value of $\X = \x$, characterizes all possible values of $\Y$ and therefore contains the effect of the forward model and the measurement operator. Notably, when solving the inverse problem, we will not require knowledge of the explicit form of this conditional distribution. Rather, we will rely only on the ability to generate samples from it for a given value of $\X$. The inverse problem can be stated as: given the forward model and the measurement operator, some prior information about $\X$, and a measurement $\Y = \hat{\y}$ characterize the conditional distribution $\prob{\X\vert\Y}\!\left(\x \vert \hat{\y}\right)$. 

The ``typical'' approach to solving the inverse problem involves using Bayes' theorem to write 
\begin{equation}\label{eq:Bayes}
\prob{\X\vert\Y}\!\left(\x \vert \hat{\y}\right) \propto \prob{\Y\vert\X}\!\left(\hat{\y} \vert \x \right) \probb{\X}{\x},
\end{equation}
where $\prob{\X}$ denotes the density of the \emph{prior} probability distribution of $\X$, $\prob{\Y \vert \X}$ denotes the density corresponding to the \emph{likelihood} of $\Y$ conditioned on $\X$, and $\prob{\X\vert\Y}\!\left(\cdot \vert \hat{\y}\right)$ is the \emph{posterior} distribution. Therefore, the goal of solving the inverse problem is to sample the posterior distribution. These samples represent candidate solutions to the inverse problem. There is extensive research on using the Bayesian approach to solving the probabilistic inverse problem, and the reader is referred to several outstanding texts and reviews on this topic \cite{kaipio2006statistical,stuart2010inverse,calvetti2018inverse}. 

When the prior probability distribution is Gaussian, the forward model and the measurement operator are linear, and the measurement noise is additive and Gaussian, the posterior distribution is also Gaussian. In this case, the mean and covariance of the posterior distribution can be evaluated relatively easily in order to completely quantify the posterior distribution. This approach can be extended to the case where the forward model and/or the measurement operator are non-linear, by working with their linearization about the maximum a-posteriori point (MAP). This approximation is often referred to as the Laplace approximation \cite{laplace1986memoir,bui2012extreme}. 

When the probability densities $\prob{\X}$ and $\prob{\Y \vert \X}$ are non-Gaussian but explicitly known so that they can be evaluated for given values of $\x$ and $\hat{\y}$, the right hand side of \Cref{eq:Bayes} can be determined. In this case, techniques based on Markov Chain Monte Carlo (MCMC) methods and their variants are used to construct a Markov chain whose stationary measure is the desired conditional distribution. This allows a practitioner to generate samples from the desired conditional distribution $\prob{\X\vert\Y}$. However, these methods do not scale well to large dimensions ($\Nx > 10^2$) and to cases where the inverse problem must be solved for multiple measurements. Further, these techniques cannot be applied to problems where the expression for $\prob{\Y \vert \X}$ is not explicitly known, which can often be the case in real experiments where the measurement operator is complex, such as those containing complex, non-additive noise. 

\subsection{Inverse problem as a conditional generative problem}

The key idea in addressing the challenges described above is to recognize that we can generate samples from the conditional distribution that we wish to characterize, that is $\prob{\X\vert\Y}\!\left(\x \vert \hat{\y}\right)$, if we are able to 
\begin{enumerate}
    \item Develop an algorithm that can generate samples from a conditional distribution and can be trained using samples from the joint distribution. 
    \item Find a means to generate samples from the joint distribution $\prob{\X\Y}$. 
\end{enumerate}
As described in the next section, the first part of the first requirement is solved by conditional diffusion models. The second part of the first requirement is met by generating the paired data by sampling $\Y$ from $\prob{\Y\vert\X}\!\left(\y \vert \x \right)$ for different realizations of $\X$ sampled from the prior distribution $\prob{\X}$. This also means that we can interface with the forward model and the measurement operator in a black-box fashion: we only need to generate samples from $\prob{\Y\vert\X}\!\left(\y \vert \x \right)$ for different realizations of $\X$. This is particularly useful for problems with complex physics, considering that most forward models involve sophisticated computational codes that are difficult to alter in any significant way. Moreover, depending on the measurement modality, the measurement noise may be non-Gaussian and non-additive. These scenarios usually pose significant challenges when sampling posterior distributions with conventional MCMC methods but are easily handled within the approach described above. 

In the following section, we will show how conditional diffusion models can be used to sample $\prob{\X\vert\Y}$ using a neural network approximation of the conditional density's score function.  Mainly, we will show that the score function that is necessary for sampling can be derived from the forward and reverse diffusion processes for a given realization of $\Y$, and then derive the score matching loss for training the score network using samples from the joint distribution. Our presentation in \Cref{sec:cond-gen} will closely follow \Cref{sec:uncond-gen} so that reader can draw one-to-one correspondence between unconditional and conditional generation. 	
	
\section{Conditional diffusion models}
\label{sec:cond-gen}
	
In \Cref{sec:uncond-gen}, we derived a family of diffusion models for unconditional generation. Now, we develop conditional diffusion models to solve probabilistic inverse problems. 
	
\subsection{Theory of conditional diffusion}
	
We first derive the forward process that evolves the density $\prob{t}(\x \vert \y)$ such that $\prob{0}(\x \vert \y)$ is the target conditional density $\prob{\X\vert\Y}(\x\vert \y)$. As before, let $\prob{t}(\x\vert\y)$ satisfy the drift-diffusion equation
    \begin{equation}
\label{eq:pde-forward-combined-cond}
    \frac{\partial \prob{t}(\x\vert \y) }{\partial t} -\frac{b(t)}{2}  \nabla \cdot(\x \prob{t}(\x\vert \y)) -\frac{g(t)}{2} \Delta  \prob{t}(\x\vert \y)  = 0,
\end{equation}
which has the solution
\begin{equation}
\prob{t}(\x \vert \y)  = \int_{\Omega_{\mathcal{X}}} \prob{t}{ (\x \vert \x^\prime) } \prob{\X \vert \Y}{(\x^\prime \vert \y)} \mathrm{d}\x^\prime,
\end{equation}
where, as before, $\prob{t}{ (\x \vert \x^\prime) }$ is a Gaussian kernel with mean $m(t) \x^\prime$ and variance $\sigma^2(t) \mathbb{I}$. The functions $g(t), b(t), m(t)$ and $\sigma(t)$ have the same definition in the unconditional case; see \Cref{tab:diff-formulations}. The $\nabla$ and the $\Delta$ operators in this equation contain derivatives only along the $\x$ (and not $\y$) coordinates. It is apparent from this that we treat the random variables $\X$ and $\Y$ differently. We apply diffusion to the former, but not to the latter.  

To derive the drift-diffusion equation associated with the reverse process, we use the variable transformation $\tau = T - t$ again, let $\probt{\tau}(\x \vert \y) = \prob{t}(\x \vert \y)$, and follow the same steps as before to obtain the counterpart of \Cref{eq:reverse-drift-diffusion},
\begin{equation}\label{eq:reverse-drift-diffusion-cond}
\frac{\partial \probt{\tau}(\x\vert \y)}{\partial \tau} = - \nabla \cdot \left(  \left( \frac{b(t)}{2} \x + \frac{(1+ \alpha) g(t)}{2}  \bm{s}_t(\x, \y) \right) \probt{\tau}(\x\vert \y) \right) + \frac{\alpha g(t)}{2} \Delta  \probt{\tau}(\x\vert \y),
\end{equation}
where $\bm{s}_t(\x,\y) = \nabla \log \prob{t}(\x\vert \y)$ is the score function of the conditional distribution for the forward process. If we set $\probt{0}(\x \vert \y) = \mathcal{N}(\bm{0}, \sigma^2(T)\mathbb{I})$, and solve this PDE, then by construction $\probt{\tau}(\x \vert \y) = \prob{t}(\x \vert \y)$, and therefore $\probt{T}(\x \vert \y) = \prob{0}(\x \vert \y) = \prob{\X \vert \Y}(\x \vert \y)$. 

The drift-diffusion equation above is the Fokker-Planck equation for the evolution of the probability density of a stochastic process governed by the Ito SDE 
\begin{equation}\label{eq:reverse_SDE-cond}
		\mathrm{d}\x_\tau =  \left( \frac{b(t)}{2} \x + \frac{(1+ \alpha) g(t)}{2}  \bm{s}_t(\x,\y) \right) \mathrm{d}\tau + \sqrt{\alpha g(t) } \mathrm{d}\w_\tau,
\end{equation}
where $\w_\tau$ denotes the $\Nx$-dimensional Wiener process. Now, if $\x_0 \sim \mathcal{N}(\bm{0}, \sigma^2(T)\mathbb{I})$ and is evolved according to this SDE, then at $\tau = T$, $\x_T \sim \prob{\X\vert\Y}(\x\vert \y)$, the desired conditional density. 

In practice, this SDE may be integrated using the Euler-Maruyama method which provides an update for $\x_{\tau + \Delta \tau} $, given $\x_\tau$, 
\begin{equation}\label{eq:euler_maruyama-cond}
\x_{\tau + \Delta \tau} = \x_{\tau } + \left( \frac{b(t)}{2} \x + \frac{(1+ \alpha)g(t)}{2}  \bm{s}_t(\x,\y) \right) \Delta \tau + \sqrt{\alpha g(t) \Delta \tau} \z,
\end{equation}
where $\z$ are sampled independently from the $\Nx$-dimensional standard normal distribution.

With the choice $\alpha = 0$, \Cref{eq:reverse-drift-diffusion-cond} reduces to, 
\begin{equation}\label{eq:reverse-drift-cond}
		\frac{\partial \probt{\tau}(\x\vert \y)}{\partial \tau} = - \nabla \cdot \left(  \left( \frac{b(t)}{2} \x + \frac{g(t)}{2}  \bm{s}_t(\x,\y) \right) \probt{\tau}(\x\vert \y) \right),
\end{equation}
which is the continuity equation for particles being advected by the velocity $\frac{b(t)}{2} \x + \frac{g(t)}{2}  \bm{s}_t(\x, \y)$. Thus, at $\tau = 0$, if particles are selected so that $\x_0 \sim \mathcal{N}(\bm{0}, \sigma^2(T)\mathbb{I})$ and evolved according to
\begin{equation}\label{eq:reverse_ODE-cond}
\frac{\mathrm{d}\x_\tau}{\mathrm{d} \tau} = \frac{b(t)}{2} \x + \frac{g(t)}{2}  \bm{s}_t(\x,\y) 
\end{equation}
then at $\tau = T$, $\x_T \sim \prob{\X\vert\Y}(\x\vert \y)$, the desired conditional data density. This ODE may be integrated using any explicit time integration scheme.

\subsection{Conditional score matching}

The process of generating samples in a conditional diffusion model requires the knowledge of the score function of the time-dependent conditional probability distribution for all values of the random vectors $\X$ and $\Y$. This function is denoted by $\bm{s}_t(\x,\y)$ in the previous section and is approximated using a neural network denoted by $s_{\thetaa}(\x, \y, t)$, where the subscript $\thetaa$ denotes the learnable parameters of the score network. This network has to be learned using samples from the joint distribution $\prob{\X \Y}$. It is not immediately clear how this can be accomplished; however, as described below a loss function that measures the difference between the true score function and its approximation can be formulated as a Monte Carlo sum that only requires samples from the joint distribution. 

We begin with a loss function defined as 
\begin{equation}\label{eq:fisher-divergence-cond}
\mathcal{L}(\thetaa) = \int_{\Omega_{\mathcal{Y}}} \!\Big[ \int_{0}^{T}\!\!\int_{\Omega_{\mathcal{X}}}\!\lvert  s_{\thetaa}(\x, \y, t) - \nabla \log \prob{t}(\x|\y) \rvert_2^2  \; \prob{t}(\x)   \mathrm{d}\x \mathrm{d}t \Big] \prob{\Y}(\y) \mathrm{d}\y.
\end{equation}
Thereafter, we recognize that the expression within the square parentheses is of the same form as the loss function for the unconditional case \Cref{eq:fisher-divergence}, and following the steps outlined in \Cref{sec:uncond-gen}, we arrive at 
\begin{equation}\label{eq:score-matching-loss-cond}
\begin{split}
\mathcal{L}(\thetaa) &= \int_{\Omega_{\mathcal{Y}}} \Big[ \int_{0}^{T} \int_{\Omega_{\mathcal{X}}} \int_{\Omega_{\mathcal{X}}}   \Big\lvert s_{\thetaa}(\x, \y, t) -  \frac{ m(t) \x^\prime  - \x }{\sigma^2(t)} \Big\rvert_2^2 \prob{t}{ (\x \vert \x^\prime) } \prob{\X|\Y}{(\x^\prime \vert \y)} \; \mathrm{d}\x^\prime  \mathrm{d}\x \mathrm{d}t \Big] \prob{\Y}(\y) \mathrm{d}\y +K \\
&= \int_{0}^{T} \int_{\Omega_{\mathcal{X}}} \int_{\Omega_{\mathcal{Y}}}\int_{\Omega_{\mathcal{X}}}   \Big\lvert s_{\thetaa}(\x, \y, t) -  \frac{ m(t) \x^\prime  - \x }{\sigma^2(t)} \Big\rvert_2^2 \prob{t}{ (\x \vert \x^\prime) } \prob{\X \Y}{(\x^\prime, \y)} \; \mathrm{d}\x^\prime \mathrm{d}\y \mathrm{d}\x \mathrm{d}t +K. 
\end{split}
\end{equation}
To arrive at the second equality above we have changed the order of integrations, and used the fact that $\prob{\X \Y}{(\x^\prime, \y)}  = \prob{\X|\Y}{(\x^\prime \vert \y)} \prob{\Y}(\y) $. The integral above can be approximated by a Monte Carlo sum obtained by sampling from the joint distribution as follows,
\begin{equation}\label{eq:score-matching-loss2-cond}
\mathcal{L}(\thetaa) = \sum_{i=1}^{N} \Bigg\lvert s_{\thetaa}(\x^{(i)}, \y^{(i)}, t^{(i)}) + \frac{\z^{(i)}}{\sigma(t^{(i)})} \Bigg\rvert_2^2,
\end{equation}
where $t^{(i)} \sim \mathcal{U}(0,T)$, $(\x^{\prime (i)},\y^{(i)}) \sim \prob{\X \Y}{(\x^\prime,\y)}$, and from \Cref{eq:forward_sample} we have used $\x^{(i)} = m(t^{(i)}) \x^{\prime (i)} + \sigma(t^{(i)}) \z^{(i)}$. Thus, we have achieved the desired goal of approximating the loss function for the score function with a sum that utilizes samples from the joint distribution.  

Similar to the denoising score matching objective, the loss above is also scaled by $\sigma^2(t)$ to ensure numerical stability for small values of $\sigma(t)$, which leads to the \emph{conditional denoising score matching} loss~\cite{dasgupta2025conditional}
\begin{equation}\label{eq:denoise-score-matching-loss-conditional}
\mathcal{L}(\thetaa) = \sum_{i=1}^{N} \Big\lvert \sigma(t^{(i)}) s_{\thetaa}(\x^{(i)}, \y^{(i)}, t^{(i)}) + \z^{(i)} \Big\rvert_2^2 
\end{equation}
which can be estimated using the paired data. The trained network with parameters $\thetaa^\ast$ is used to sample from $\prob{\X \vert \Y}$ by replacing $s_{t}(\x, \y)$ with $s_{\thetaa^\ast}(\x, \y,  t)$ in \Cref{eq:euler_maruyama-cond} or \Cref{eq:reverse_ODE-cond}. 

\section{Accounting for a family of measurement operators}\label{sec:stochastic-obs}

The conditional diffusion model described in the previous section can be used to solve an inverse problem corresponding to any instance of the measurement $\Y$. However, if the measurement operator is changed, the corresponding diffusion model also must be changed by retraining it with a new set of data generated with the new measurement operator. As described below, this inconvenience may be avoided by training a single diffusion model that would apply to the family of observation operators.

We consider the case where the measurement operator itself is parameterized by a random vector $\Mm$. Thus, the conditional distribution $\prob{\Y\vert\X}\!\left(\y \vert \x \right)$ generalizes to $\prob{\Y\vert\X \Mm}\!\left(\y \vert \x, \mm \right)$ which characterizes the measurements obtained when the measurement operator is defined by $\Mm = \mm$ and in the forward model the parameters to be inferred are set to $\X = \x$. We assume that in addition to the prior distribution for $\X$, we have access to the marginal distribution of $\Mm$ and denote it by $\prob{\Mm}(\mm)$. Under this generalized case, the inverse problem is defined as: given the forward model and the measurement operator, prior information about $\X$ and $\Mm$, and a measurement $\Y = \hat{\y}$ corresponding to a measurement operator $\Mm = \hat{\mm}$, characterize the conditional distribution $\prob{\X\vert\Y \Mm}\!\left(\x \vert \hat{\y}, \hat{\mm} \right)$. 

We convert this problem to a conditional generative problem using the same approach described in the previous section. That is, we generate samples from the joint distribution $\prob{\X \Y \Mm}$, use these to train a conditional diffusion model, and then use the trained diffusion model to generate samples from desired conditional density $\prob{\X\vert\Y \Mm}\!\left(\x \vert \hat{\y}, \hat{\mm} \right)$. In essence, we replace the conditioning random vector $\Y$ in the previous section with the expanded set of random vectors $(\Y, \Mm)$.

In order to generate samples from the joint distribution $\prob{\X \Y \Mm}$, we recognize 
\begin{eqnarray}\label{eq:cond-meas-op}
\prob{\X \Y \Mm}(\x,\y,\mm)  &=&  \prob{\Y \vert \X \Mm}(\y \vert \x, \mm) \prob{\X  \Mm}(\x,\mm)  \nonumber \\
    &=&  \prob{\Y \vert \X \Mm}(\y \vert \x, \mm) \prob{\X}(\x) \prob{\Mm}(\mm).
\end{eqnarray}
In arriving at the second line in the equation above, we have assumed that $\Mm$ and $\X$ are independent random vectors. That is, the prior distribution of $\X$ is independent of the choice of the measurement operator. This is a modeling choice that appears to be reasonable for most scenarios. If this were not the case, then one would replace $\prob{\X}(\x)$ with $\prob{\X \vert \Mm}(\x \vert \mm)$ in the equation above. 

\Cref{eq:cond-meas-op} suggests that the training data for $\X$, $\Y$ and $\Mm$ may be generated by first generating the samples $\Mm$ and $\X$ from $\prob{\Mm}(\mm)$ and $\prob{\X}(\x)$, respectively, and using these in $\prob{\Y \vert \X \Mm}(\y \vert  \x, \mm)$ to generate the corresponding samples of $\Y$.
    
\section{Results} \label{sec:results}

In this section, we present results to assess the performance of the methods described in this study. They include generating samples from a one-dimensional conditional distribution given data from the underlying two-dimensional joint distribution. They also include inferring the boundary flux of a chemical species entering a fluid domain based on sparse and noisy measurements of concentration within the domain. In both cases, we assess the performance of the variance exploding and preserving methods, and sampling based on the probability flow ODE and a SDE, which correspond to setting $\alpha = 0$ and $\alpha = 1$ in \Cref{eq:euler_maruyama-cond}, respectively. 

We perform all numerical experiments using PyTorch~\cite{pytorch2019}. For the experiments in \Cref{subsec:flux-problem1}, we min-max normalize the training data, which includes both the flux and concentration values, between $[0,1]$. We provide additional details regarding the score network and associated training hyper-parameters, variance schedules, and sampling algorithms in \ref{appsubsec:models}.

\subsection{Conditional density estimation}\label{subsec:cond-dens-toy-examples}
\begin{figure}[t]
    \centering
    \captionsetup[subfigure]{justification=centering}
    \begin{subfigure}[b]{0.32\textwidth}
        \centering
        \includegraphics[width=\textwidth]{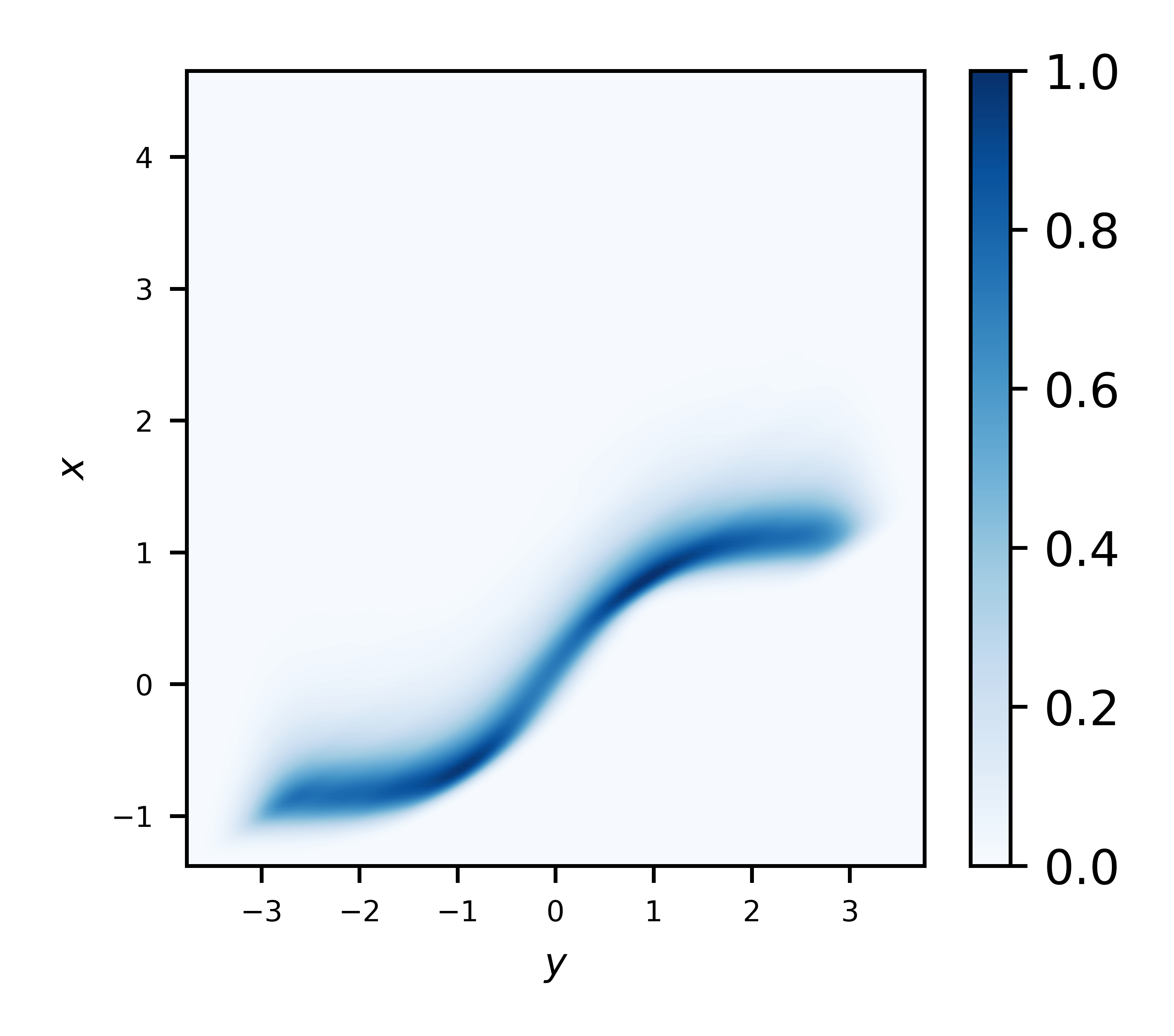}
        \caption{Tanh}
        \label{fig:plot1}
    \end{subfigure}
    \hfill
    \begin{subfigure}[b]{0.32\textwidth}
        \centering
        \includegraphics[width=\textwidth]{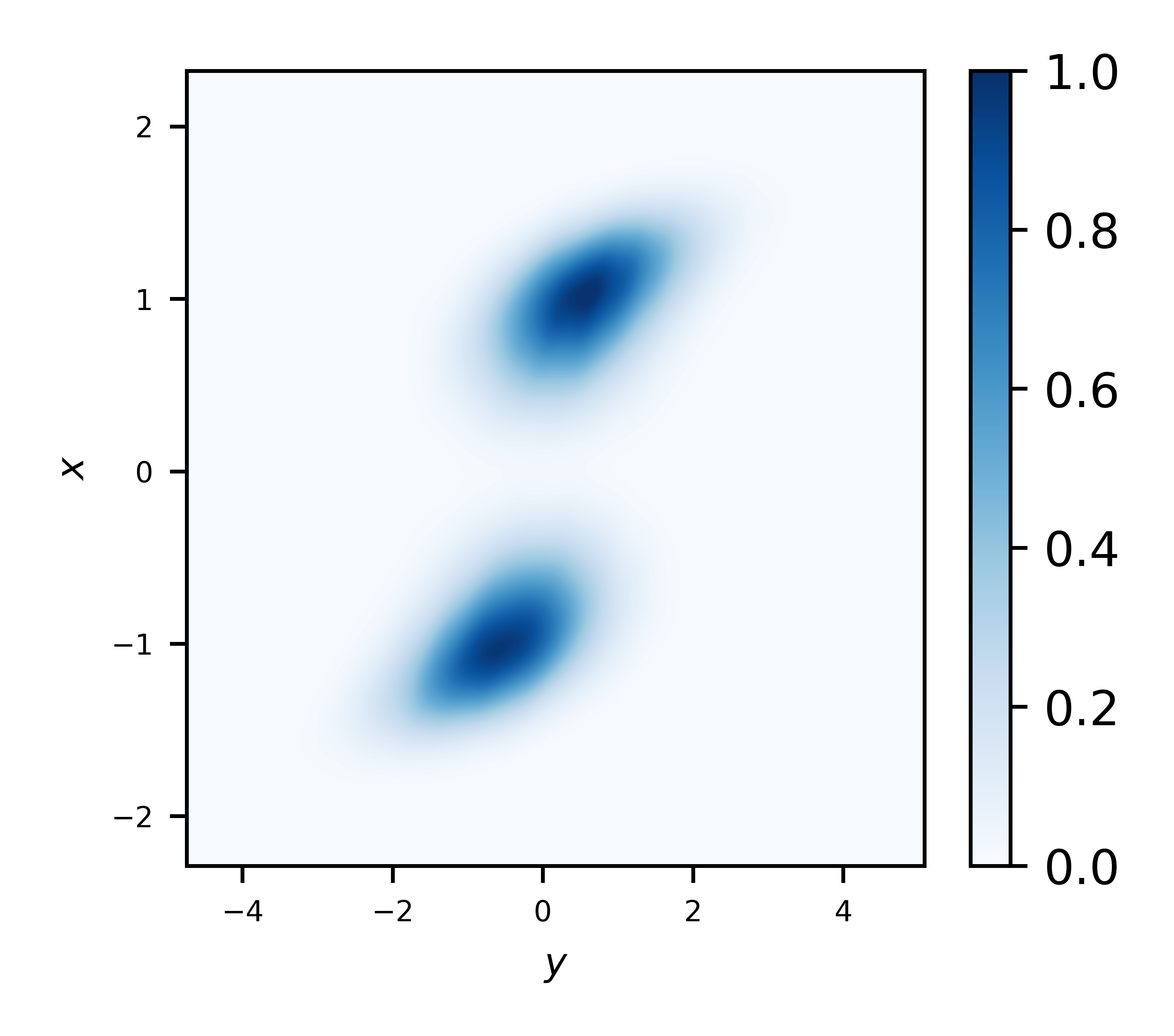}
        \caption{Bimodal}
        \label{fig:plot2}
    \end{subfigure}
    \hfill
    \begin{subfigure}[b]{0.32\textwidth}
        \centering
        \includegraphics[width=\textwidth]{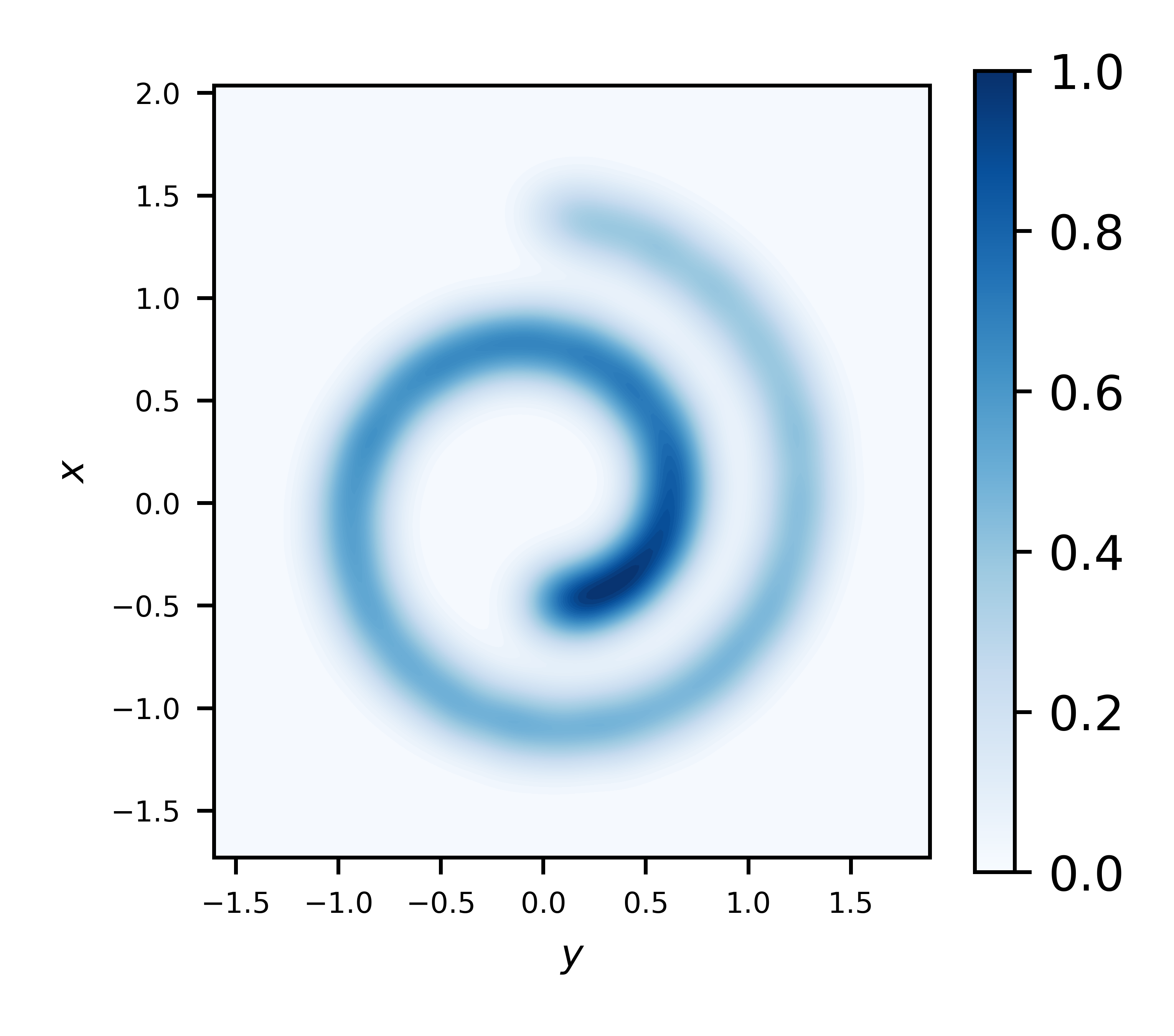}
        \caption{Spiral}
        \label{fig:plot3}
    \end{subfigure}
    \caption{Contour map of the kernel density estimate of joint distribution between $X$ and $Y$ for the (a) Tanh (\ref{eq:tanh}), (b) Bimodal (\ref{eq:bimodal}), and (c) Spiral (\ref{eq:spiral}) cases}
    \label{fig:three_plots}
\end{figure}

In this case, we consider samples from a two-dimensional joint distribution $\prob{XY}(x,y)$ and use these to train a conditional diffusion model to generate samples from the conditional distribution $\prob{X \vert Y}(x|y)$. We consider three joint densities given by 
\begin{equation}\label{eq:tanh}
    X = \tanh (Y) + C_1, \quad Y \sim \mathcal{U}(-3,3), \qquad \text{where } C_1 \sim \Gamma (1,0.3). 
\end{equation}
\begin{equation}\label{eq:bimodal}
    X = (Y+C_2)^{1/3}, \quad  Y \sim \mathcal{N}(0,1), \qquad \text{where } C_2 \sim \mathcal{N}(0,1). 
\end{equation}
\begin{align}
    X &= 0.1 (W \sin(W) +  C_3), \qquad Y = 0.1 (W \cos(W) + C_4), \nonumber \\
    & \text{where } W = 1.5 \pi (1+ 2H), \;  H \sim \mathcal{U}(0,1), \; C_3 \sim \mathcal{N}(0,1),  \; C_4 \sim \mathcal{N}(0,1) \label{eq:spiral}
\end{align}
that are motivated by the examples considered in \cite{ray2023solution}. We refer to the first of these as the Tanh case, the second as the Bimodal case, and the third as the Spiral case. The contour map of the joint density for each case is shown in \Cref{fig:three_plots}.

Using 10,000 samples from the joint density for each case we train a neural network to learn the score of the conditional density for the variance exploding and variance preserving formulations. Thereafter, we specify values of $Y$ and using the trained network generate 10,000 samples of $X$. For sampling, we consider both the ODE sampler ($\alpha = 0$) and the SDE sampler ($\alpha = 1$). 

\begin{figure}[h!]
    \centering
    \captionsetup[subfigure]{justification=centering}

    \begin{subfigure}[b]{\textwidth}
        \centering
        \includegraphics[width=\textwidth,clip, trim={0 0 0 20}]{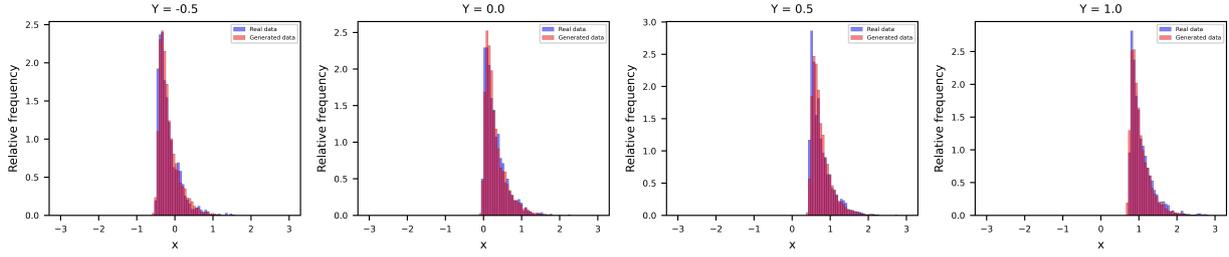}
        \caption{Tanh}
    \end{subfigure}

    \vskip 1em 

    \begin{subfigure}[b]{\textwidth}
        \centering
        \includegraphics[width=\textwidth,clip, trim={0 0 0 20}]{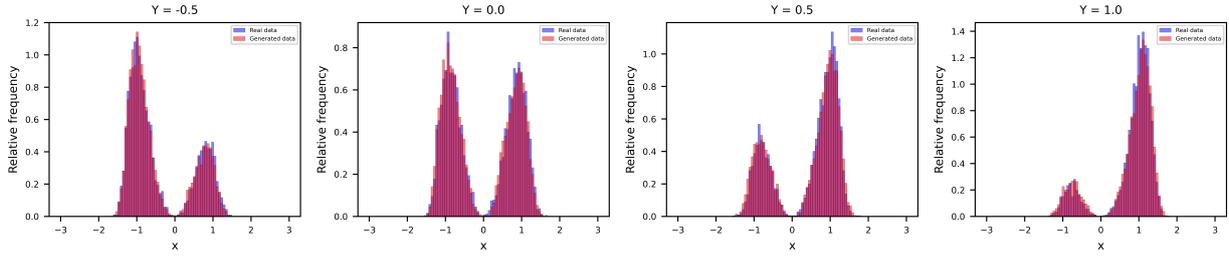}
        \caption{Bimodal}
    \end{subfigure}

    \vskip 1em

    \begin{subfigure}[b]{\textwidth}
        \centering
        \includegraphics[width=\textwidth,clip, trim={0 0 0 20}]{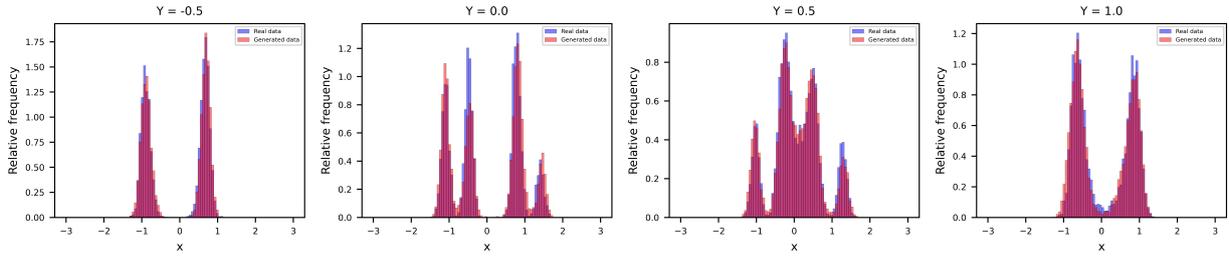}
        \caption{Spiral}
    \end{subfigure}

    \caption{Histograms of generated samples using variance exploding formulation and SDE sampler ($\alpha = 1$) compared to the histograms of the true conditional distribution for $Y \in \{-0.5, 0, 0.5, 1\}$}
    \label{fig:VE_histograms}
\end{figure}

\begin{figure}[h!]
    \centering
    \captionsetup[subfigure]{justification=centering}

    \begin{subfigure}[b]{\textwidth}
        \centering
        \includegraphics[width=\textwidth,clip, trim={0 0 0 20}]{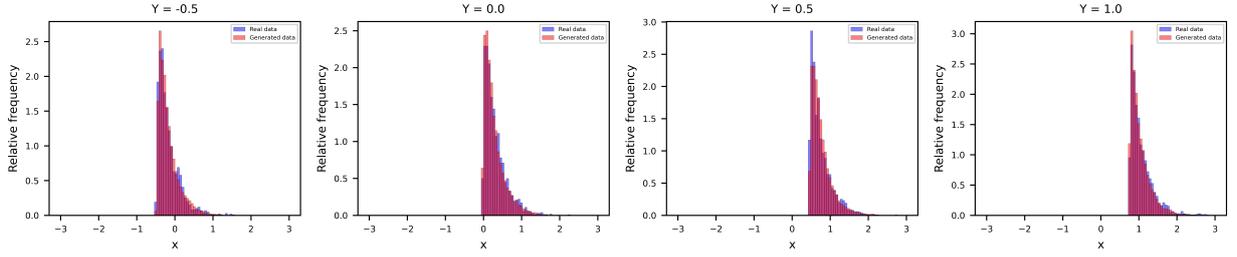}
        \caption{Tanh}
    \end{subfigure}

    \vskip 1em 

    \begin{subfigure}[b]{\textwidth}
        \centering
        \includegraphics[width=\textwidth,clip, trim={0 0 0 20}]{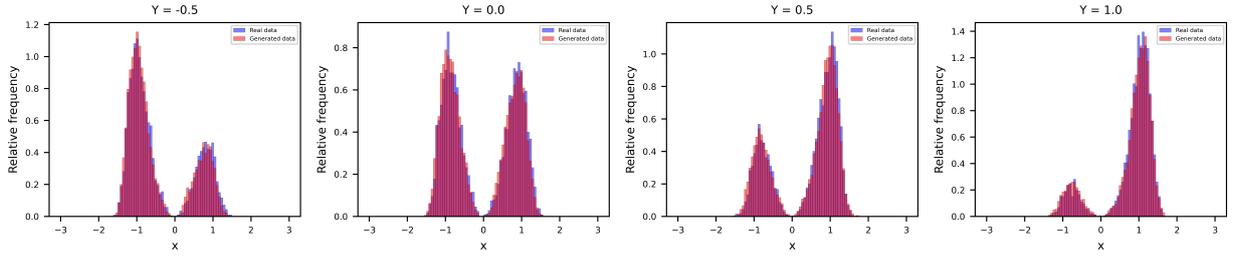}
        \caption{Bimodal}
    \end{subfigure}

    \vskip 1em

    \begin{subfigure}[b]{\textwidth}
        \centering
        \includegraphics[width=\textwidth,clip, trim={0 0 0 20}]{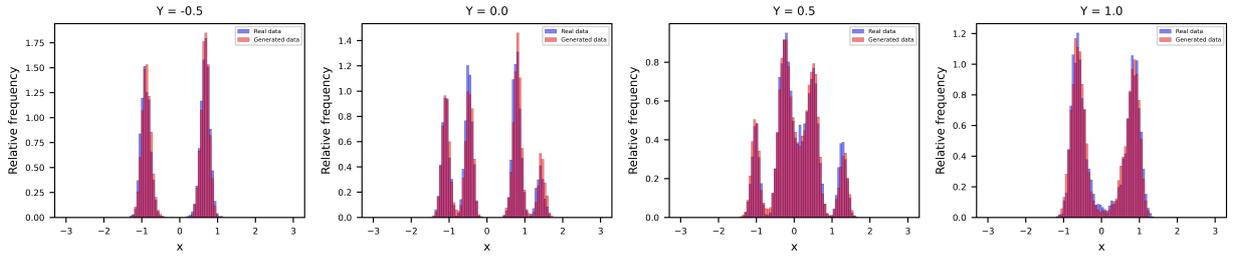}
        \caption{Spiral}
    \end{subfigure}

    \caption{Histograms of generated samples obtained using the variance preserving formulation and SDE sampler compared to the histograms of the true conditional distribution for $Y \in \{-0.5, 0, 0.5, 1\}$}
    \label{fig:VP_histograms}
\end{figure}

In \Cref{fig:VE_histograms}, we present histograms of the generated samples from the conditional density for the variance exploding formulation using the SDE sampler. In each plot, we also include the ``true'' conditional distribution, which is obtained by sampling points in a band of width 0.1 around the specified value of $Y$ from a test set containing 100,000 realizations of $X$ and $Y$. In each case, we observe that histograms generated by the diffusion model closely match the histogram of the true distribution. \Cref{fig:VP_histograms} contains the corresponding results for the variance preserving formulation. Once again, we observe that the generated samples closely conform to the underlying conditional density. 

In \Cref{tab:VE&VP_results}, we quantify the accuracy of diffusion models by computing the regularized optimal transport distance between the generated samples and the reference conditional density, averaged over the four $Y$ values, using the Sinkhorn-Knopp algorithm~\cite{NIPS2013_af21d0c9}.  We use the Python Optimal Transport library \cite{flamary2021pot} to calculate this distance with the regularization term set to 0.01. From \Cref{tab:VE&VP_results}, we observe that while both models perform well, the variance preserving formulation is performs marginally better than the variance exploding formulation for the ODE sampler. For the SDE sampler, the variance exploding formulation performs better in the Spiral case.

\begin{table}[h!]
\centering
\caption{Average regularized optimal transport distance over $Y \in \{-0.5, 0, 0.5, 1\}$ using the variance exploding and variance preserving formulations}
\label{tab:VE&VP_results}
\begin{tabular}{l cccc}
\toprule
\multirow{2}{*}{Formulation} & \multirow{2}{*}{Sampler}& \multicolumn{3}{c}{Dataset}\\
\cline{3-5}
 &  & Tanh & Bimodal &  Spiral \\
\midrule
Variance Exploding & ODE & 0.051 & 0.059  & 0.047\\
 & SDE & 0.040 & 0.043 & 0.052\\
\hline
Variance Preserving & ODE & 0.046 & 0.046 & 0.045\\
 & SDE & 0.041 &0.044 & 0.041\\
\bottomrule
\end{tabular}
\end{table}

\subsection{Estimating boundary flux in an advection diffusion problem}\label{subsec:flux-problem1}

\begin{figure}
    \centering
    \includegraphics[width=0.8\linewidth]{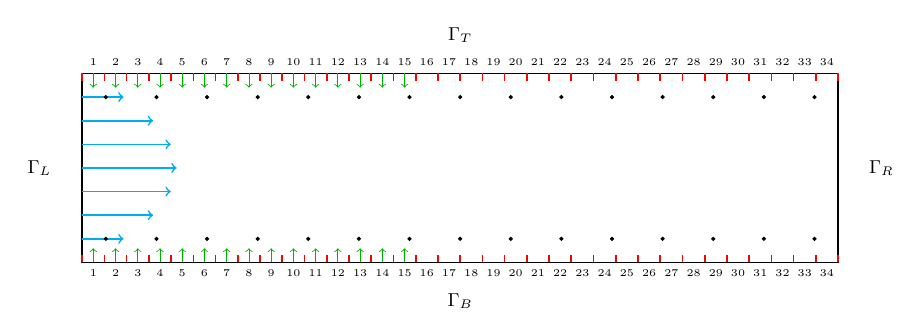}
    \caption{Problem setup illustrating the rectangular domain with parabolic profile flow from left to right. The inflow boundary \( \Gamma_L \) maintains a fixed concentration \( u(0,y) = u_i \), the right boundary \( \Gamma_R \) has a no-flux condition, and the top \( \Gamma_T \) and bottom \( \Gamma_B \) boundaries emit a substance with flux \( q(x, y) \). The dotted points indicate sensor locations where concentration is measured. Top and bottom boundaries are divided into 34 segments each, and green arrows indicate segments that can have non-zero concentration flux} 
    \label{fig:adv-diff-setup}
\end{figure}

In this section, we describe the results obtained by solving a physics-driven inverse problem using conditional diffusion models. The problem setup is described in \Cref{fig:adv-diff-setup}. The domain is a rectangle of dimensions $16 \times 4$ units where the concentration of a chemical species is observed. The concentration is required to obey the advection-diffusion equation
\begin{eqnarray}
\label{eq:advection_diffusion}
    \nabla \cdot (\bm{a} u) - \kappa \nabla^2 u = 0,
\end{eqnarray}
in the domain. Here $\bm{a}$ is directed along the horizontal coordinate and has a parabolic profile with a maximum value of 0.1 units, and $\kappa = 0.07$. Zero concentration is imposed on the left boundary, whereas zero flux is prescribed on the right boundary. The flux is allowed to be non-zero on the top and bottom boundaries, beginning from the left edge to a distance of 7 units. This prescribed flux is constrained to be piecewise constant over 15 segments, each of length 0.47 units. The value of this flux on the top and bottom boundaries, upper and lower walls, respectively, is denoted by the vector $\X$ and is the quantity we wish to infer. Therefore, for this problem $\Nx = 30$. The measurement $\Y$ comprises the noisy concentration values measured by 30 equi-spaced sensors ($\Ny = 30$) located at a distance of 0.5 units away from the bottom and top edges. The noise is additive and independent Gaussian with zero mean and a variance equal to $\sigma^2_{\epsilon}$. 

The prior distribution of $\mathbf{X}$ is defined by a Gaussian process. Flux values for each segment are sampled from a multivariate normal distribution with a radial basis function (RBF) kernel as the covariance function. This kernel is based on the Euclidean distance between segment locations, with a length scale of two, ensuring stronger correlations between nearby segments while still allowing variability in the flux values across segments. The values are initially sampled with zero mean and then shifted by adding 2 to obtain a positive mean flux. Negative values are clipped to zero to ensure all flux values are non-negative.

\begin{figure}[htbp]
    \centering

    \includegraphics[width=0.3\textwidth]{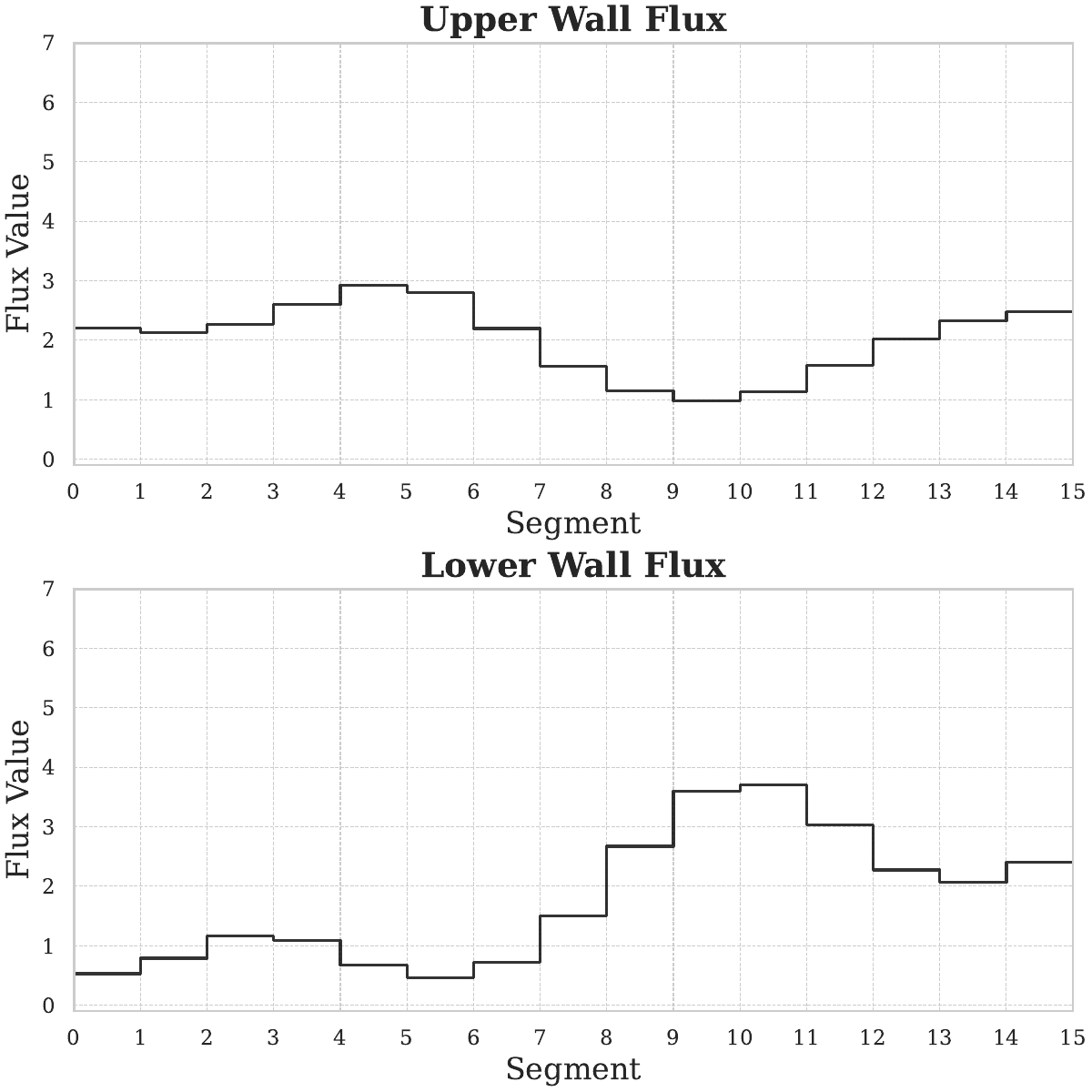}
    \hfill
    \includegraphics[width=0.3\textwidth]{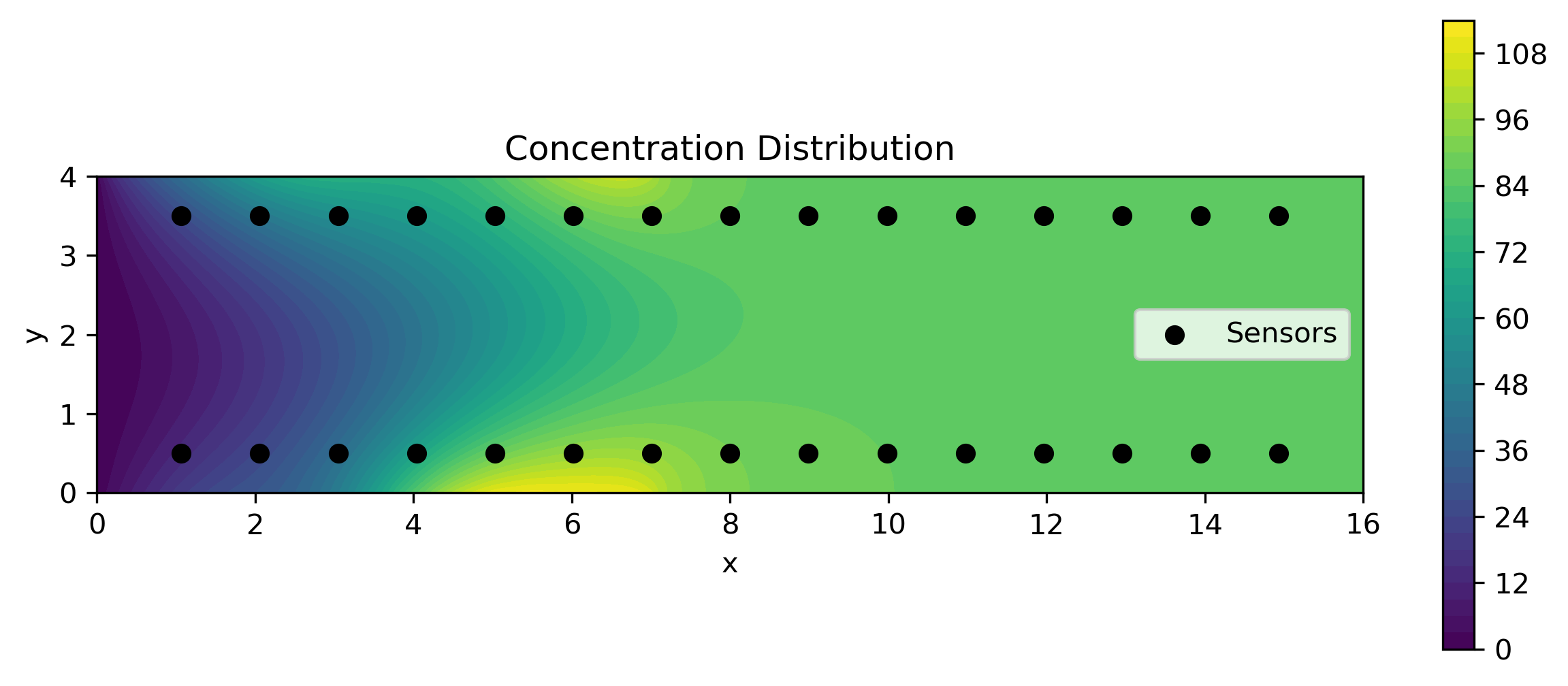}
    \hfill
    \includegraphics[width=0.3\textwidth]{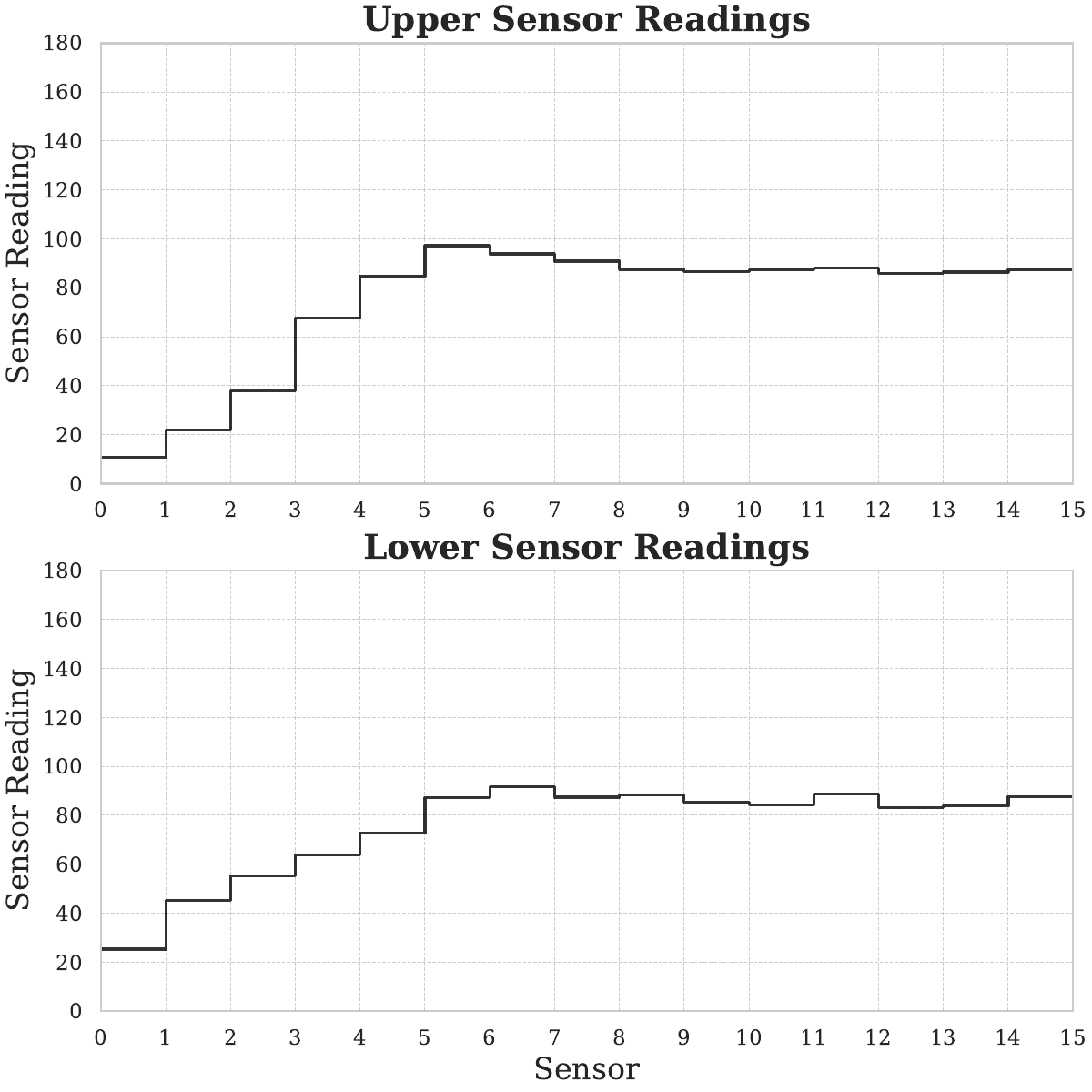}

    \vspace{0.5cm}

    \includegraphics[width=0.3\textwidth]{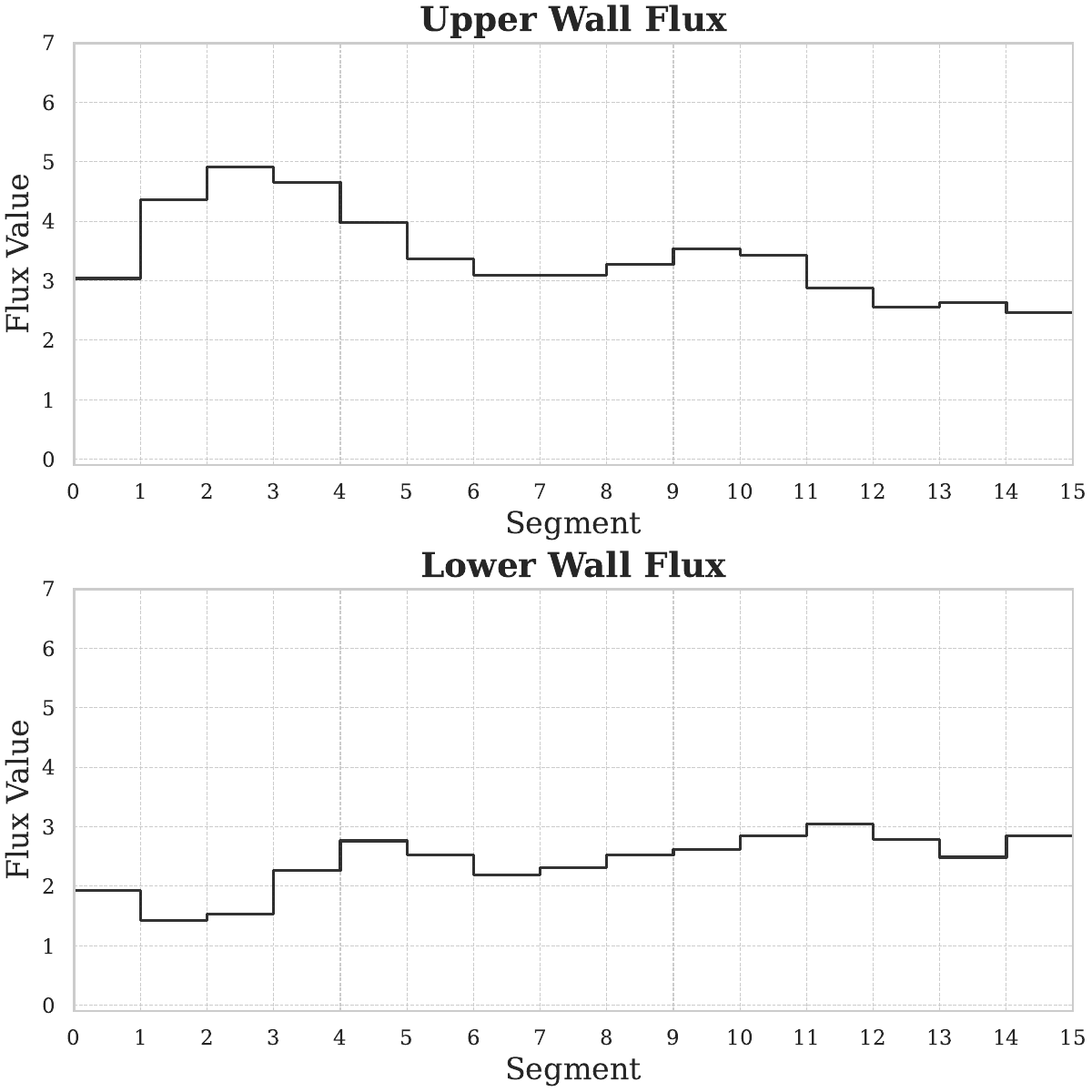}
    \hfill
    \includegraphics[width=0.3\textwidth]{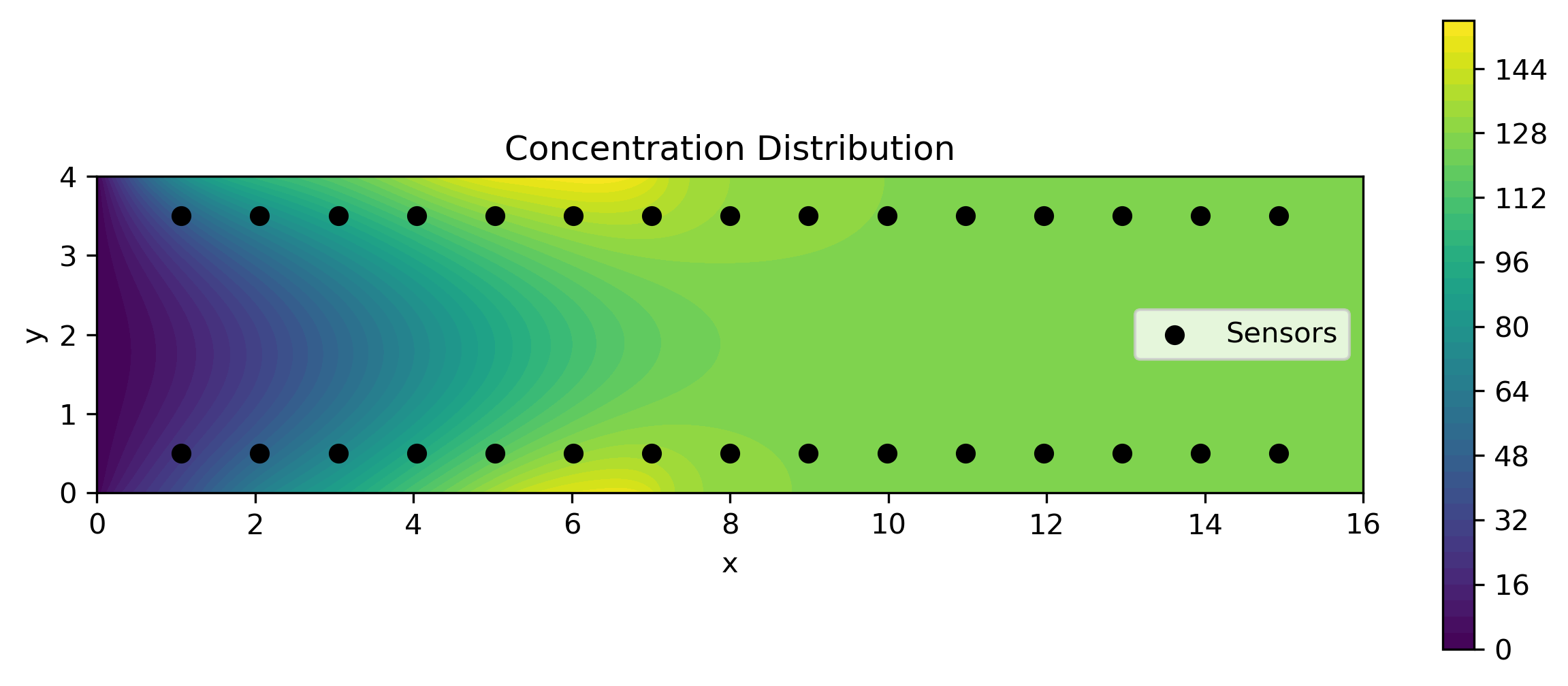}
    \hfill
    \includegraphics[width=0.3\textwidth]{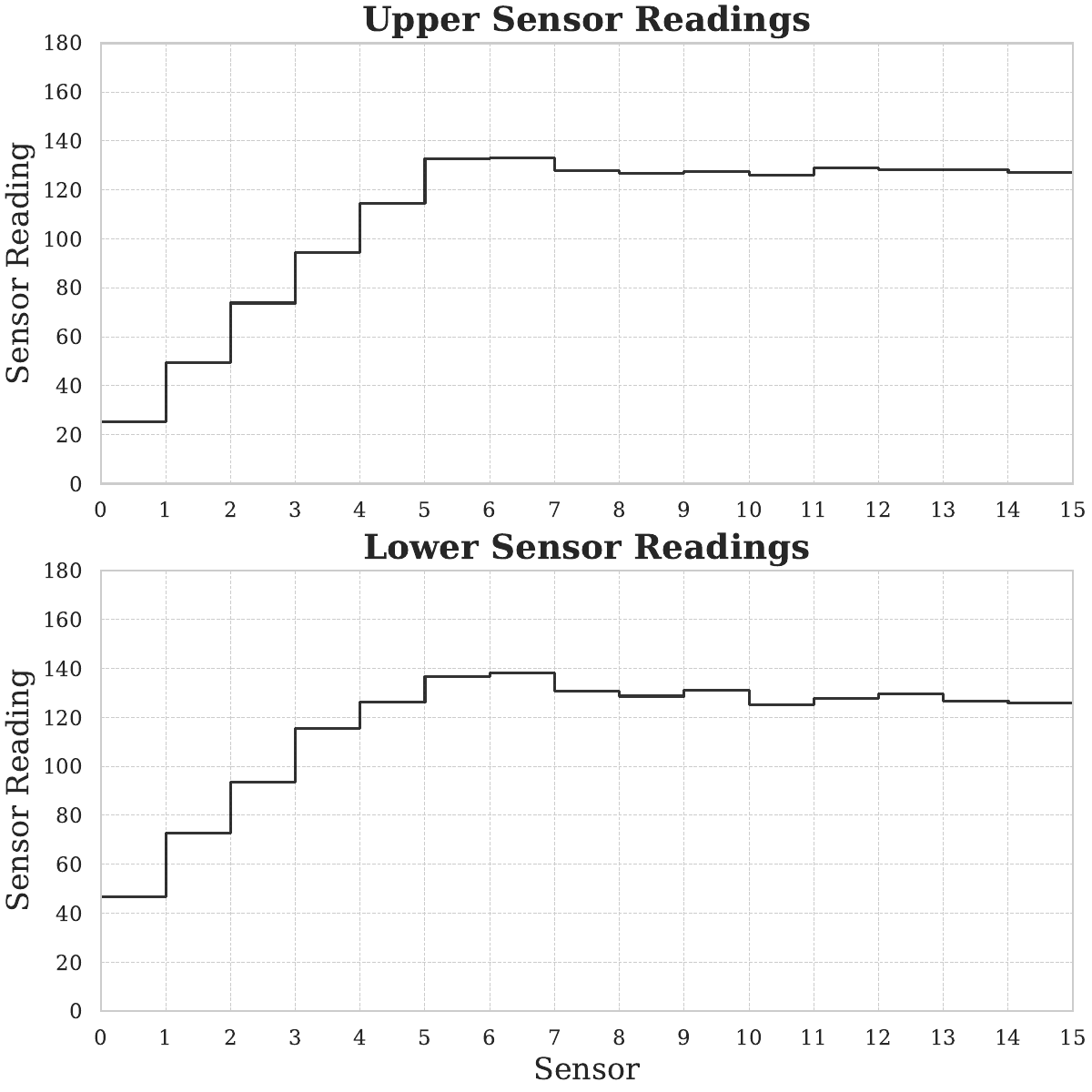}

    \vspace{0.5cm}

    \includegraphics[width=0.3\textwidth]{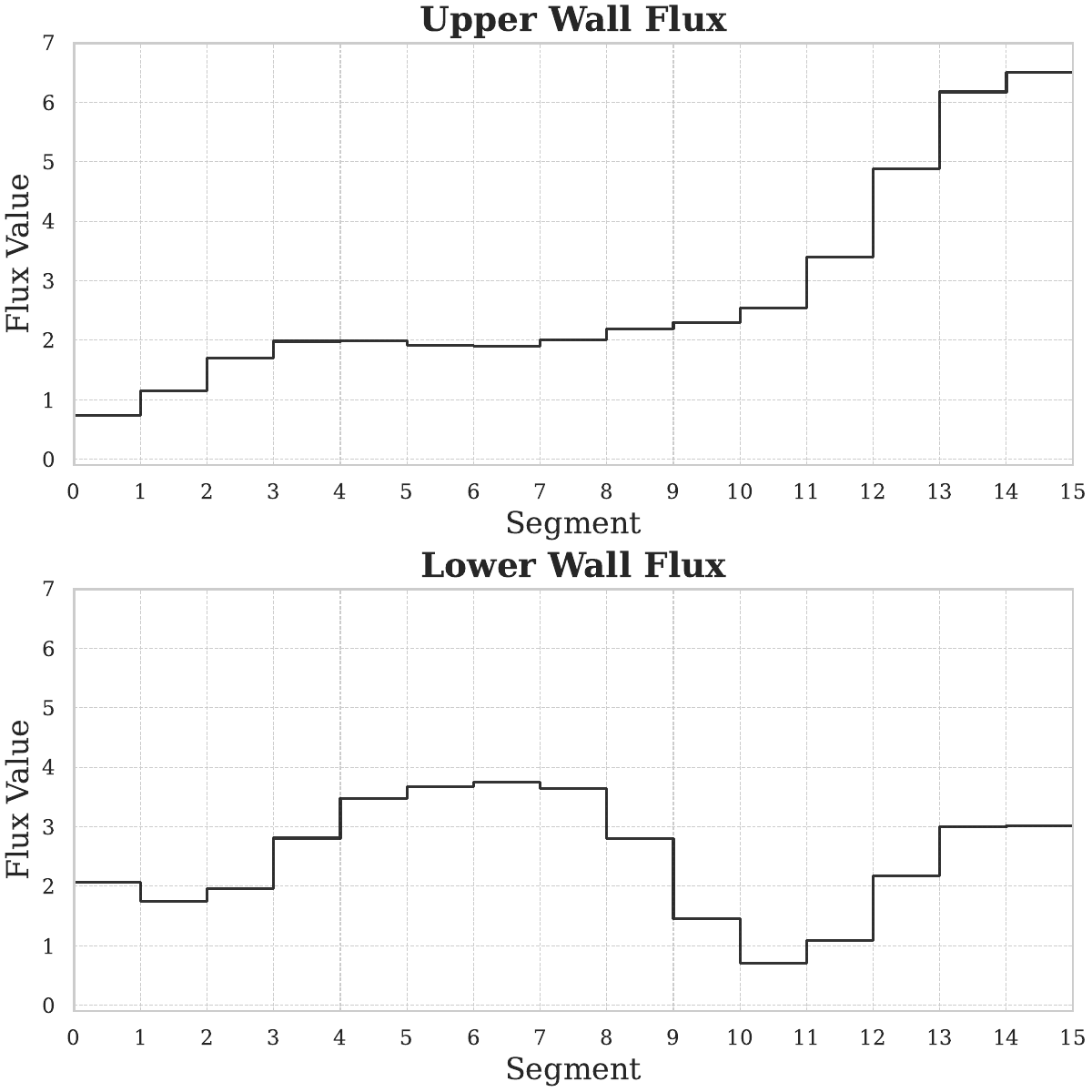}
    \hfill
    \includegraphics[width=0.3\textwidth]{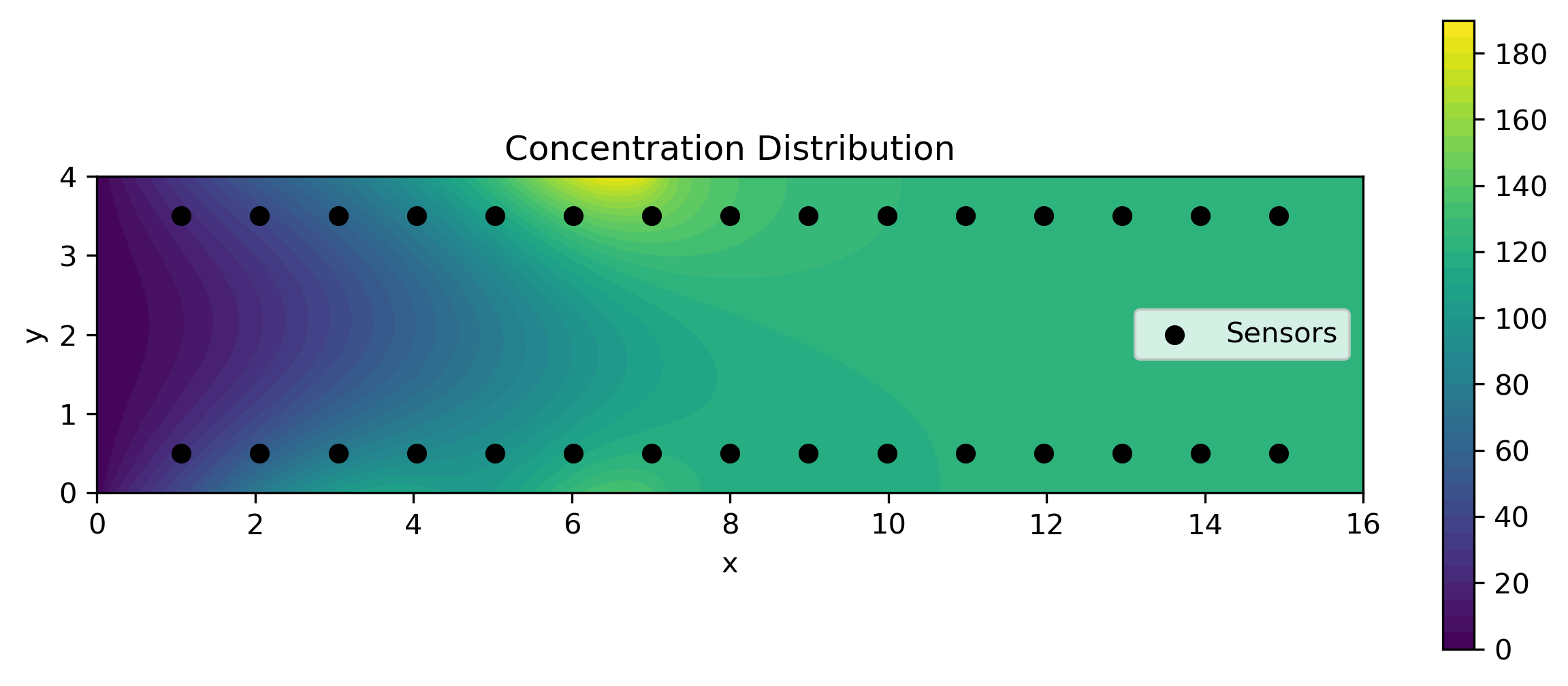}
    \hfill
    \includegraphics[width=0.3\textwidth]{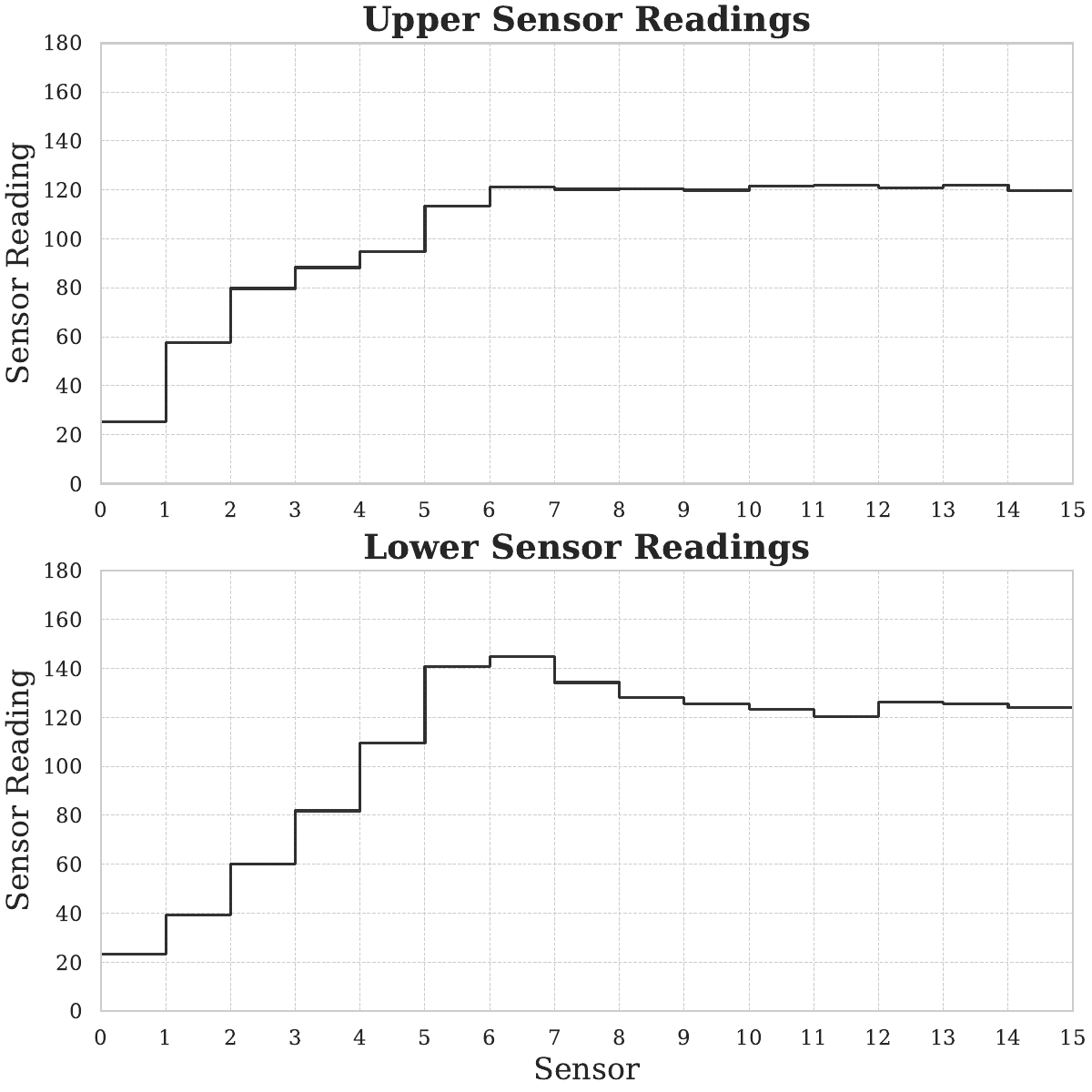}

    \caption{Three realizations of the top and bottom wall flux ($\X$) sampled from the prior (first column), corresponding concentration fields obtained after solving \Cref{eq:advection_diffusion}, and corresponding measurements ($\Y$) at various sensor locations (third column)} 
    \label{fig:dataset_description}
\end{figure}

In \Cref{fig:dataset_description}, we plot three instances of $\X$, the corresponding concentration field obtained by solving \Cref{eq:advection_diffusion}, and the corresponding noisy measurement. The forward model \Cref{eq:advection_diffusion} was solved using the FEniCS \cite{Fenics2012} finite element code over a 850$\times$200 grid containing 340,000 P1 elements. It was ensured that the solution was mesh-converged, and, therefore, a good approximation of the true solution.

For a given level of measurement noise, 9,000 realizations of $\X$ and $\Y$ were used to train the score networks for the variance exploding and preserving formulations. A test set of 1000 instances of $\X$ and $\Y$ was used to evaluate the performance of the model. For each measurement from the test set, 1000 samples of the inferred flux (that belong to the posterior distribution) were generated. Sampling was performed using both the ODE and the SDE sampler. The difference between the two sampling methods was minimal and, therefore, only results for the ODE sampler were reported. 

\begin{figure}[t]
    \centering
    \includegraphics[width=0.3\textwidth]{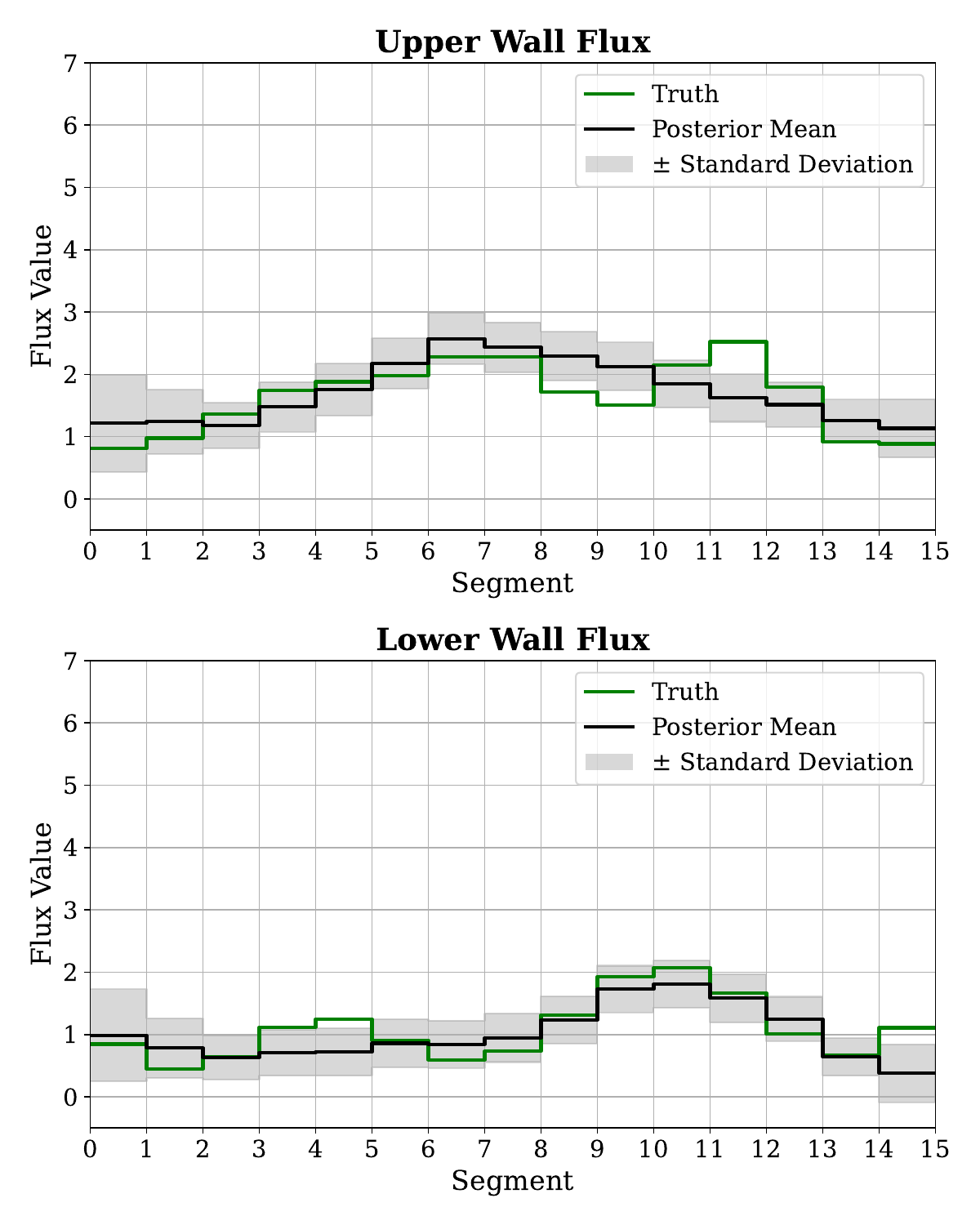}
    \hfill
    \includegraphics[width=0.3\textwidth]{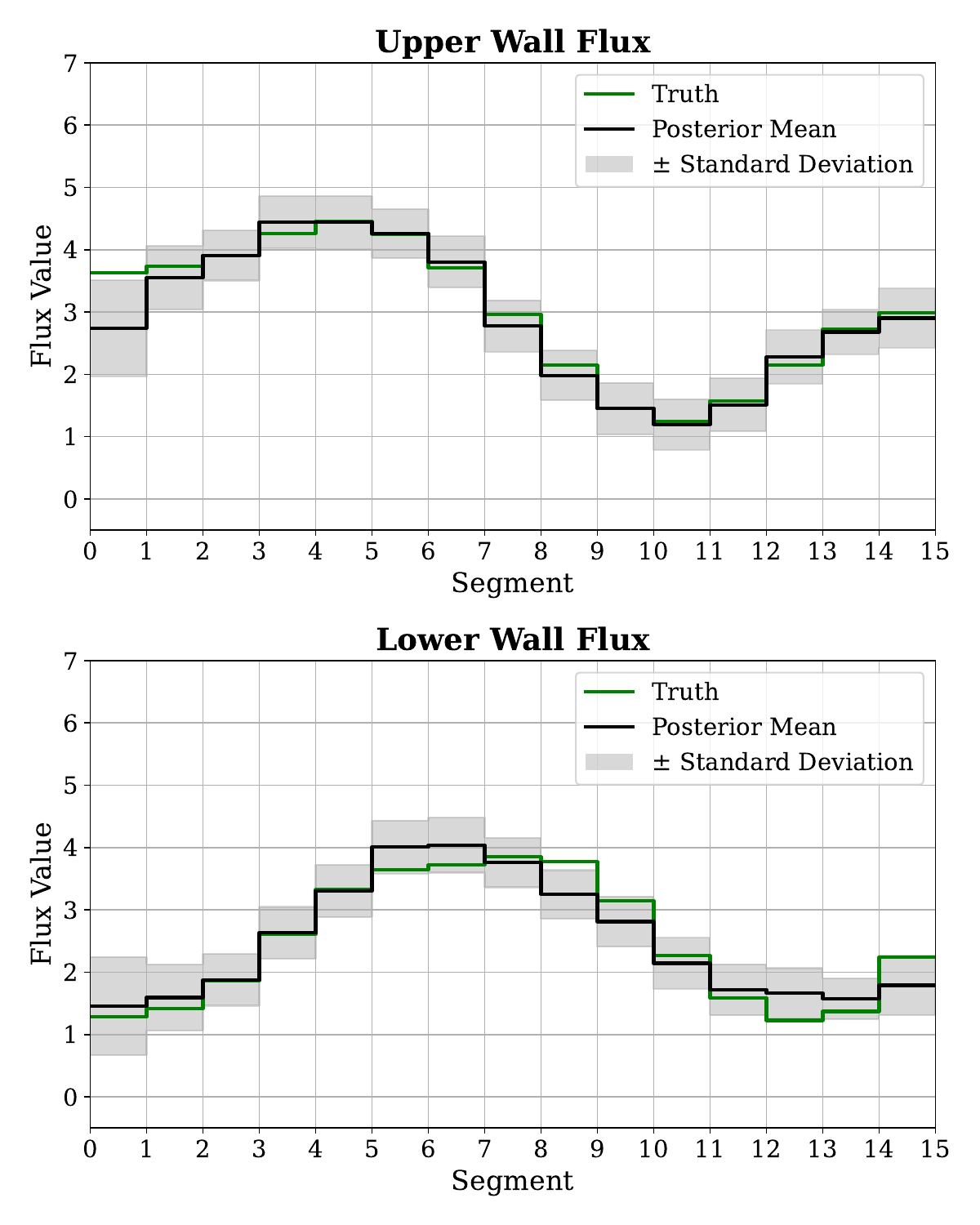}
    \hfill
    \includegraphics[width=0.3\textwidth]{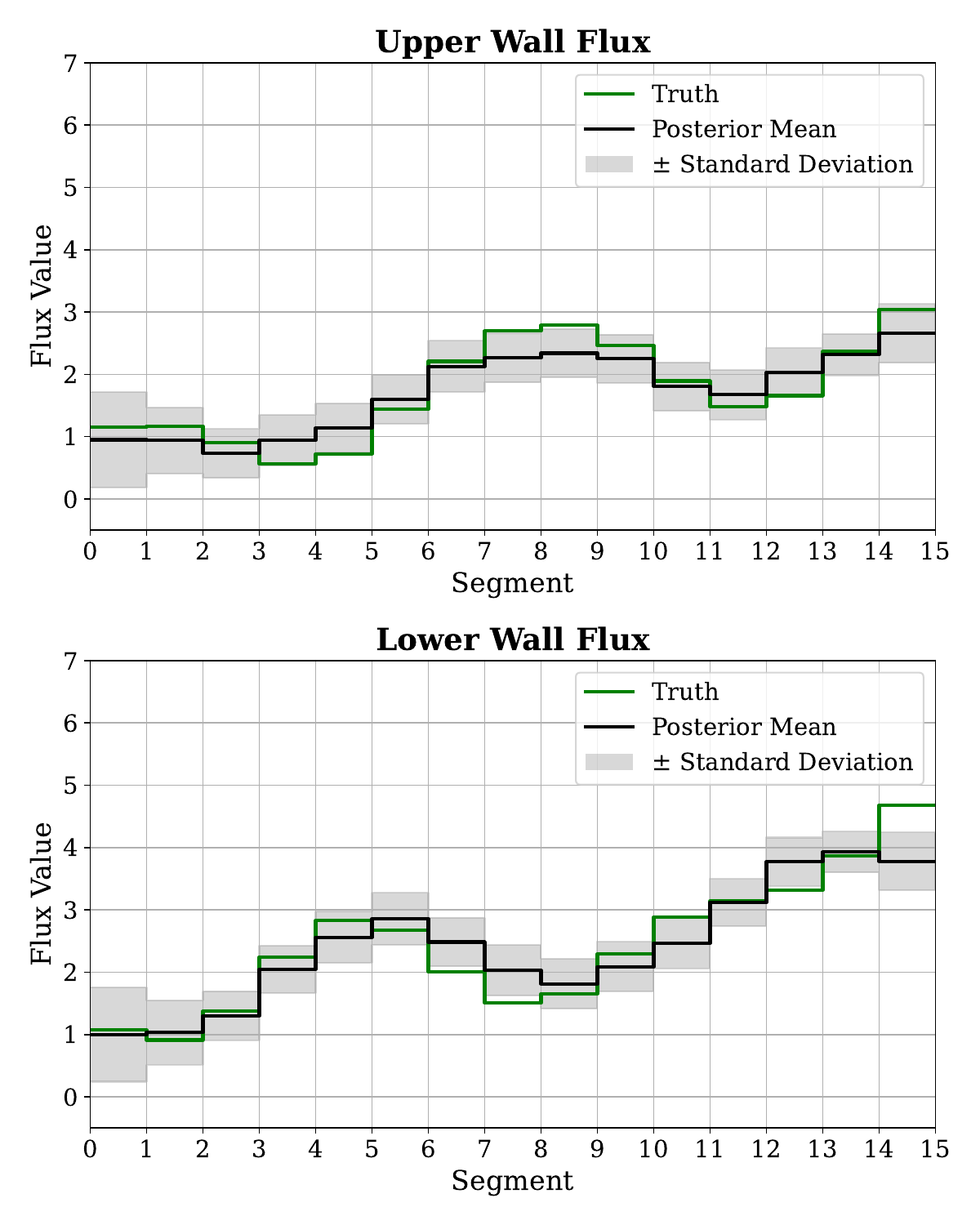}
    \caption{Posterior mean of boundary flux (black line) estimated using the variance exploding formulation with the ODE sampler for each segment in the upper and lower walls for three different measurements in the test dataset, corresponding true flux values (green line), and one standard deviation range around the posterior mean (gray shade). Each column corresponds to a different test case}
    \label{fig:VE_mean_vs_true}
\end{figure}

In \Cref{fig:VE_mean_vs_true}, we present results for the variance exploding formulation with measurement noise $\sigma_{\epsilon} = 0.02$ for three test cases. For each case, we plot the true flux distribution, the posterior mean estimated by the conditional diffusion model (considered to be the best guess), and the one standard deviation range about the mean. In each case, we observe that the posterior mean is close to the true value, and that in most cases the true value is contained within the one standard deviation range of the mean.

\begin{figure}[htbp]
    \centering
    \includegraphics[width=\linewidth]{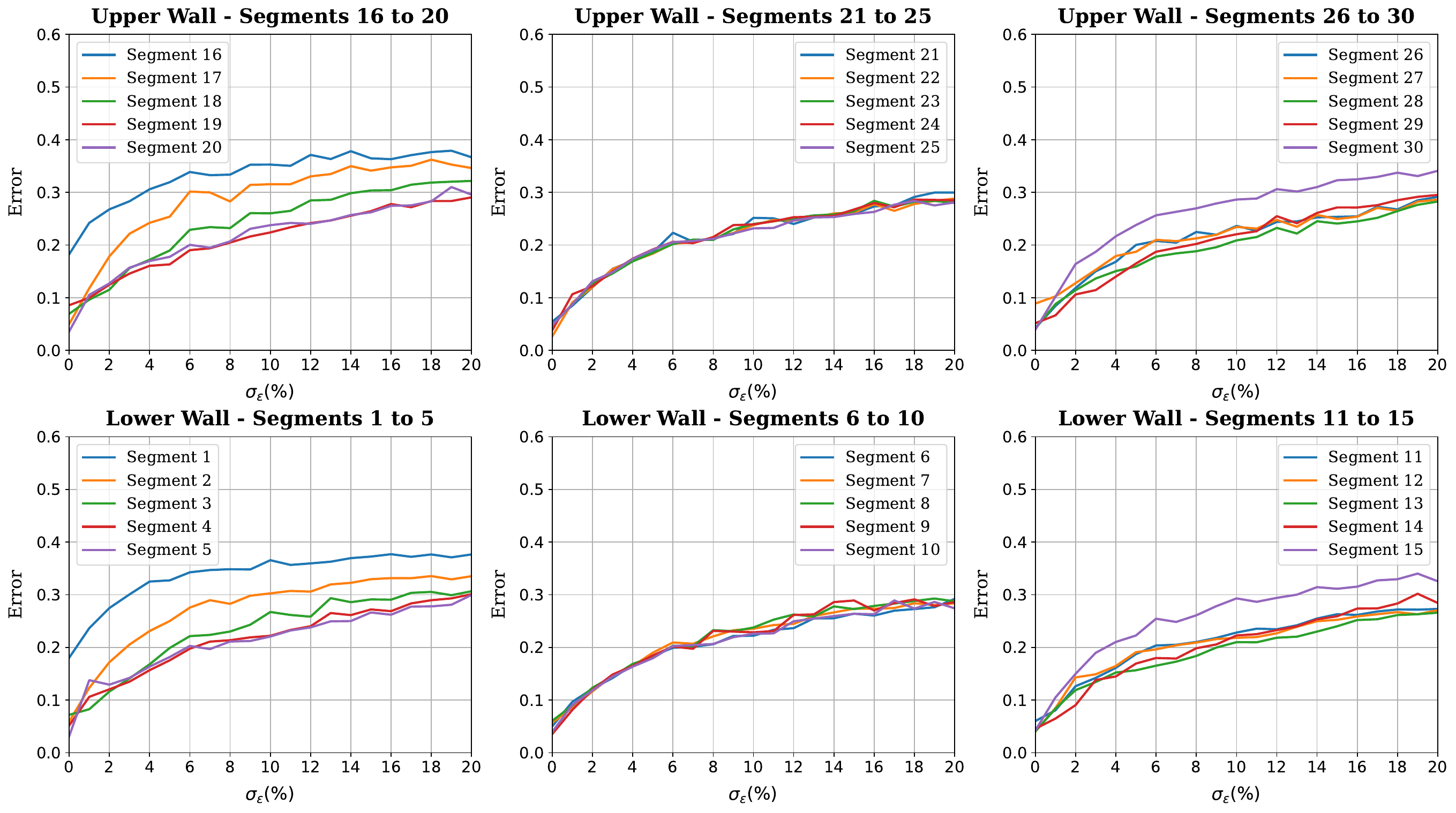}
    \caption{Sample-averaged absolute error of the generated posterior mean flux at each segment of the lower and the upper wall, normalized by the average value of flux, for all segments and all samples in the test set, for different levels of measurement noise. These results were obtained using the variance exploding formulation and ODE sampler}
    \label{fig:VE_error_vs_noise}
    \includegraphics[width=\linewidth]{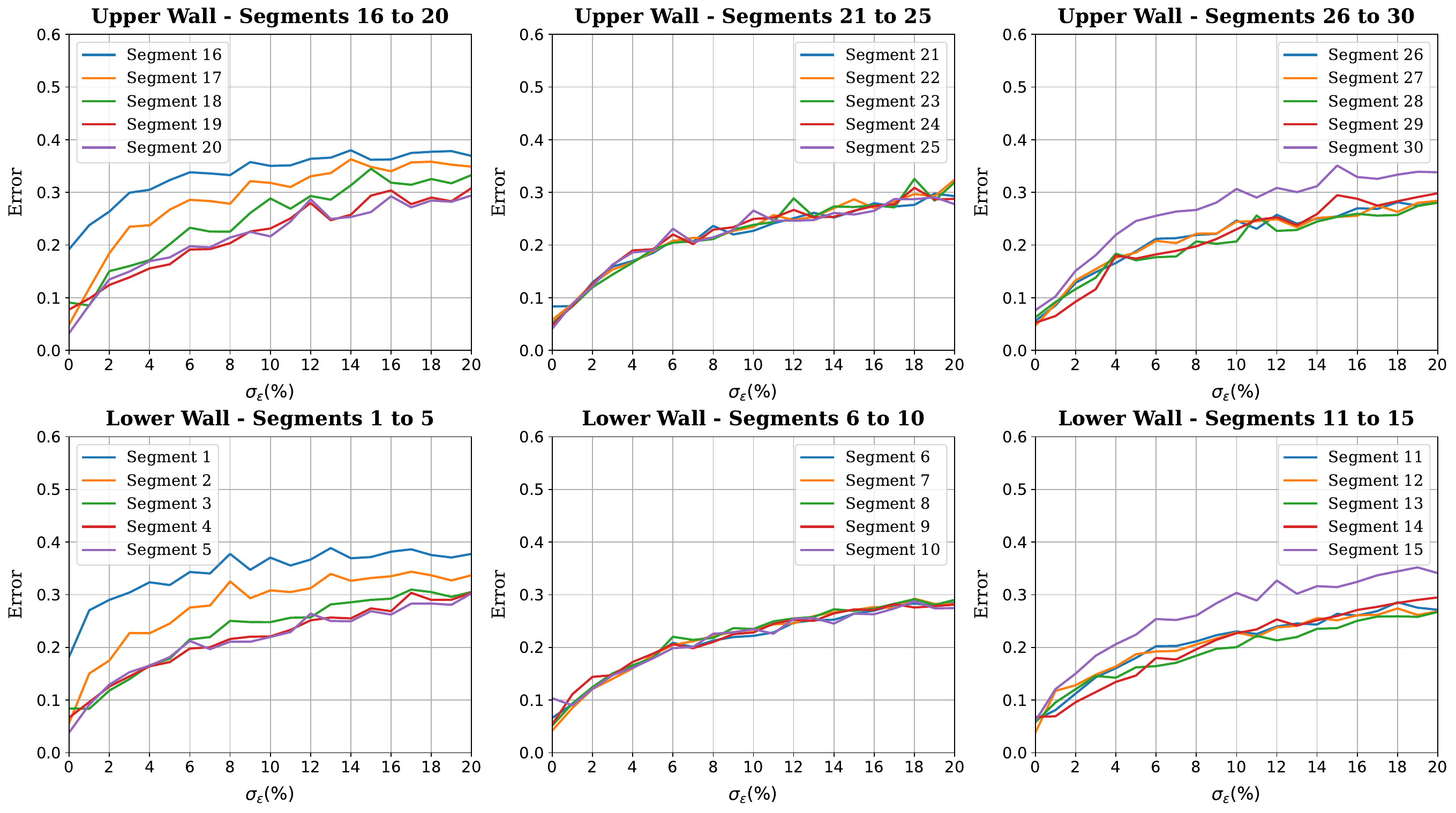}
    \caption{Sample-averaged absolute error of the generated posterior mean flux at each segment of the lower and the upper wall, normalized by the average value of flux, for all segments and all samples in the test set, for different levels of measurement noise. These results were obtained using the variance preserving formulation and ODE sampler}
    \label{fig:VP_error_vs_noise}
\end{figure}

A quantitative measure of the difference between the predicted mean and the true flux is shown in \Cref{fig:VE_error_vs_noise}. Here, for each segment, we have plotted the average of the absolute value of the difference between the mean predicted flux and the true flux. The average is taken across all test samples. Further, this value is normalized by the average value of flux for all segments and all samples in the test set. Thus, a value of 0.1 for the error implies a discrepancy of around 10\% between the predicted mean and the true value of the flux. Several interesting trends are observed in this plot.  

First, for every segment, the error increases with increasing standard deviation in measurement noise. This is to be expected, since for the same signal magnitude, increasing the magnitude of noise makes the measurements less informative and thereby increases the prediction error. Second, even with zero measurement noise, there is significant error in the prediction (around 5.9\%). This is attributed to the fact that the measurements are sparse (only 30 points in the entire domain) and the inverse problem is likely ill-posed even in the absence of noise. Third, the error in predicting the flux for the first segment (both at the bottom and the top) is larger than that for the other segments. This can be explained by recognizing that all flux segments are informed by both upstream and downstream measurements (see \Cref{fig:adv-diff-setup}), except for the first segment on the upper and lower walls, which are informed only by downstream measurements. This is the likely cause for larger errors in these segments. Finally, we note that the error appears to saturate with increasing noise. This is because with increasing noise the measurements cease to be informative and the posterior distribution reverts to the prior distribution. In \Cref{fig:VP_error_vs_noise}, we plot similar results for the variance preserving formulation. By comparing this figure with \Cref{fig:VE_error_vs_noise}, we note that the two formulations incur similar errors.

\begin{figure}[htbp]
	\centering
	\includegraphics[width=\linewidth]{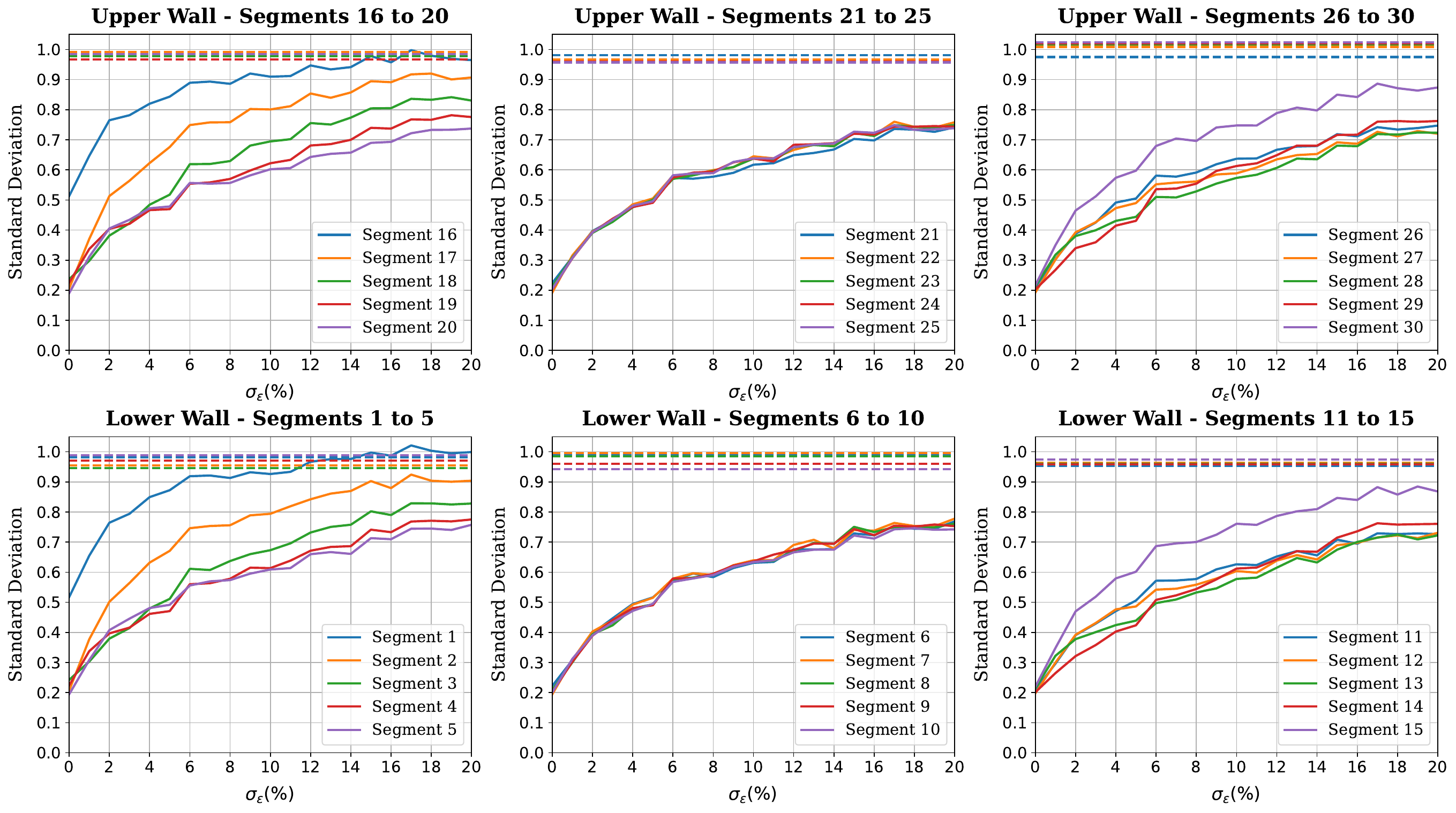}
	\caption{Sample-averaged posterior standard deviation of the flux for each segment of the lower and the upper wall for different levels of measurement noise obtained using the variance exploding formulation and ODE sampler}
	\label{fig:VE_std_vs_noise}
	\centering
	\includegraphics[width=\linewidth]{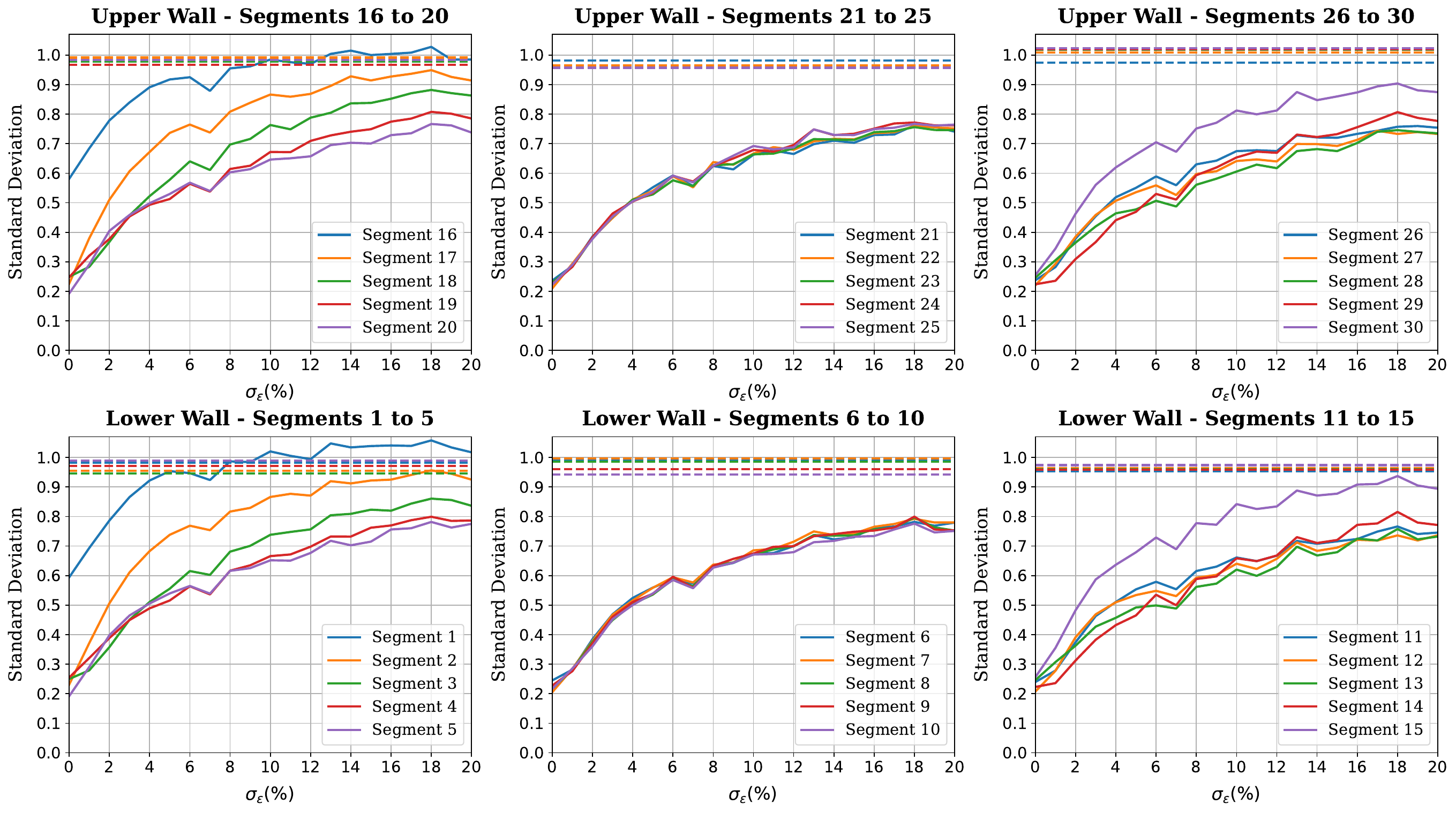}
	\caption{Sample-averaged posterior standard deviation of the flux for each segment of the lower and the upper wall for different levels of measurement noise obtained using the variance preserving formulation and ODE sampler}
	\label{fig:VP_std_vs_noise}
\end{figure}

In \Cref{fig:VE_std_vs_noise,fig:VP_std_vs_noise} we plot the average value of the standard deviation of the predicted flux computed using the variance exploding and preserving formulations, respectively. In these figures, the dashed horizontal lines denote the standard deviation of the flux across all test samples. This value approximates the standard deviation of the prior distribution for the flux for each segment. We observe that with increasing measurement noise, the standard deviation of the posterior distribution increases, and at high levels of noise it approaches the standard deviation of the prior distribution, once again underlining the fact that at these levels the noise in the measurement makes them uninformative and the posterior distribution defaults to the prior distribution.

\subsubsection{Multiple measurement operators}

Next, we demonstrate how a single diffusion model can be trained to handle multiple measurement operators. The theoretical development for this is described in \Cref{sec:stochastic-obs}. We parameterize the distribution of measurement operators by a $\Ny$-dimensional binary variable $\Mm$ defined as follows
\begin{equation}
    M_i = \left\{ \begin{array}{cc} 1 & \mbox{if the \supth{i} sensor is on} \\ 0 & \mbox{if the \supth{i} sensor is off.} \end{array} \right.
\end{equation}
For the marginal distribution of $\Mm$ we select each $M_i$ to be an independent Bernoulli variable with $p = 0.7$, that is, the probability of any sensor being on is 70\%. 

As described in \Cref{sec:stochastic-obs}, we generate the training data by sampling $\Mm$ and $\X$ from their respective priors, and then solving \Cref{eq:advection_diffusion} to determine the concentration field. Thereafter, we sample the concentration field at the locations where the sensor is on and set the measurement to the concentration plus an additive Gaussian noise. For locations where the sensor is off, we set the concentrations to -1.

\begin{figure}[htbp]
    \centering

    \includegraphics[width=0.3\textwidth]{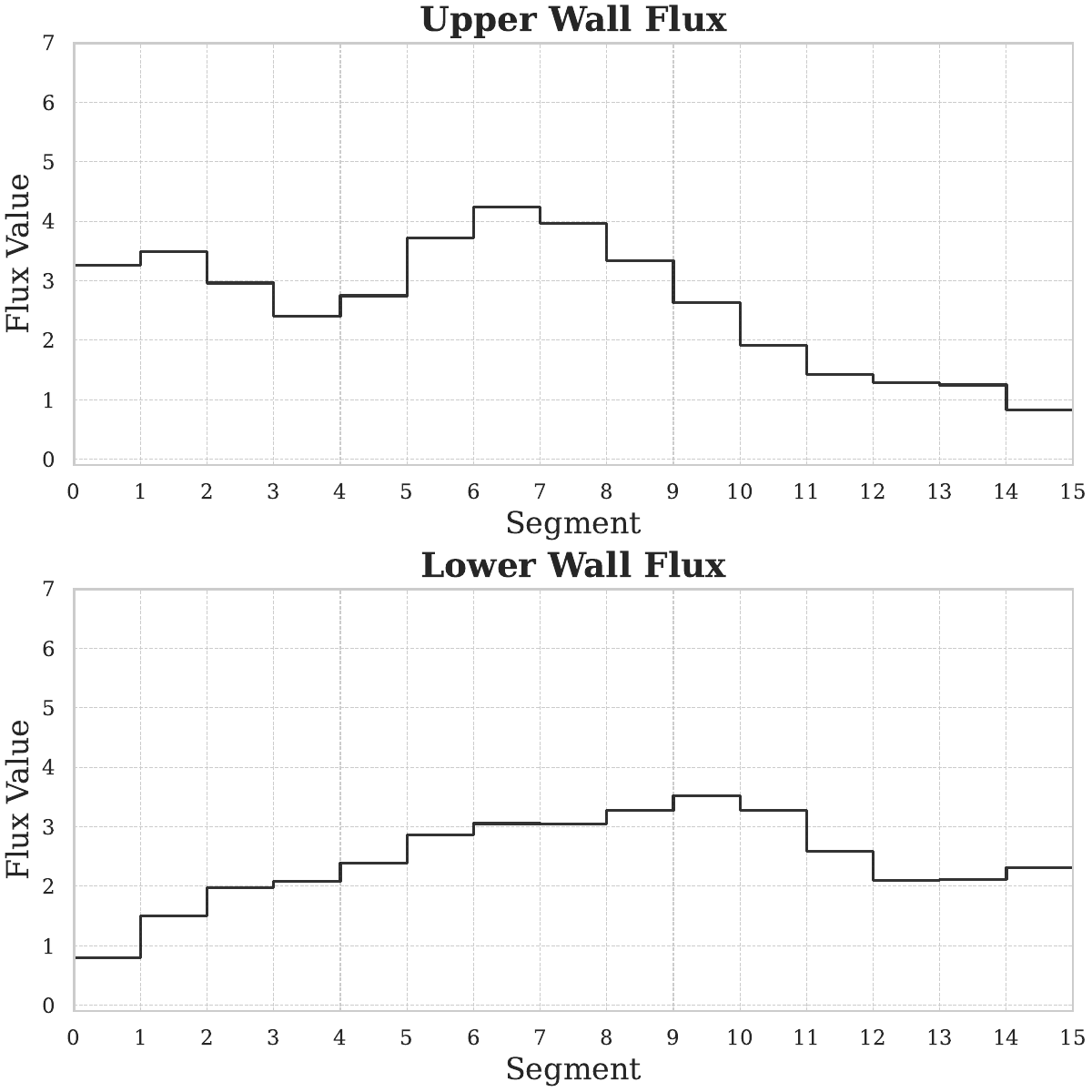}
    \hfill
    \includegraphics[width=0.3\textwidth]{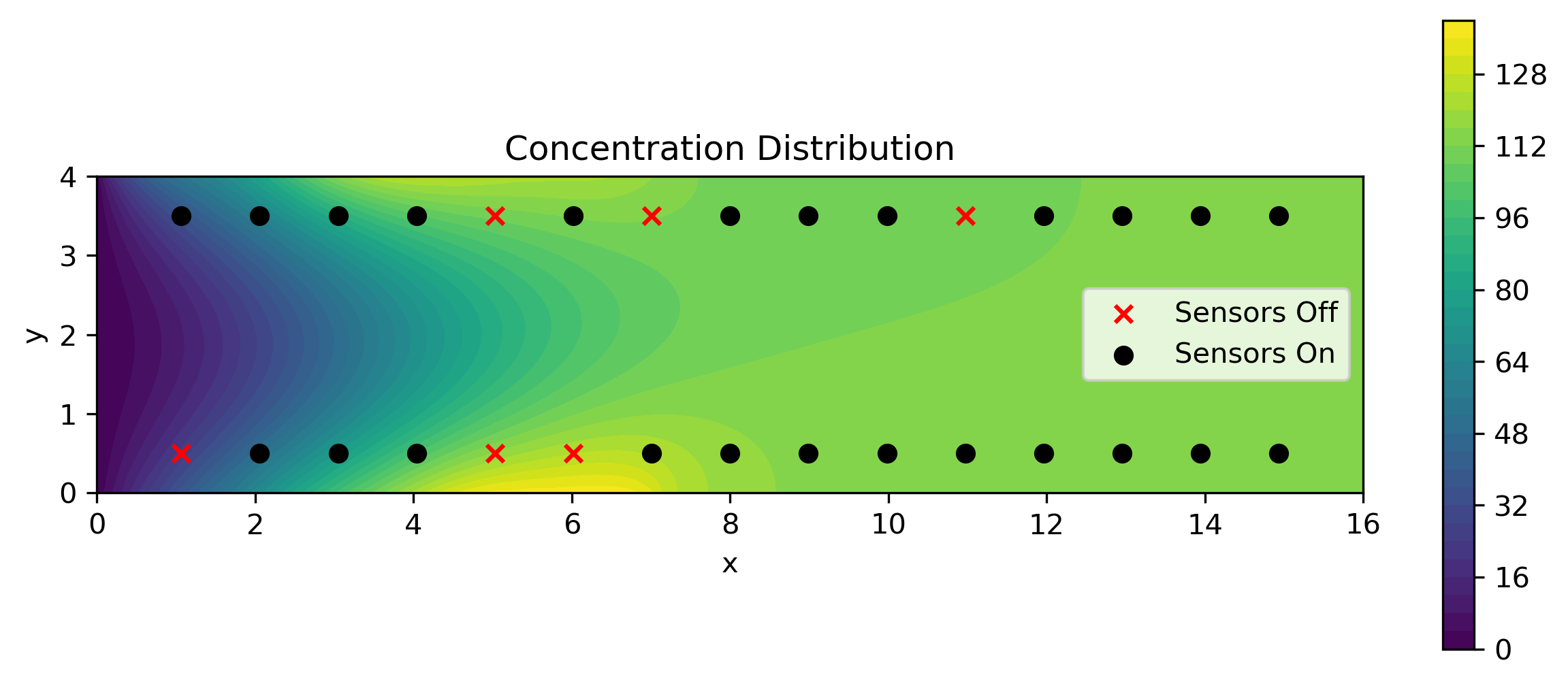}
    \hfill
    \includegraphics[width=0.3\textwidth]{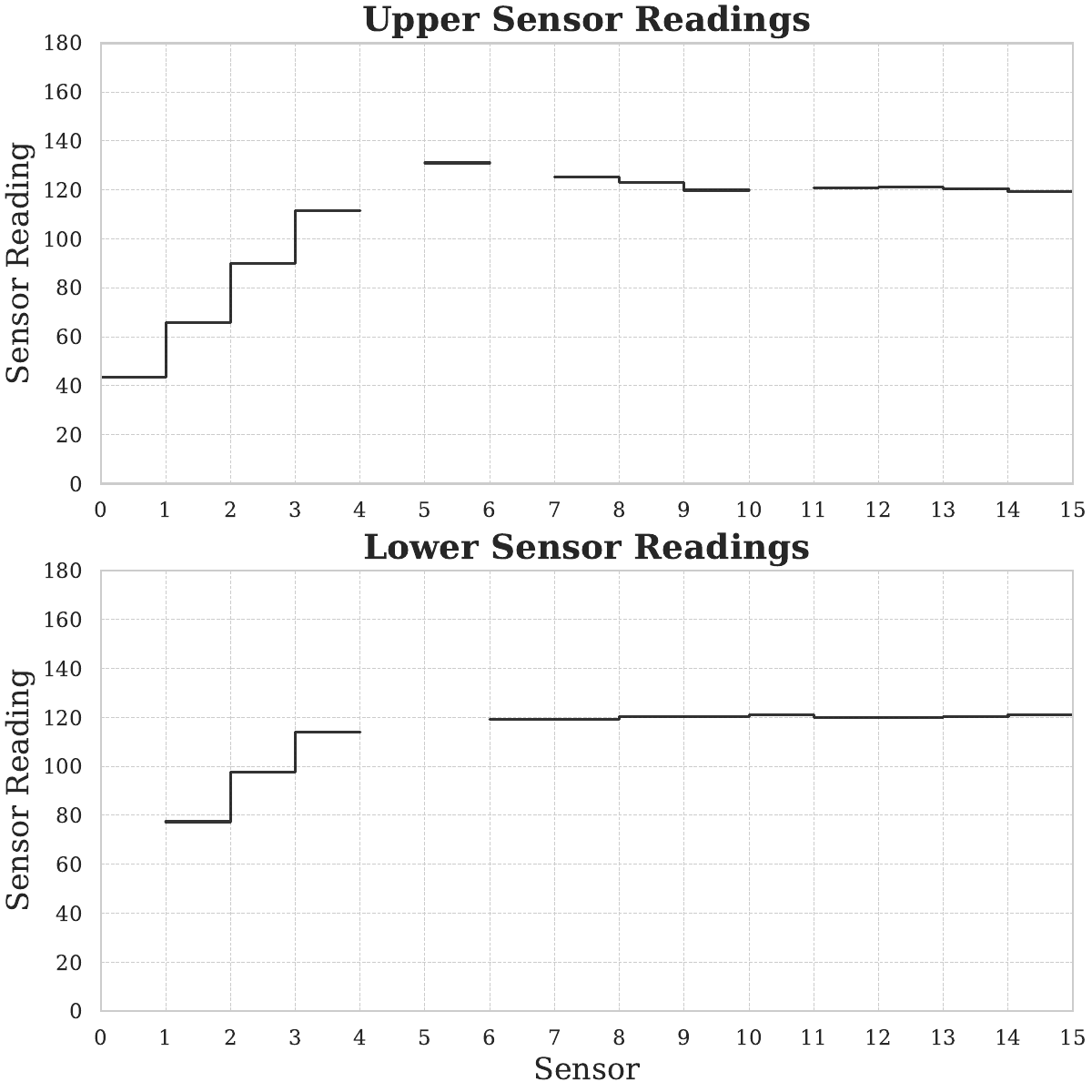}

    \vspace{0.5cm}

    \includegraphics[width=0.3\textwidth]{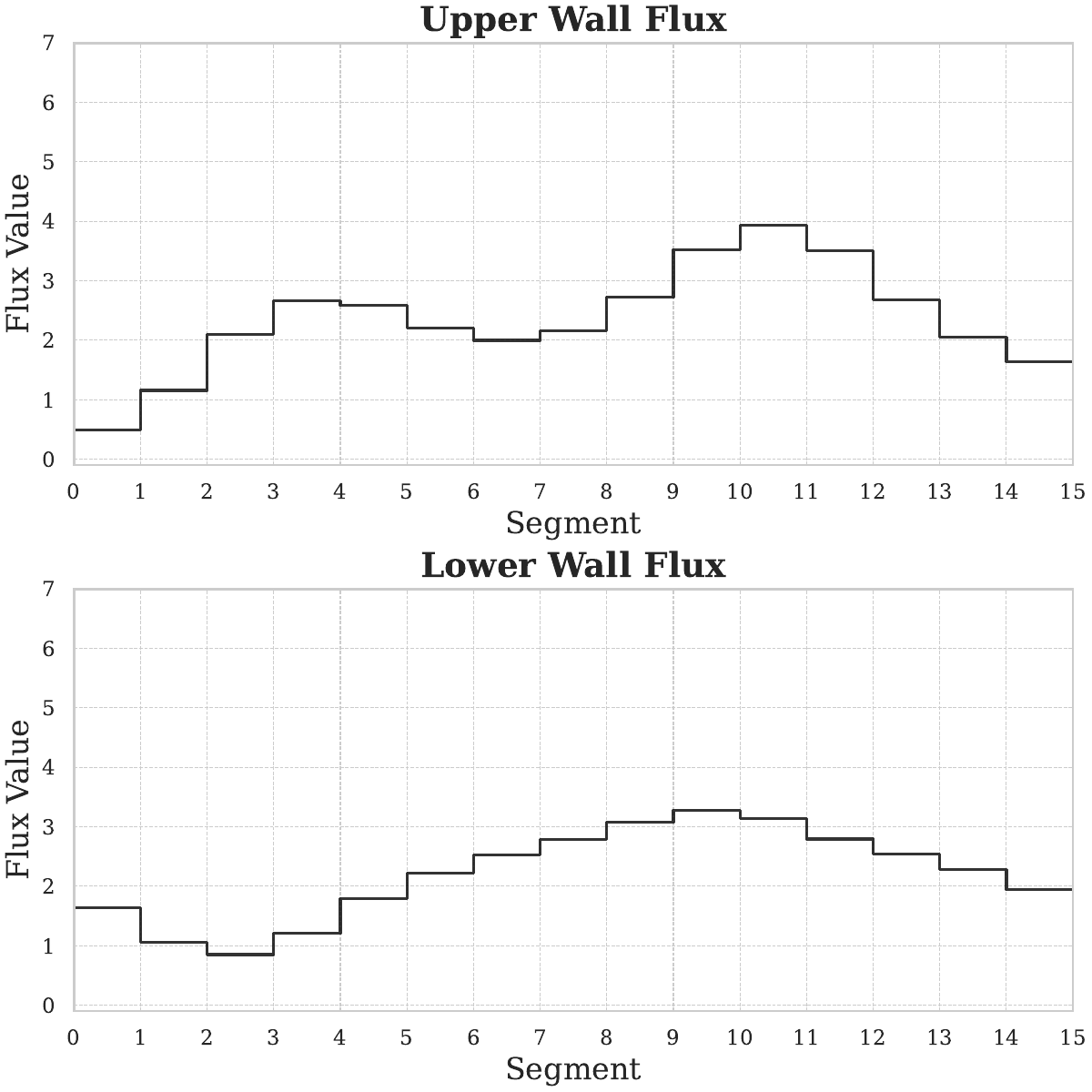}
    \hfill
    \includegraphics[width=0.3\textwidth]{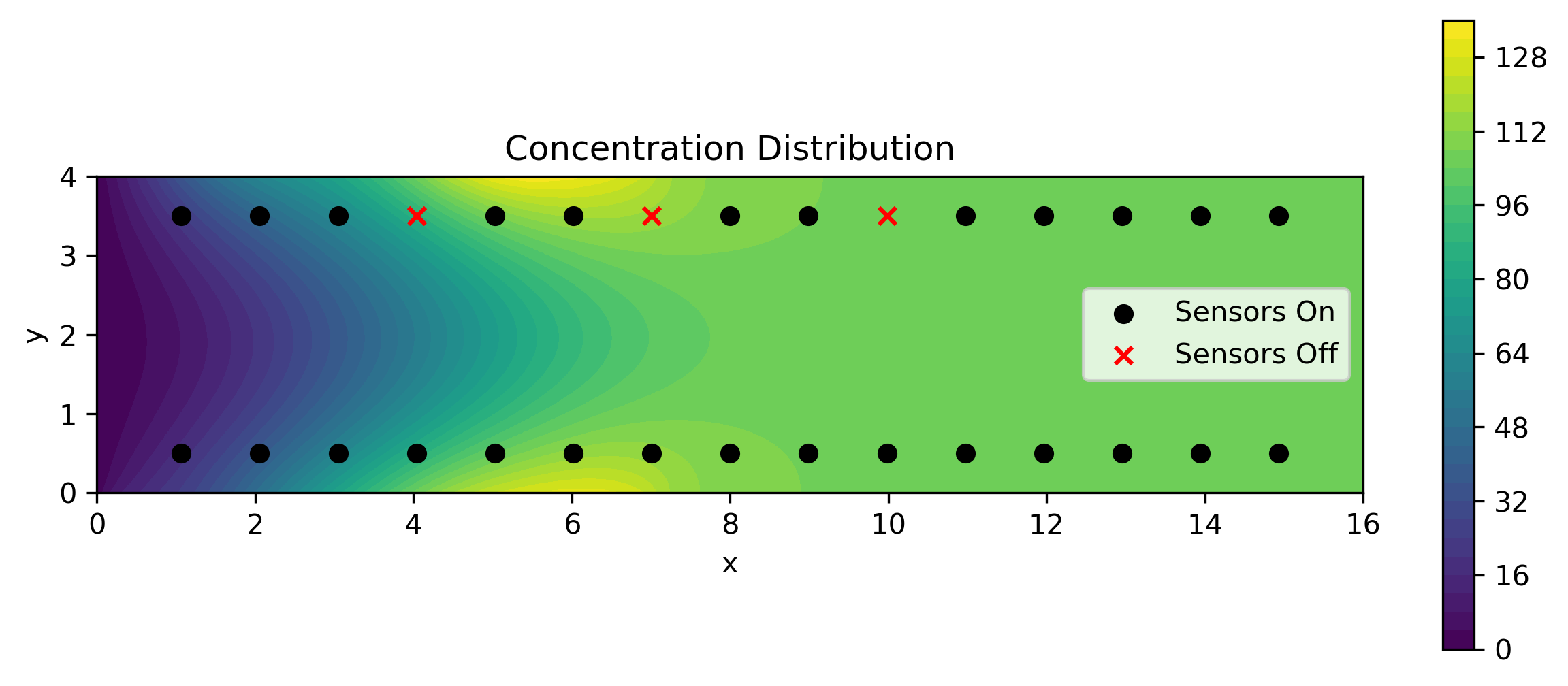}
    \hfill
    \includegraphics[width=0.3\textwidth]{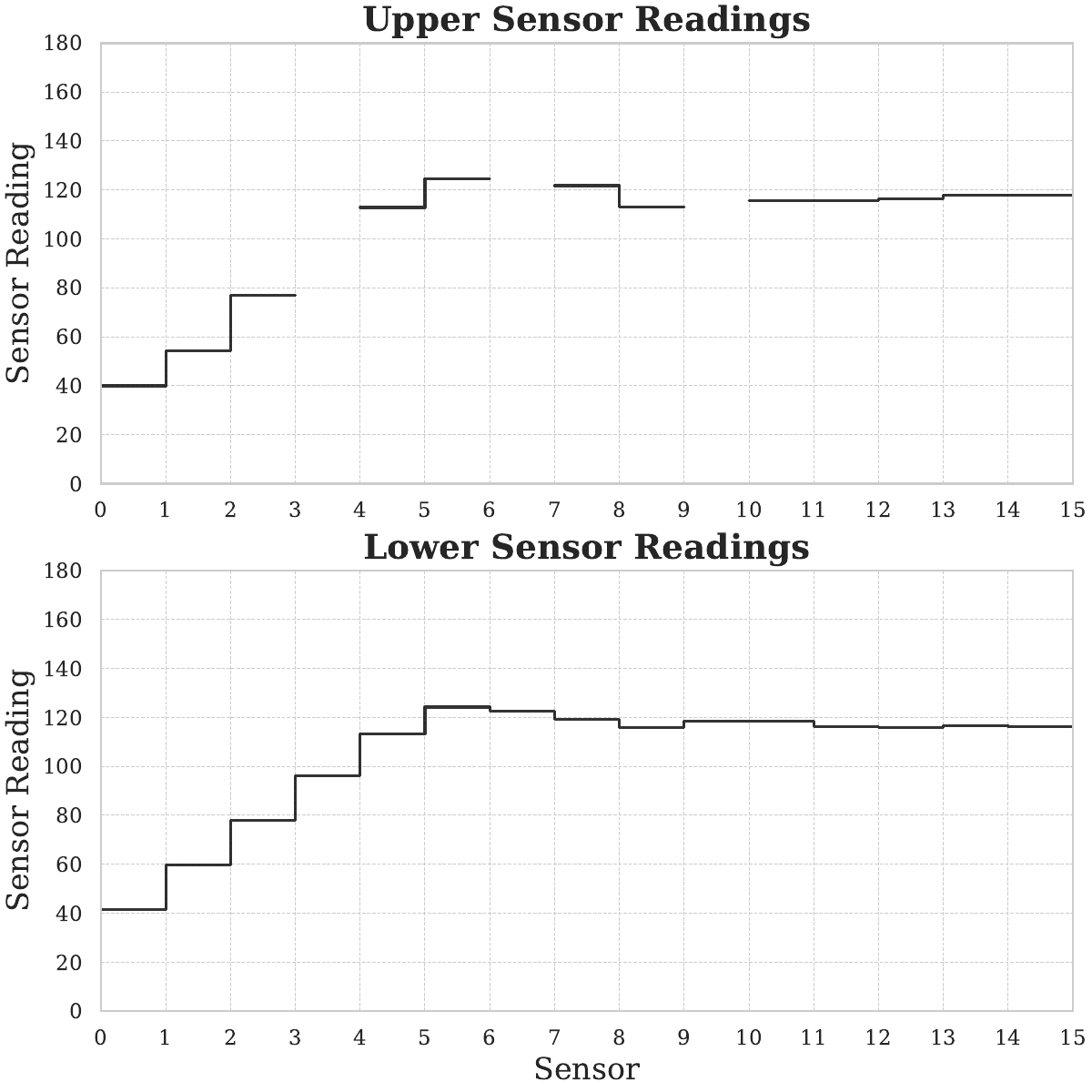}

    \vspace{0.5cm}

    \includegraphics[width=0.3\textwidth]{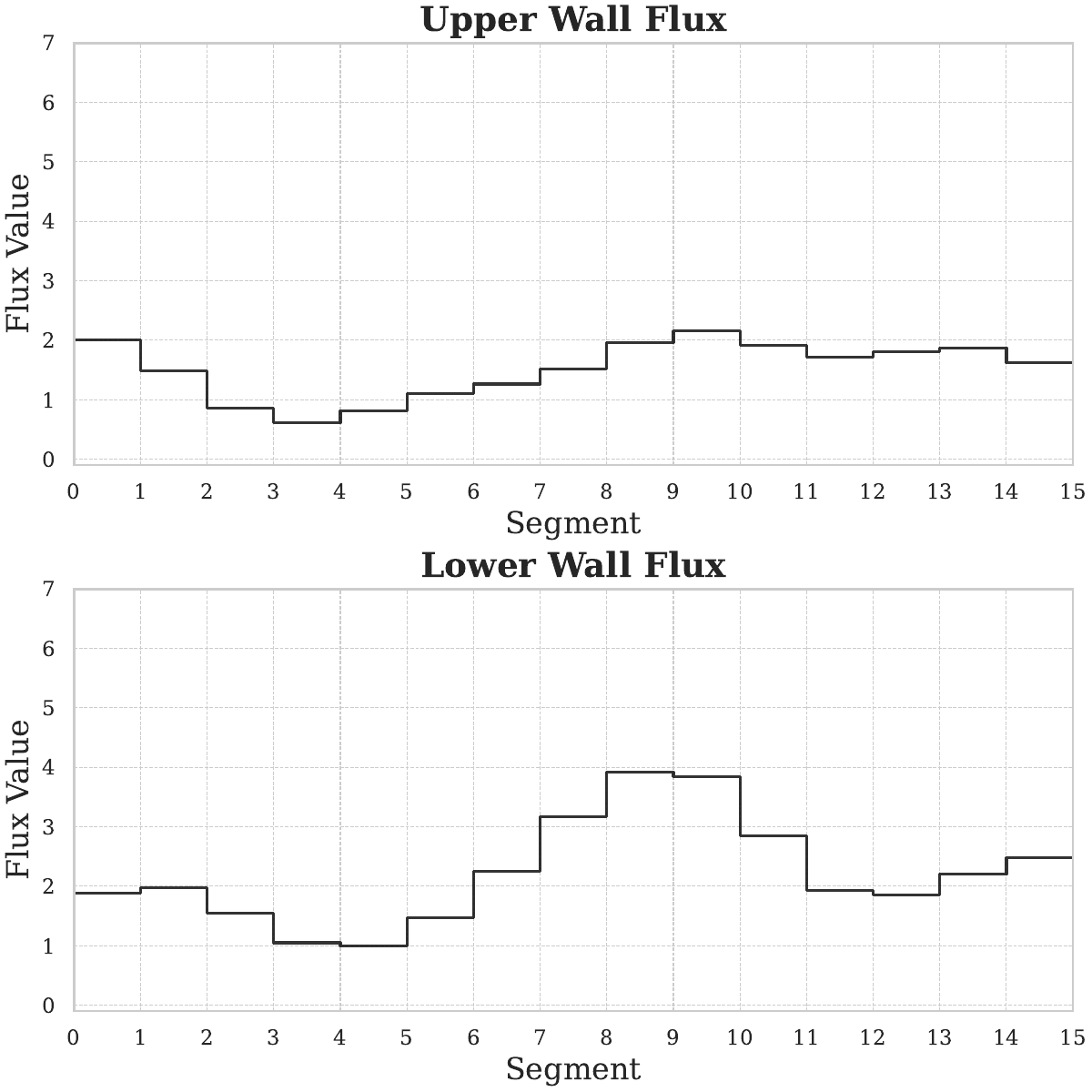}
    \hfill
    \includegraphics[width=0.3\textwidth]{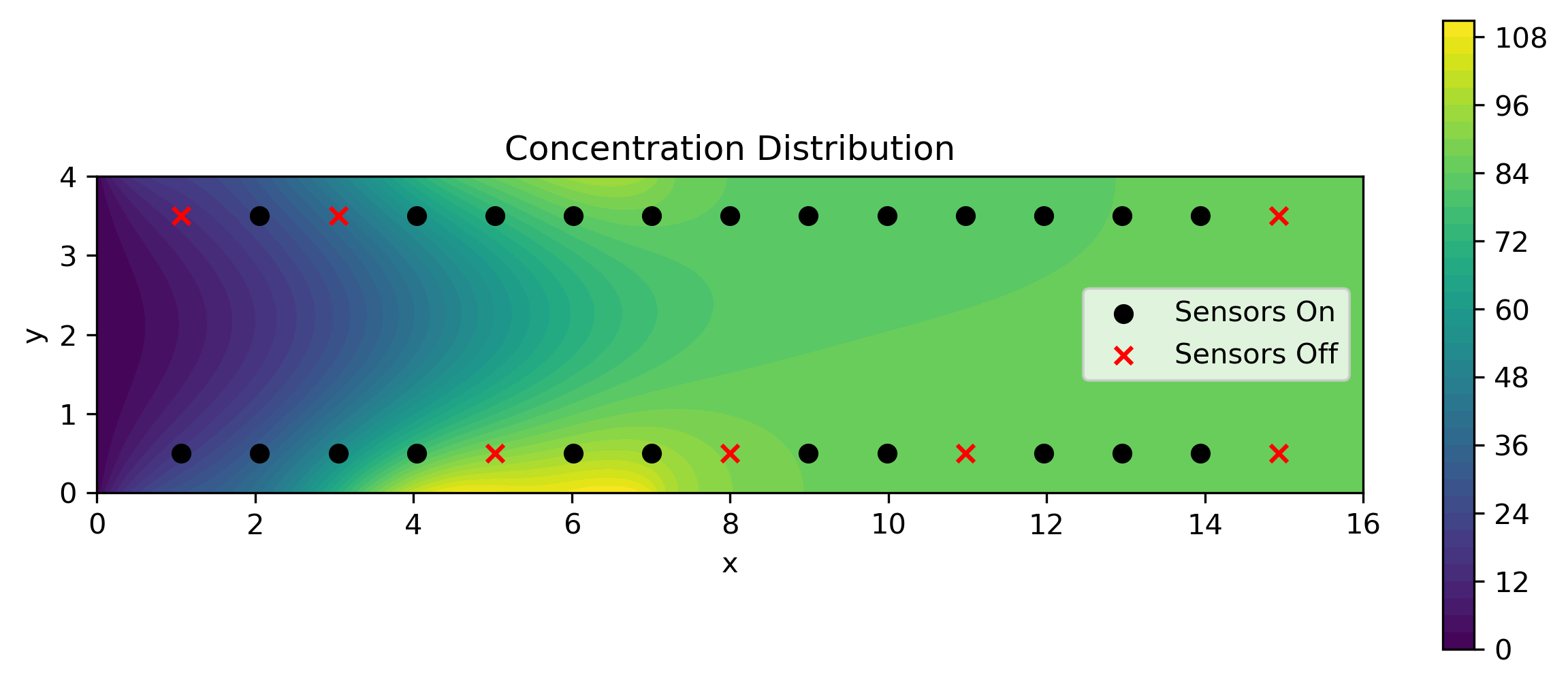}
    \hfill
    \includegraphics[width=0.3\textwidth]{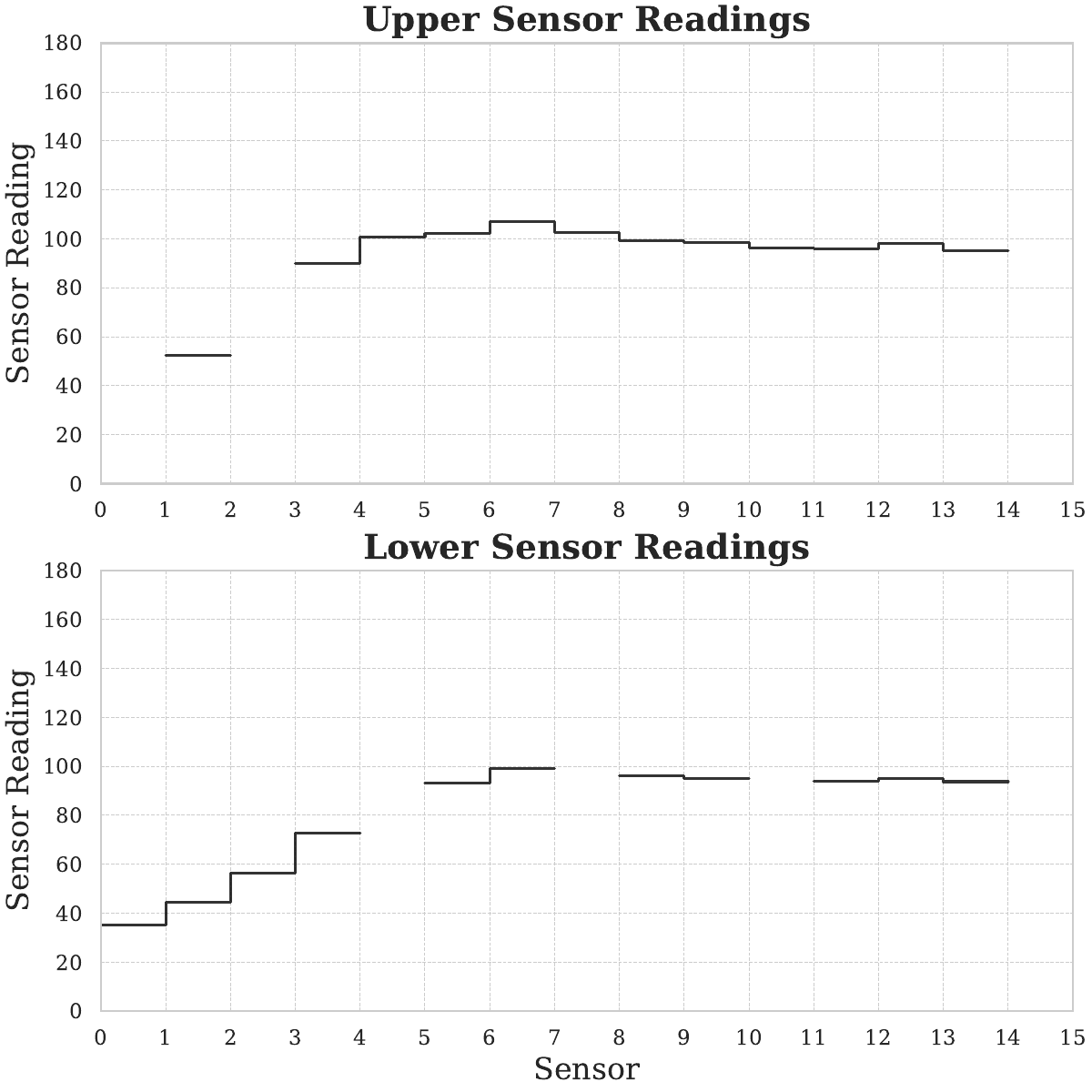}

    \caption{Three realizations of the top and bottom wall flux ($\X$) sampled from the prior (first column), corresponding concentration fields obtained after solving \Cref{eq:advection_diffusion}, and corresponding concentration measurements ($\Y$) at various sensor locations (third column) when sensors are ON with 70\% probability. Red crosses (\textcolor{red}{$\times$}) are used to denote sensors that are OFF, which are also reflected by gaps in the concentration profiles} \label{fig:mask_dataset_description}
\end{figure}

Using the training dataset, which consists of 36,000 realizations from the joint distribution of $\X$, $\Y$ and $\Mm$,  we train a score network whose input includes $\x$, $\y$, $\mm$ and $t$. We do this for both the variance exploding and preserving formulations. Three  instances of the training data are shown in \Cref{fig:mask_dataset_description}. In this figure, the sensors that are turned off are denoted by a red cross in the plot of the concentration's spatial distribution. In this example also, we find that the results for the two formulations are similar. Therefore, we only show results for the variance exploding formulation. 

\begin{figure}[htbp]
    \centering

    \includegraphics[width=0.3\textwidth]{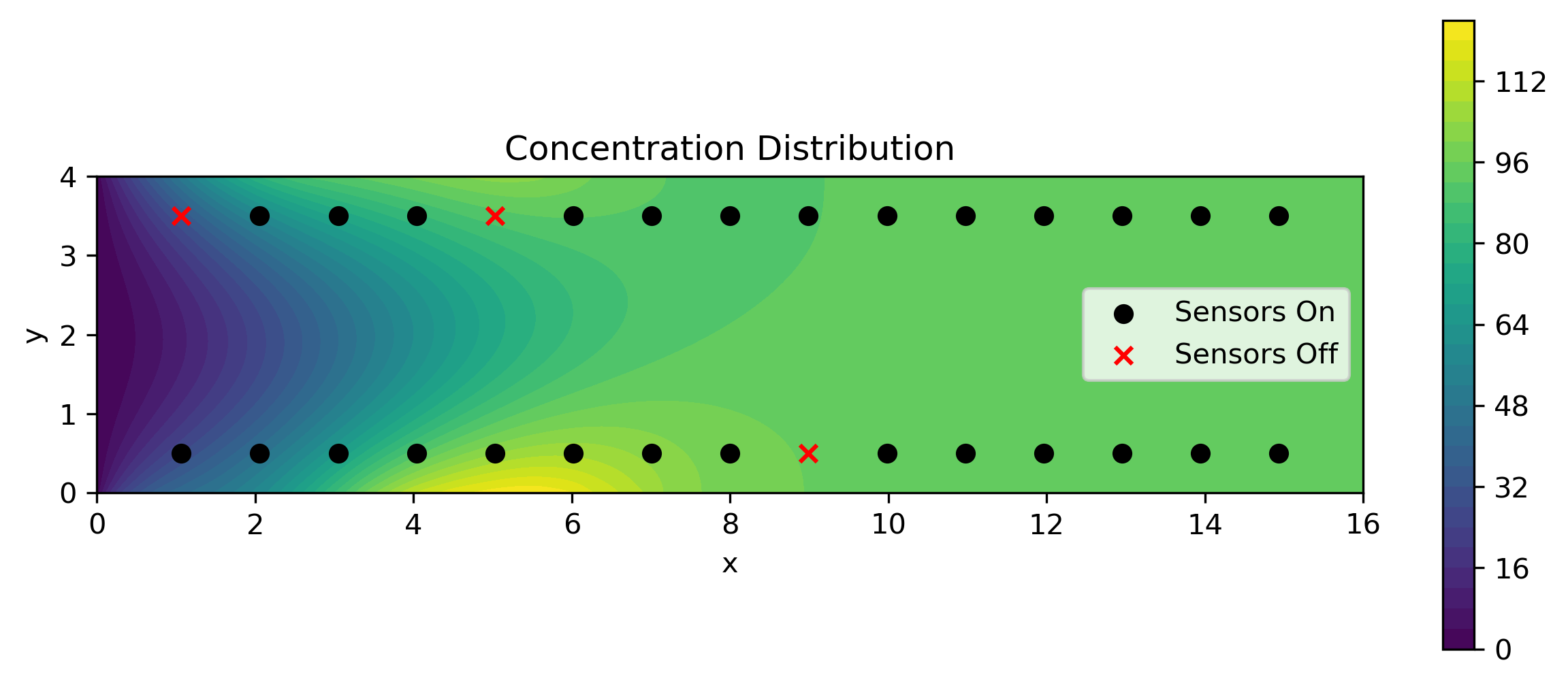}
    \hfill
    \includegraphics[width=0.3\textwidth]{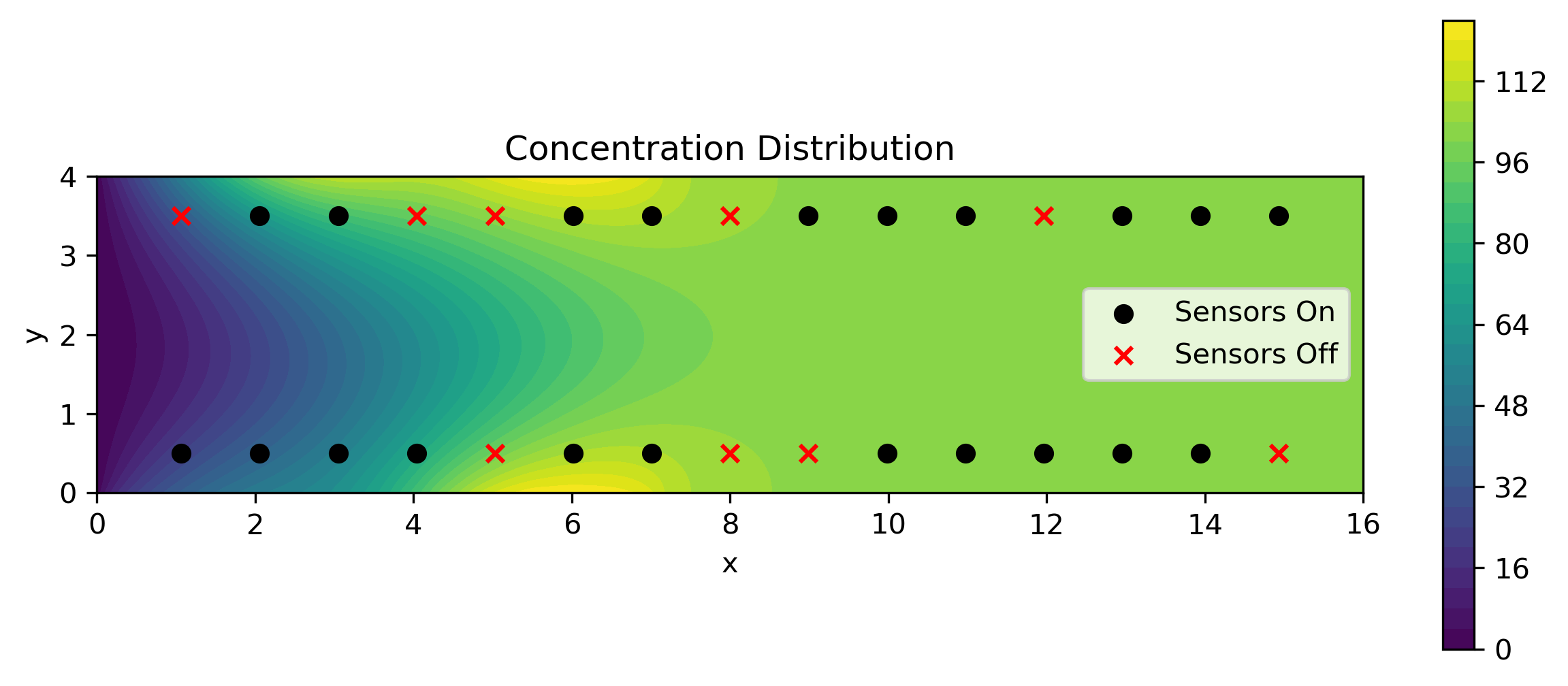}
    \hfill
    \includegraphics[width=0.3\textwidth]{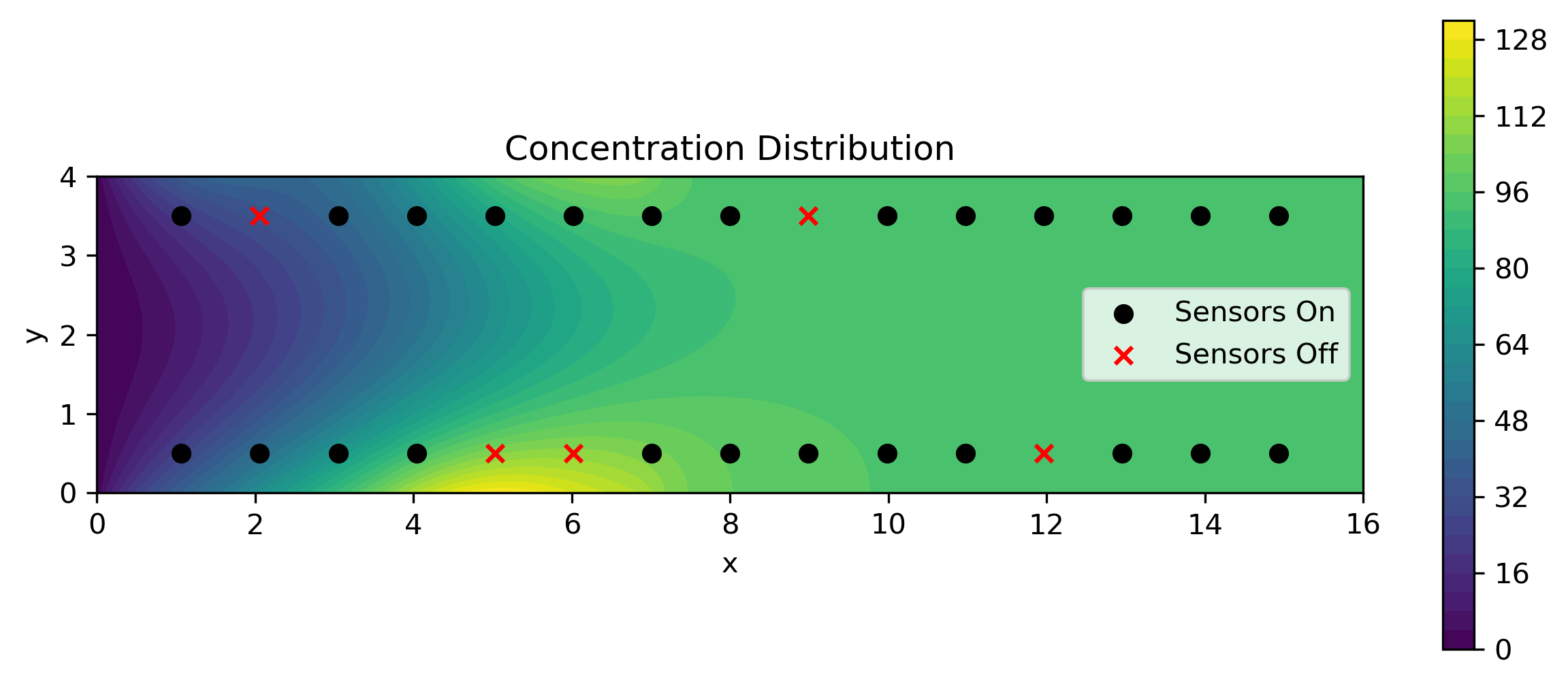}
    \vspace{0.5cm}

    \includegraphics[width=0.3\textwidth]{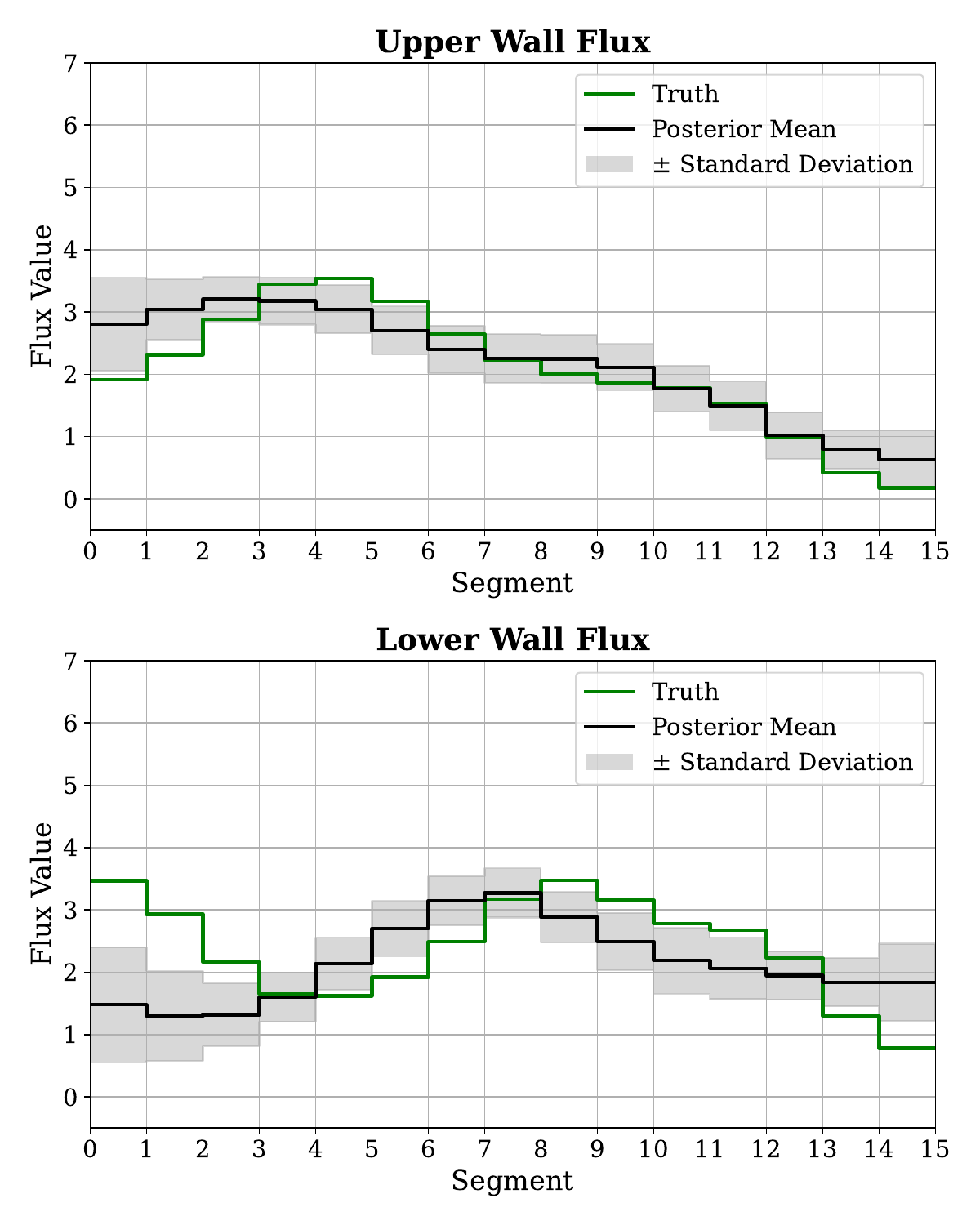}
    \hfill
    \includegraphics[width=0.3\textwidth]{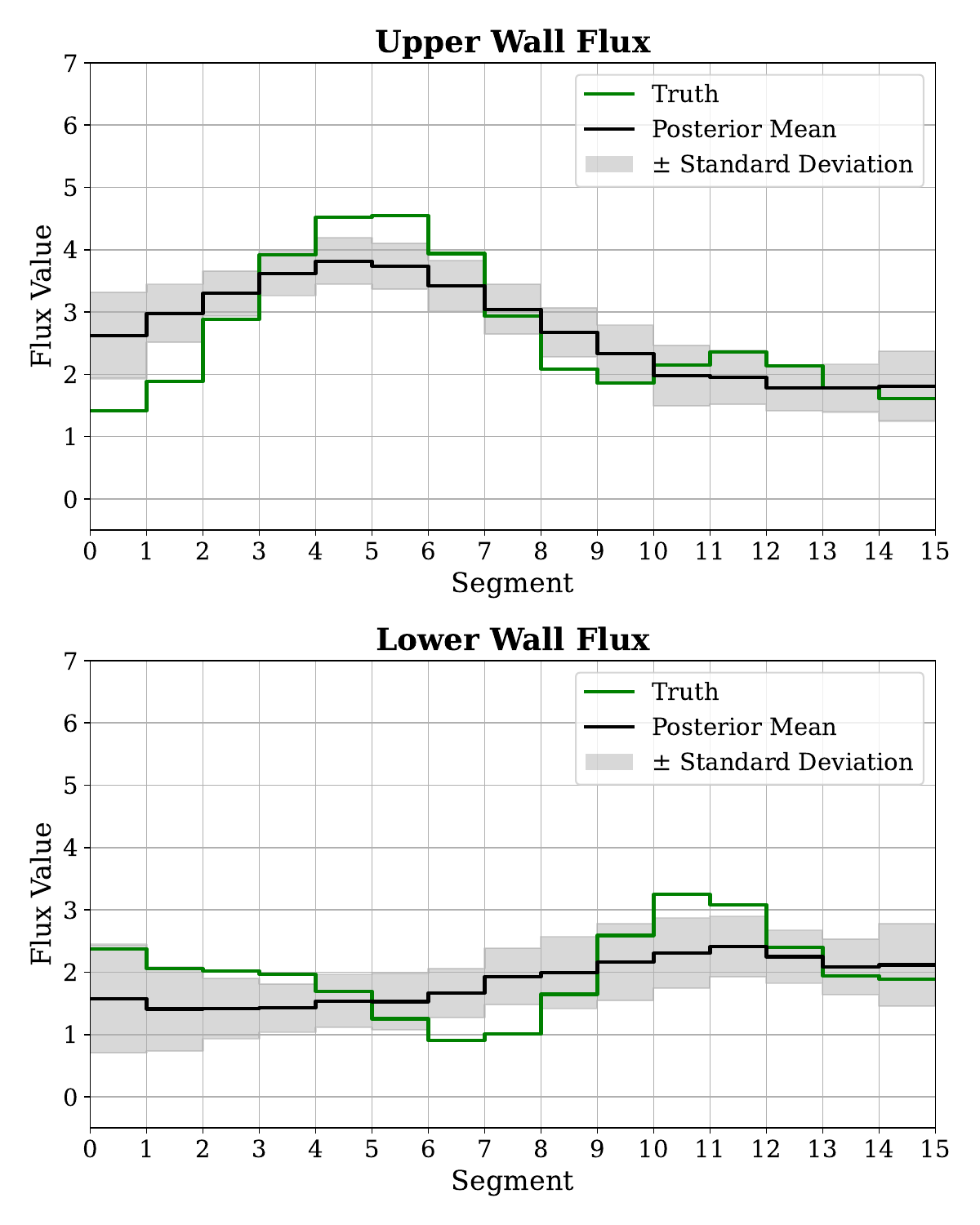}
    \hfill
    \includegraphics[width=0.3\textwidth]{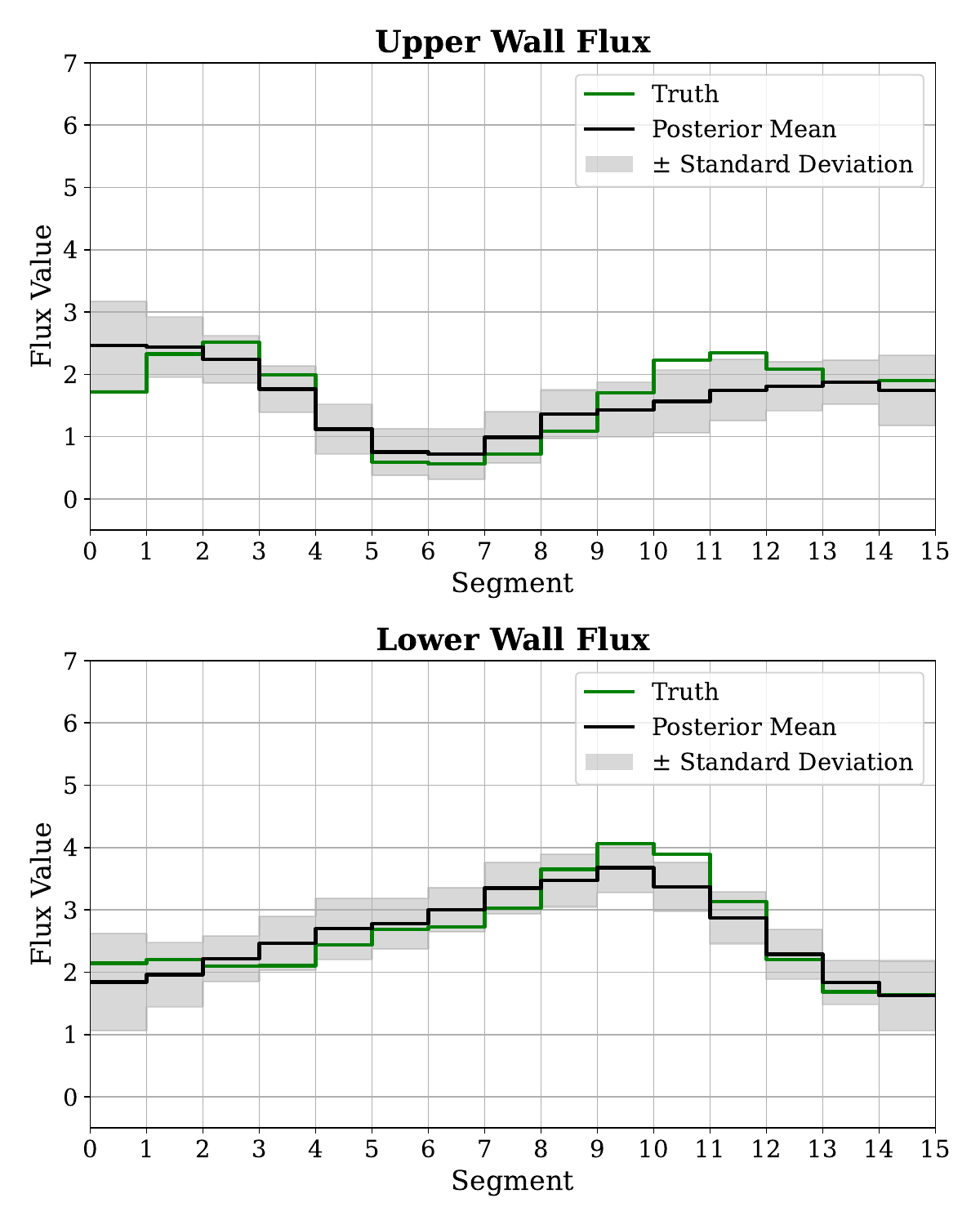}
    \caption{Posterior mean of boundary flux estimated using the variance exploding formulation with the ODE sampler for each segment in upper and lower walls for three different measurements in the test dataset (black line), true flux values (green line), one standard deviation range on either side of the generated posterior mean (gray shade). The corresponding concentration field and sensor locations are also shown for reference (first row) with \textcolor{red}{$\times$} indicating sensors that are OFF}
    \label{fig:mask_VE_mean_vs_true}
\end{figure}

In \Cref{fig:mask_VE_mean_vs_true}, we plot results for $\sigma_\epsilon = 0.02$, for three test cases. For each case, we plot the true flux distribution, the empirical mean generated by the conditional diffusion model (considered to be the best guess), and the one standard deviation range about the mean. Once again, for each case, we observe that the mean is close to the true value, and that in most cases the true value is contained within one standard deviation of the mean. 

\begin{figure}[H]
    \centering
    \includegraphics[width=\linewidth]{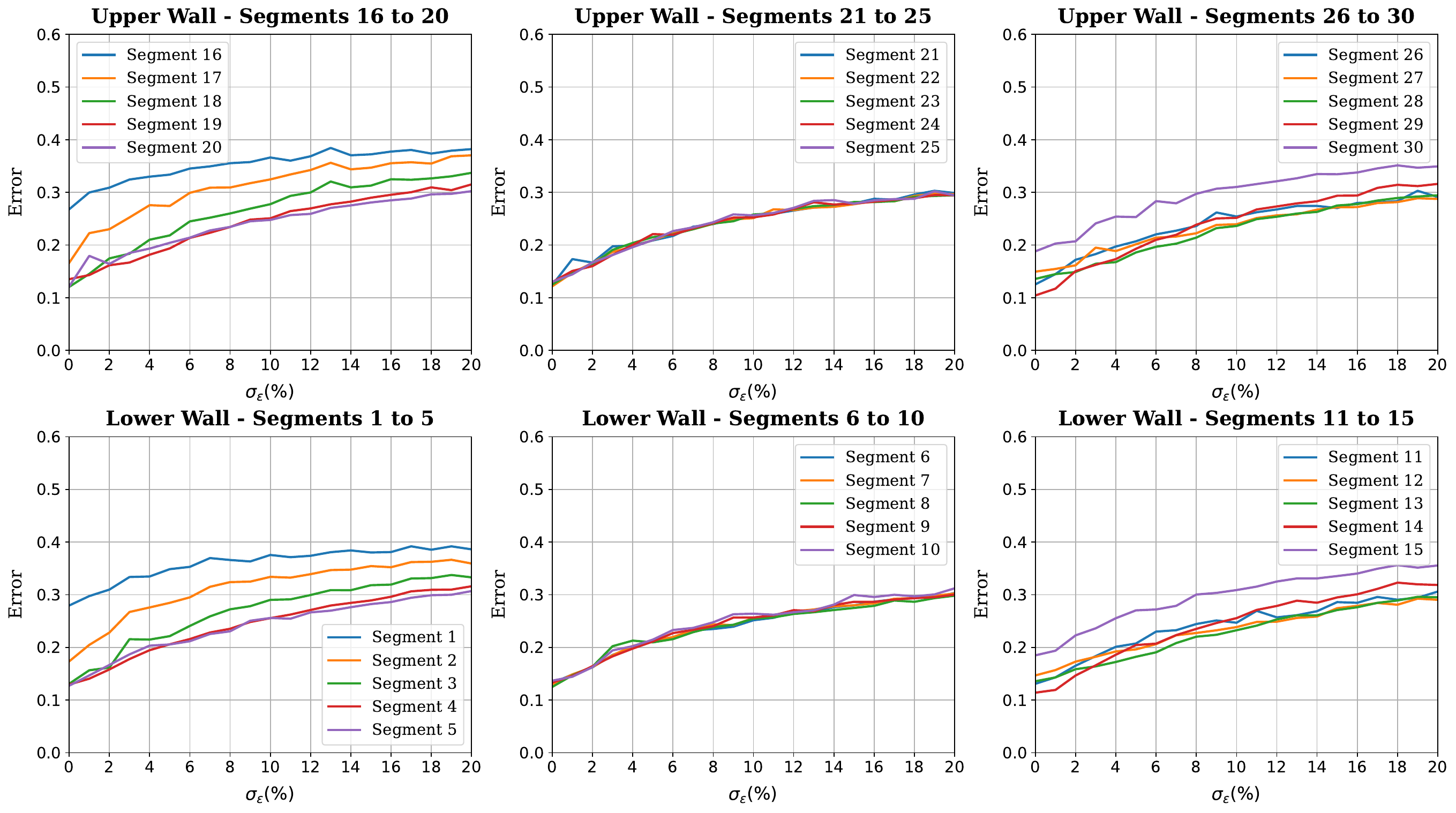}
    \caption{Sample-averaged absolute error of the generated posterior mean flux at each segment of the lower and the upper wall, normalized by the average value of flux, for all segments and all samples in the test set, for different levels of measurement noise when approximately 30\% of the sensors are OFF. These results were obtained using the variance exploding formulation and ODE sampler}
    \label{fig:mask_VE_error_vs_noise}
\end{figure}

In \Cref{fig:mask_VE_error_vs_noise}, we present a quantitative measure of the difference between the mean and the true value. For each flux segment, we plot the average absolute value of the difference between the predicted mean and the true flux, where the average is taken across all test samples. Once again, this value is normalized by the average value of flux for all segments and all samples in the test set. The trends in these plots are similar to those in \Cref{fig:VE_error_vs_noise}. However, in general, we observe that the error in this case is higher. This is to be expected; we are now working with 30\% fewer measurements on average.

\subsection{Estimating boundary flux in a nonlinear advection diffusion reaction problem}

In this section, we present results for an inverse problem governed by the nonlinear advection-diffusion-reaction equation. The domain, boundary conditions, and the measured and inferred quantities are the same as for the linear advection diffusion problem considered in \Cref{subsec:flux-problem1}. The key difference is that the concentration of the chemical species is now governed by the following nonlinear PDE,
\begin{eqnarray}
\label{eq:advection_diffusion_reaction}
\nabla \cdot (\bm{a} u) - \kappa \nabla^2 u  - u(r-u) = 0.
\end{eqnarray}
Here, $\bm{a}$ is a parabolic velocity field along the horizontal axis, with a maximum magnitude of 12 units, $\kappa = 8$ and $r = 2$. This yields a Peclet number of 6 that is close to the linear case. The reaction term models logistic growth dynamics and stabilizes the concentration near $r = 2$ units \cite{lam2020}. The boundary conditions, the parameters of the applied flux, and the measurement locations and parameters are unchanged from \Cref{subsec:flux-problem1}. 

\begin{figure}[htbp]
    \centering
    \includegraphics[width=0.3\textwidth]{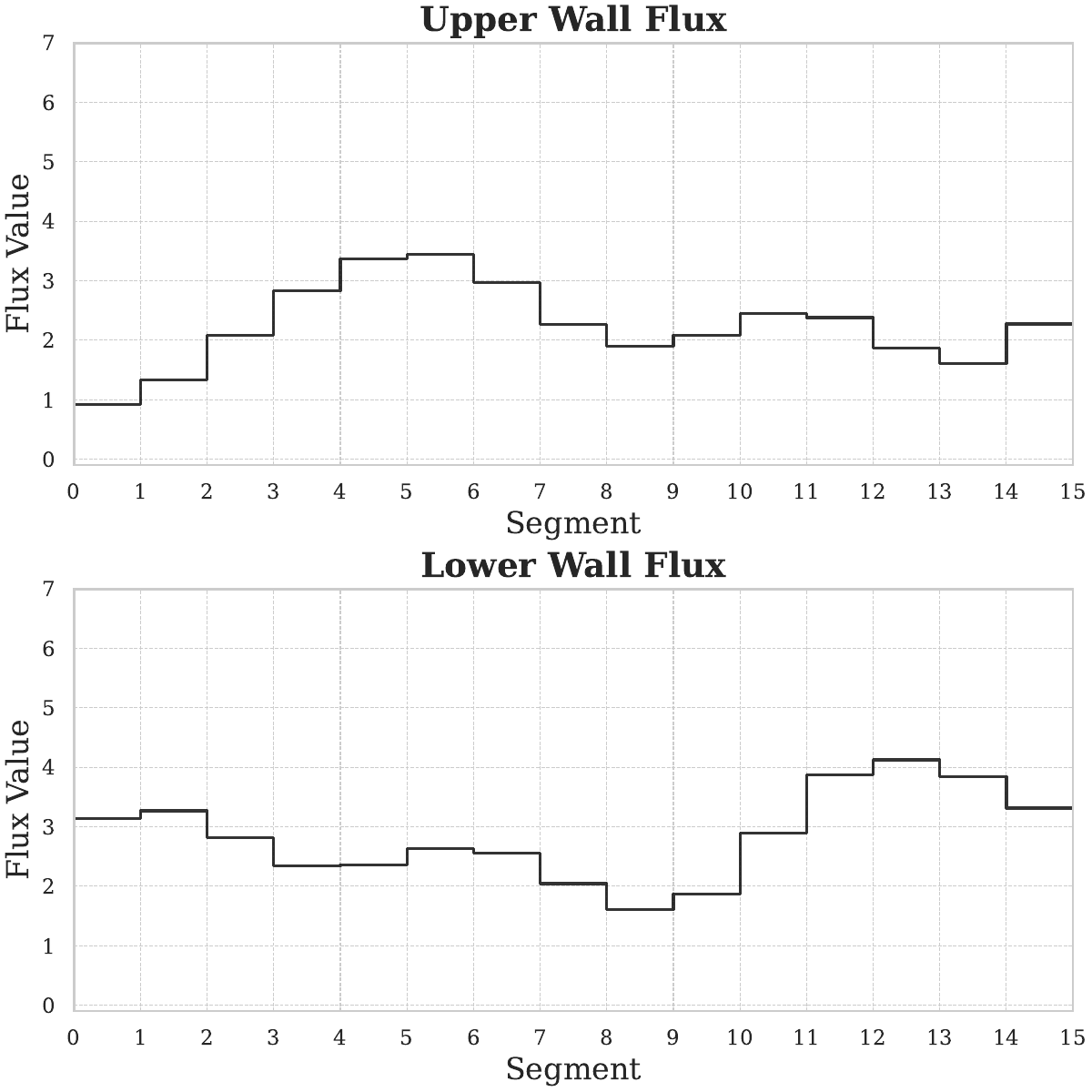}
    \hfill
    \includegraphics[width=0.3\textwidth]{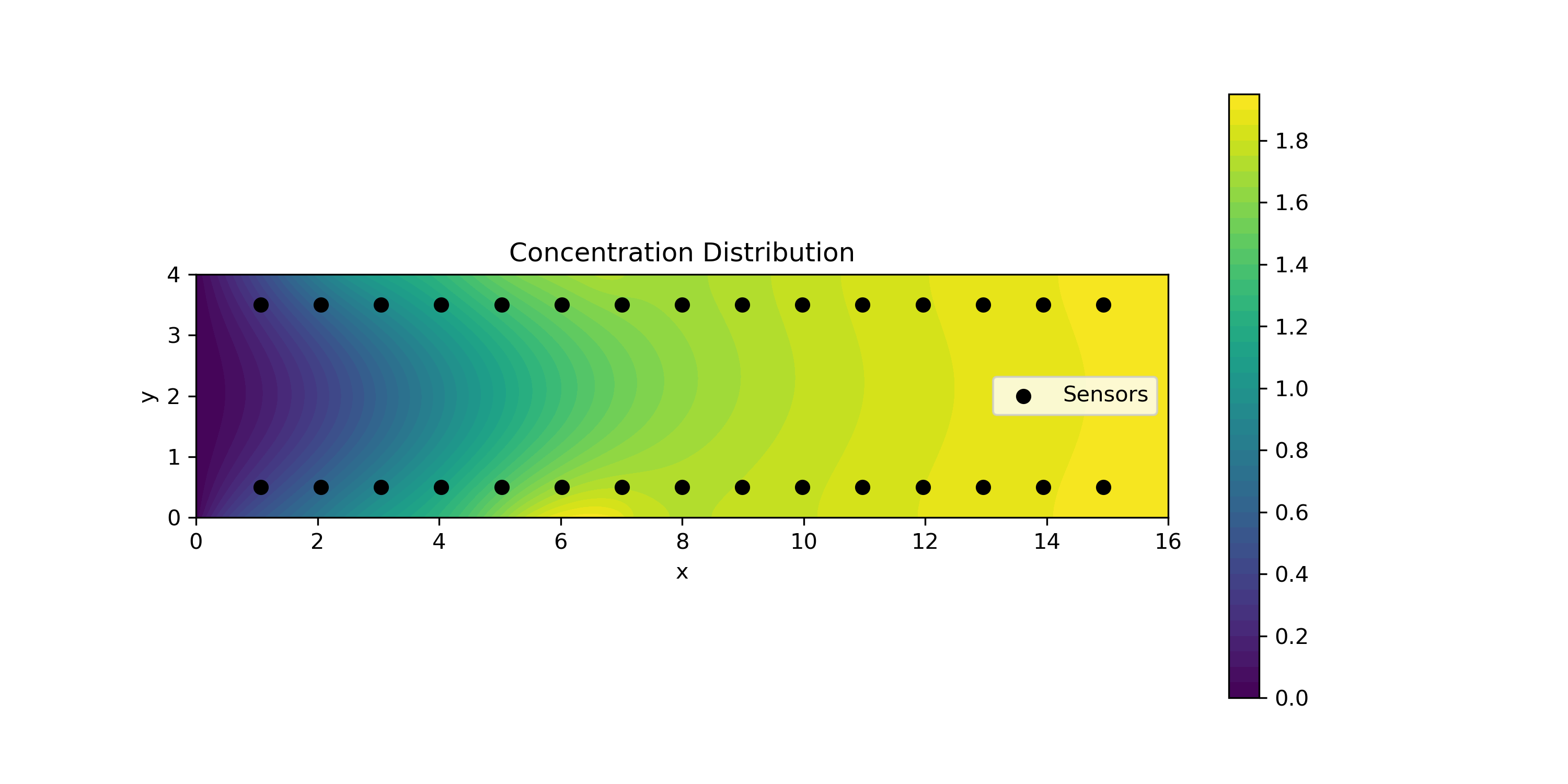}
    \hfill
    \includegraphics[width=0.3\textwidth]{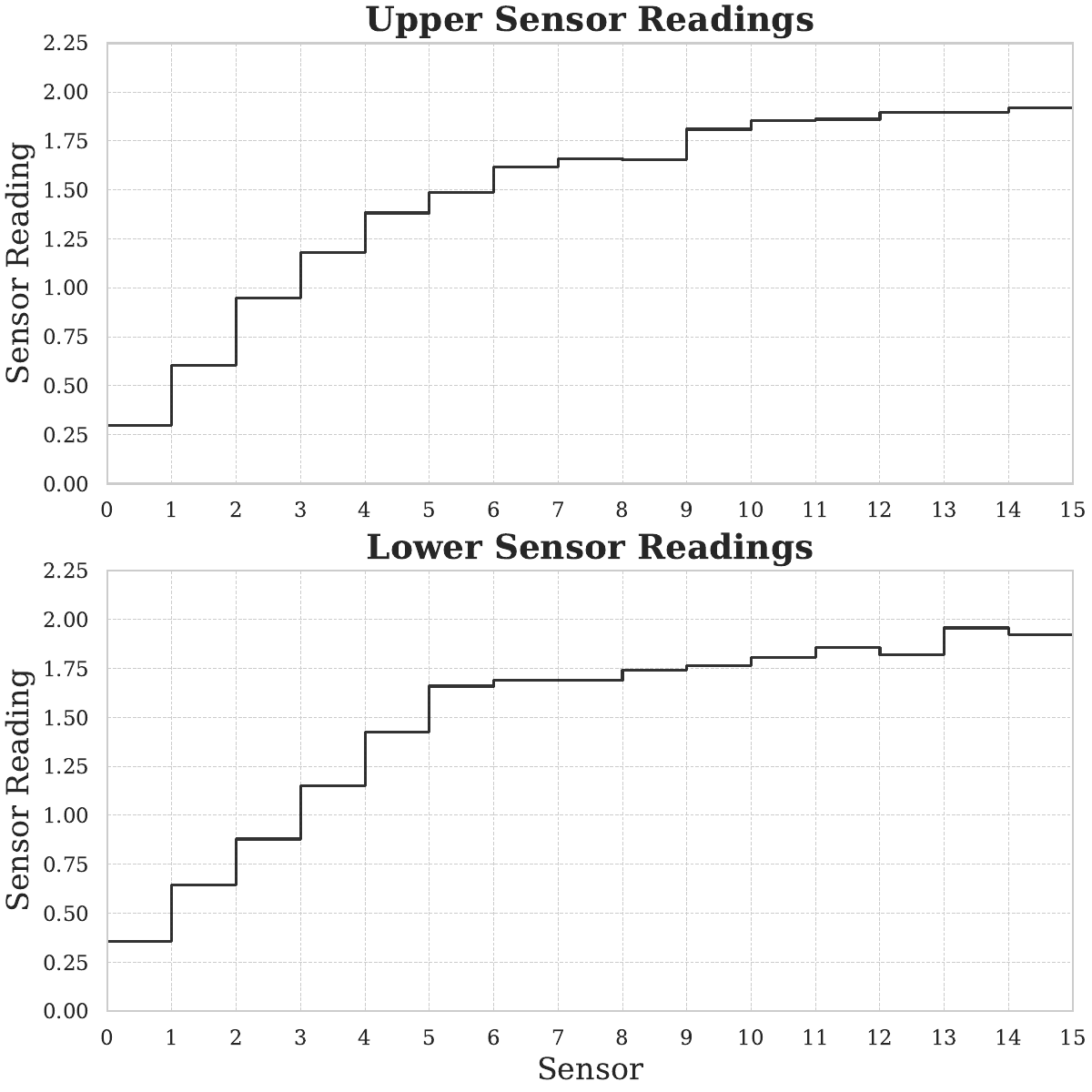}

    \vspace{0.5cm}

    \includegraphics[width=0.3\textwidth]{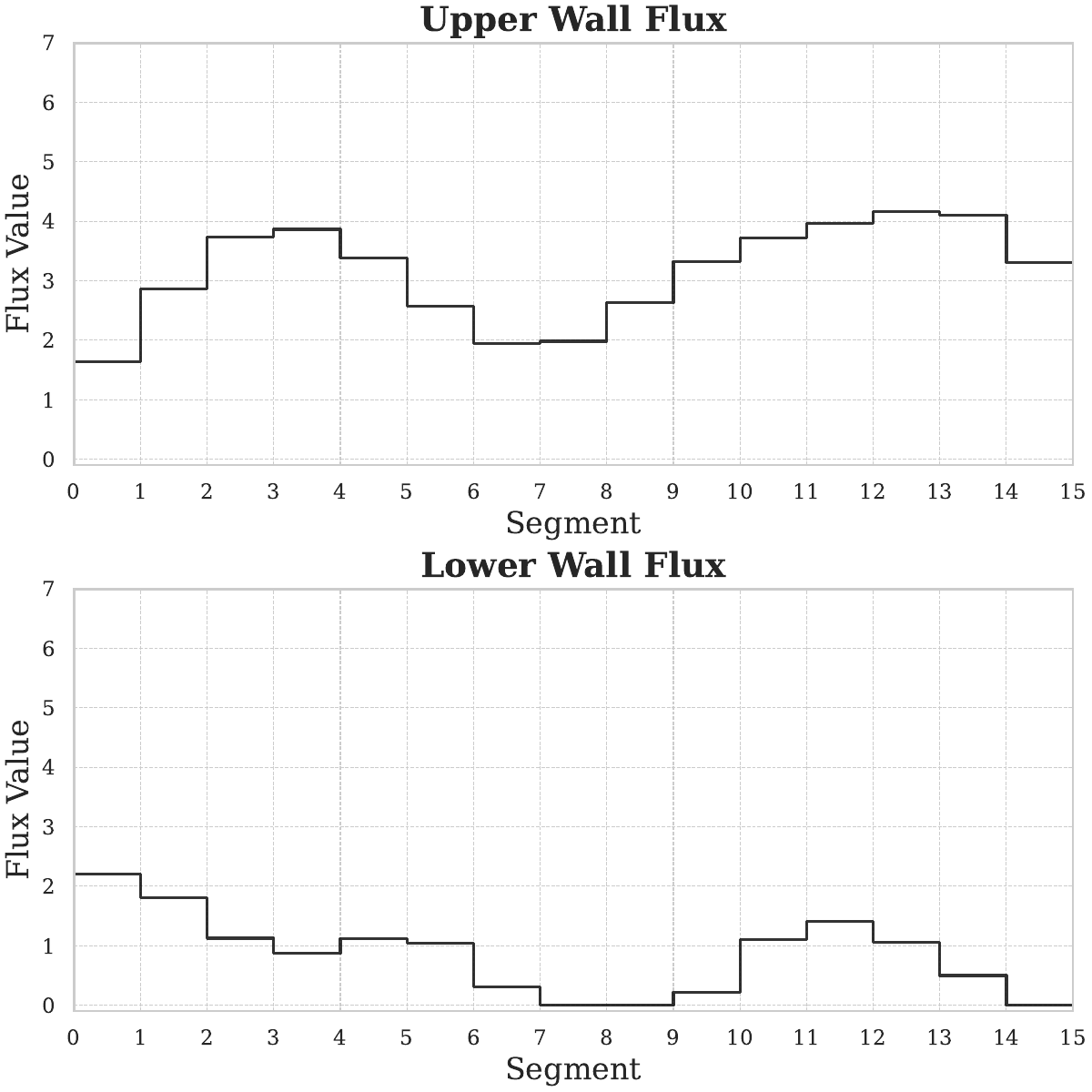}
    \hfill
    \includegraphics[width=0.3\textwidth]{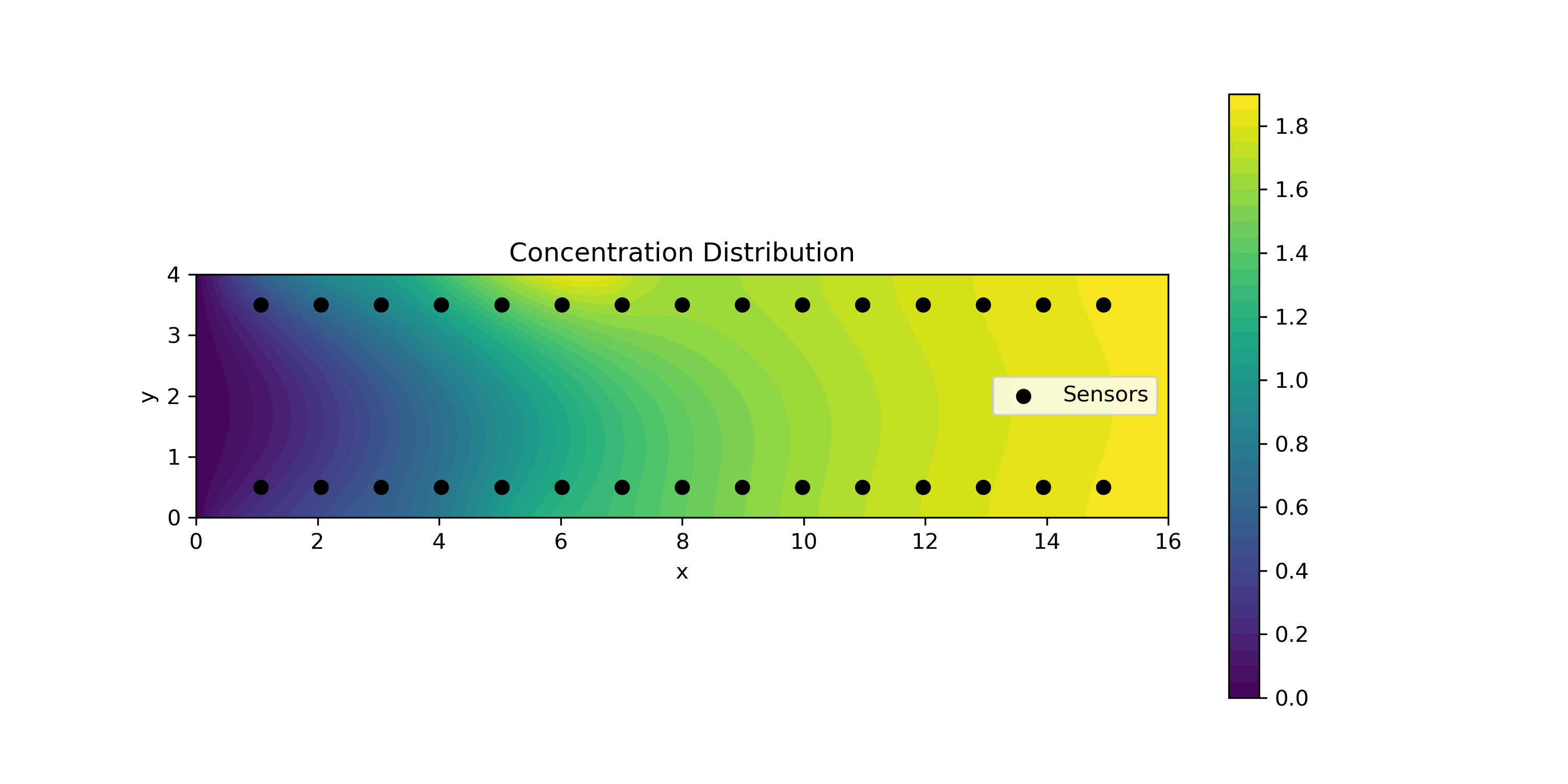}
    \hfill
    \includegraphics[width=0.3\textwidth]{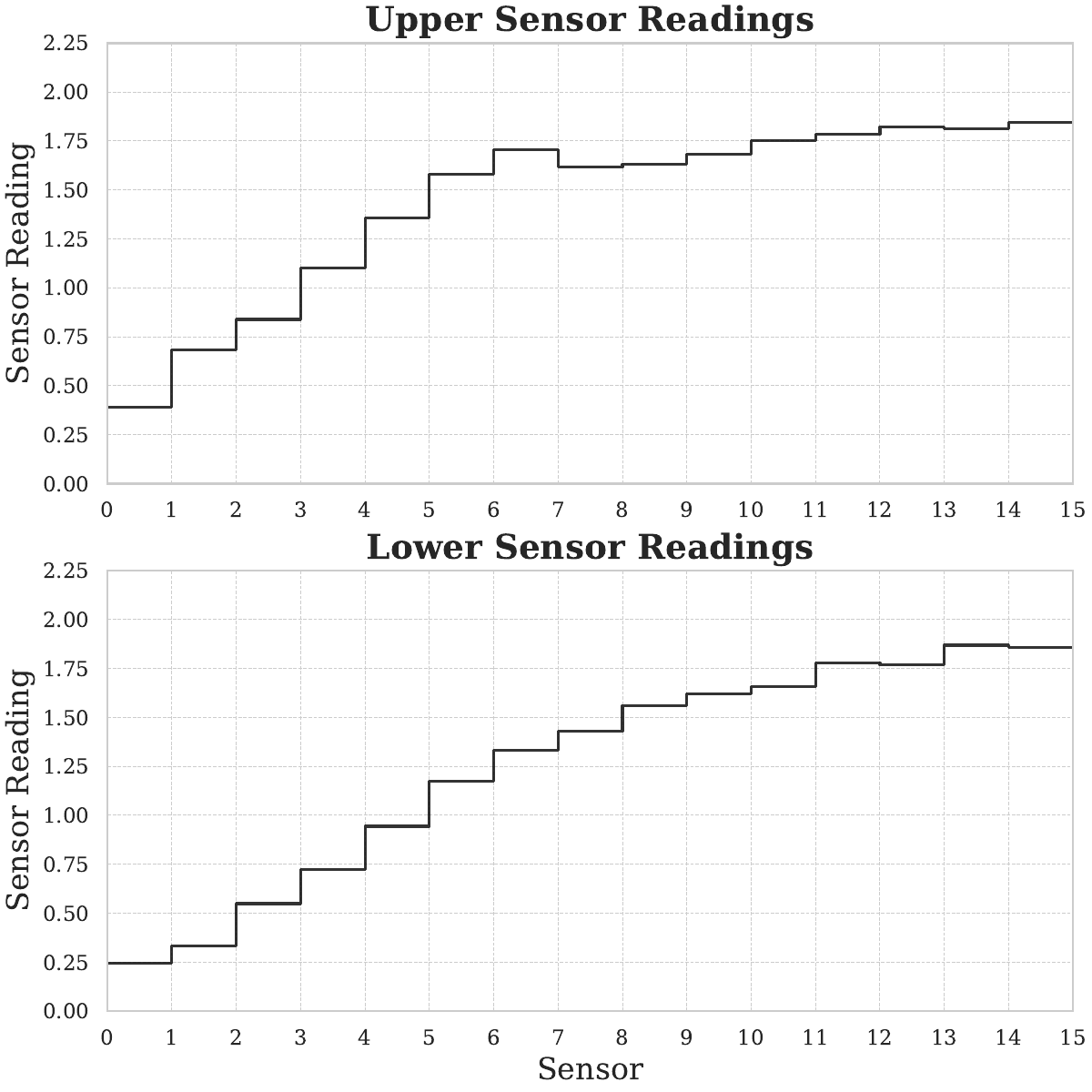}

    \vspace{0.5cm}

    \includegraphics[width=0.3\textwidth]{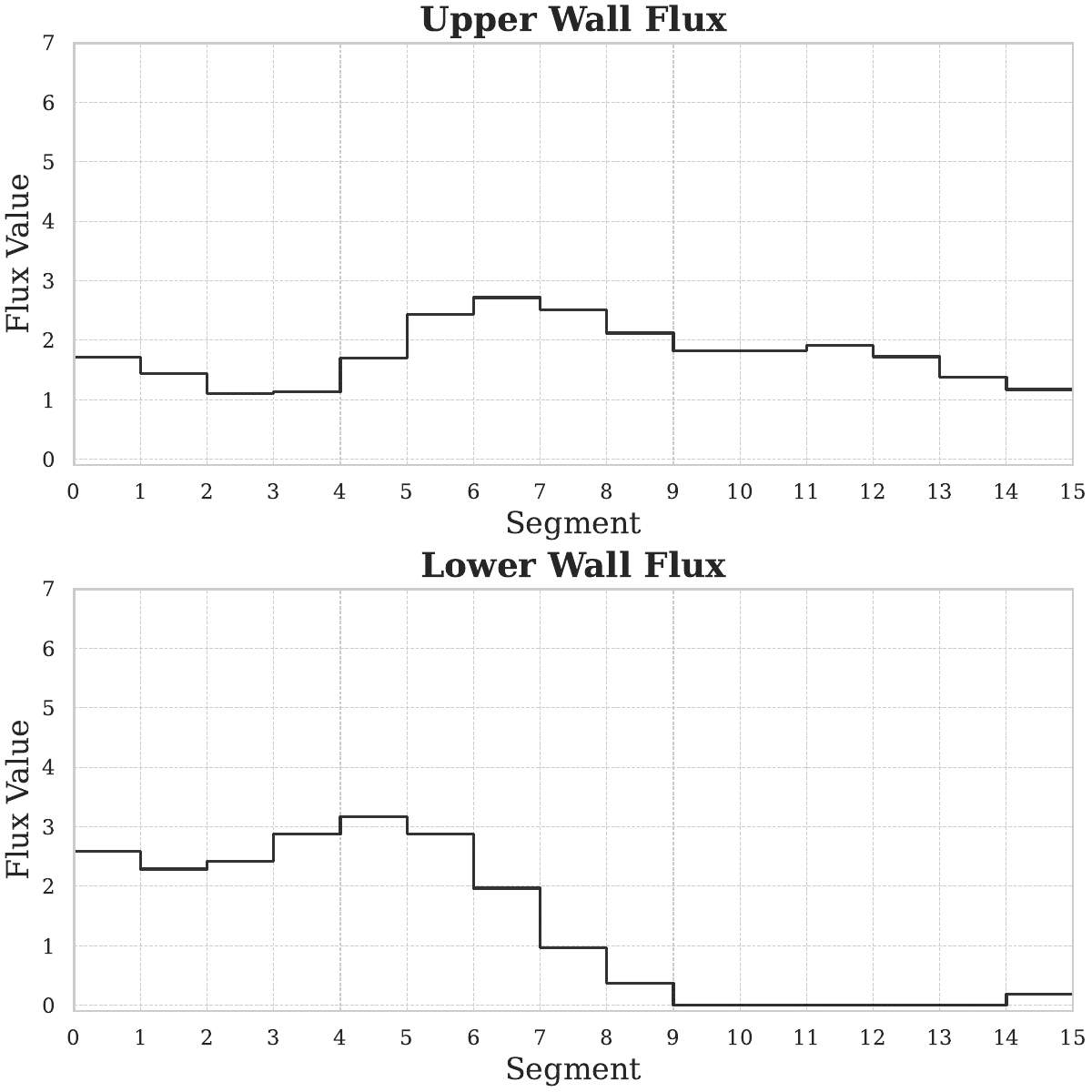}
    \hfill
    \includegraphics[width=0.3\textwidth]{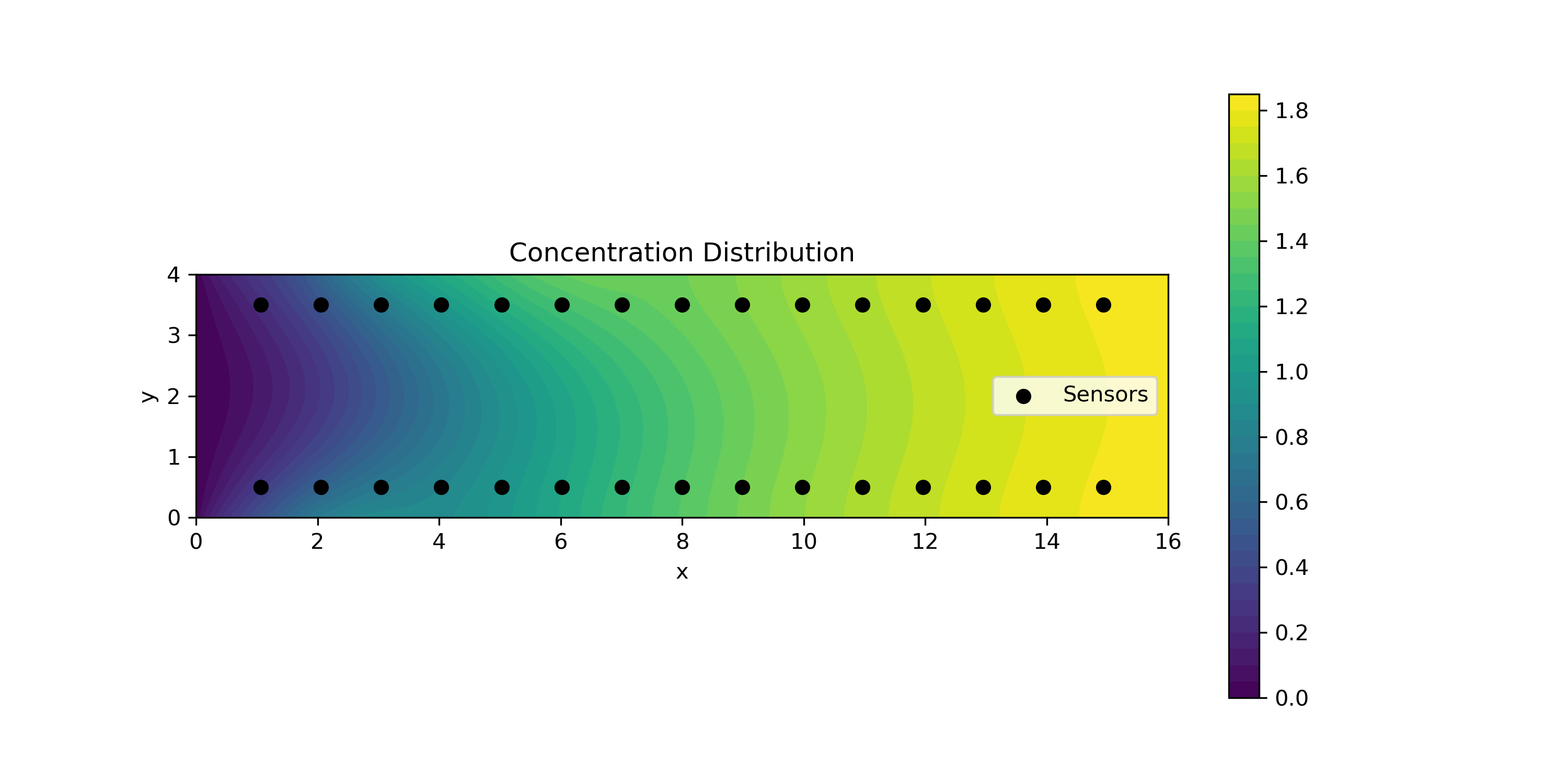}
    \hfill
    \includegraphics[width=0.3\textwidth]{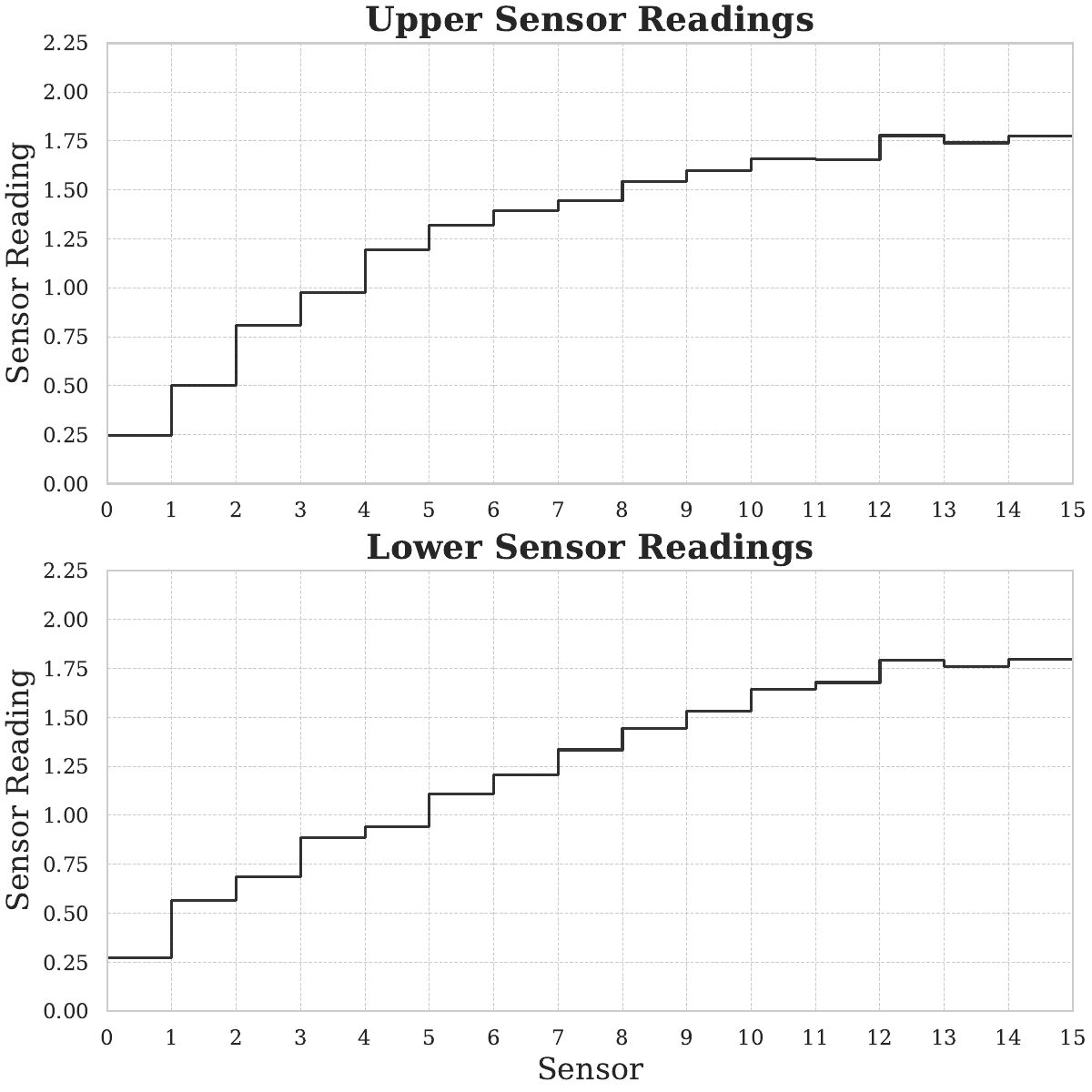}

    \caption{Three realizations of the top and bottom wall flux ($\X$) sampled from the prior (first column), corresponding concentration fields obtained after solving \Cref{eq:advection_diffusion_reaction}, and corresponding measurements ($\Y$) at various sensor locations (third column) for the nonlinear advection diffusion reaction problem}
    \label{fig:nonlinear_dataset_description}
\end{figure}

In  \Cref{fig:nonlinear_dataset_description} we plot three instances of the applied flux, the concentration in the domain, and the corresponding noisy measurement that were used as training data. The concentration was determined by solving \Cref{eq:advection_diffusion_reaction} with FEniCS on an 850$\times$200 grid, comprising 340,000 P1 elements. In these figures, we observe that the reaction term causes the concentration to tend towards a value of two as we move along the downstream direction.

The training dataset comprises 36,000 pairwise realizations of flux and concentration measurements, and the test dataset contains 4000 such samples. The nonlinearity in the problem requires a larger datset for inference using diffusion models. For each measurement from the test set, 1000 samples of the inferred flux (that belong to the posterior distribution) are generated using the trained conditional diffusion models. Sampling is performed using both the ODE and SDE-based algorithms. Since these do not differ significantly, only results from the ODE sampler are shown. 

\begin{figure}[h]
    \centering
    \includegraphics[width=0.3\textwidth]{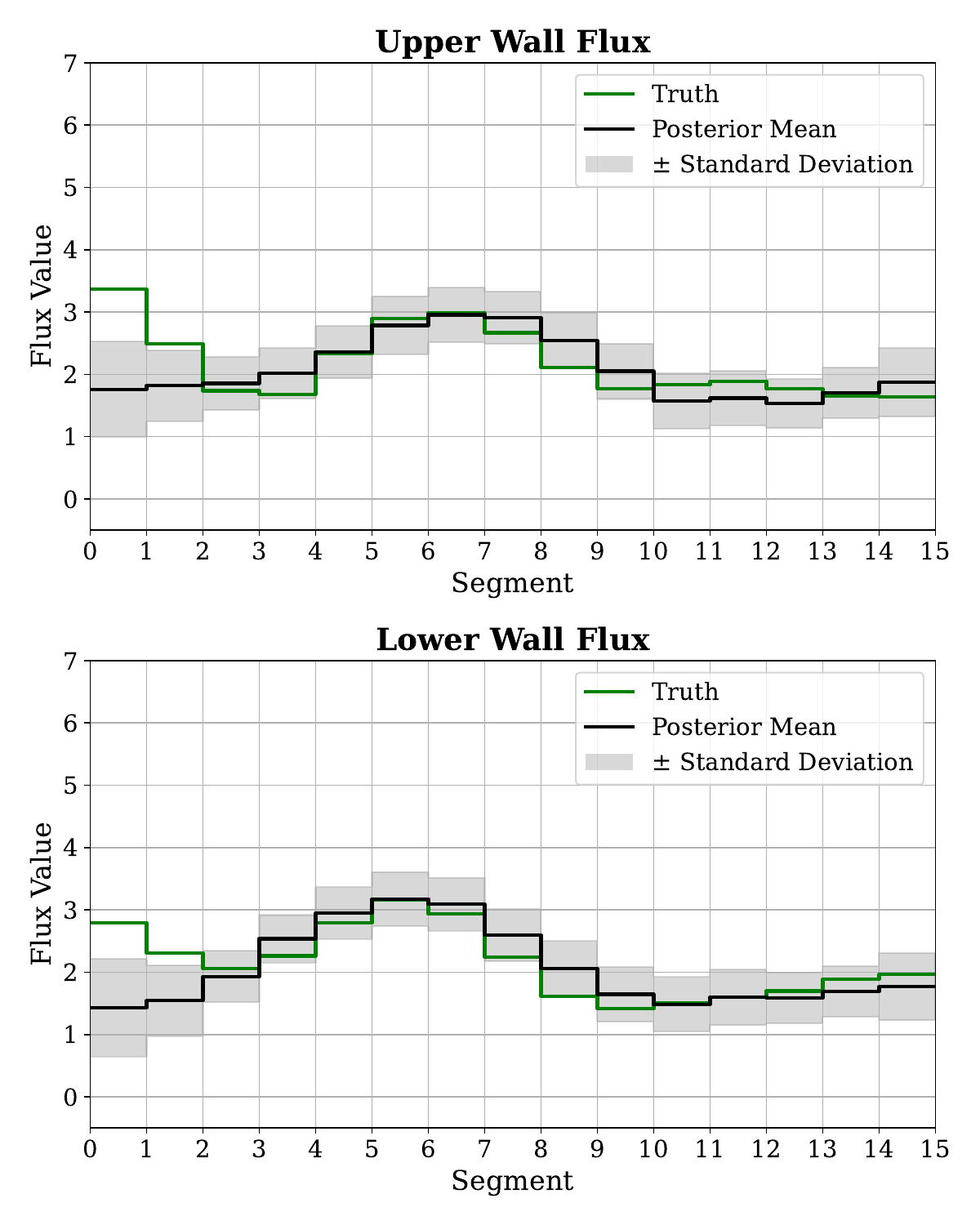}
    \hfill
    \includegraphics[width=0.3\textwidth]{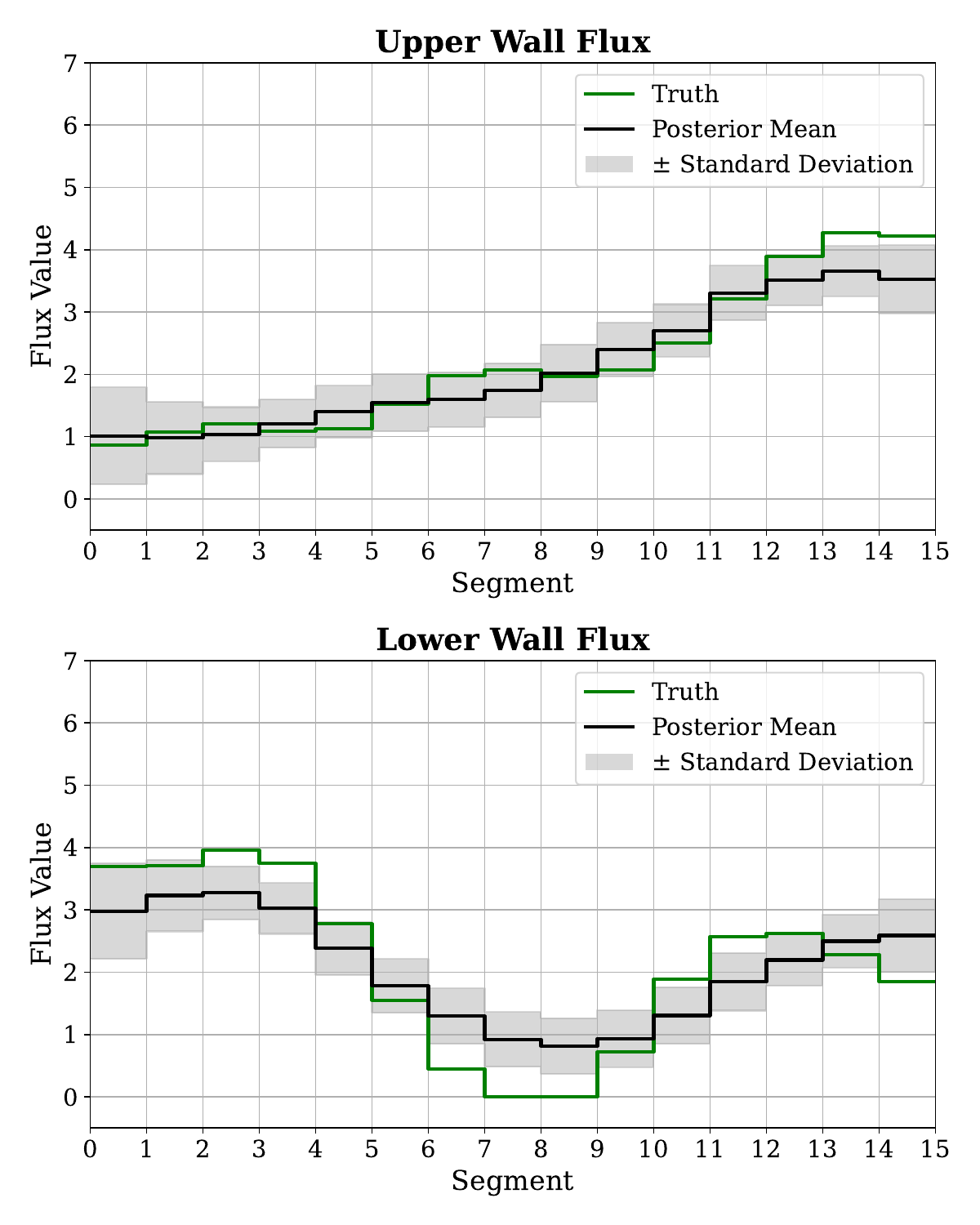}
    \hfill
    \includegraphics[width=0.3\textwidth]{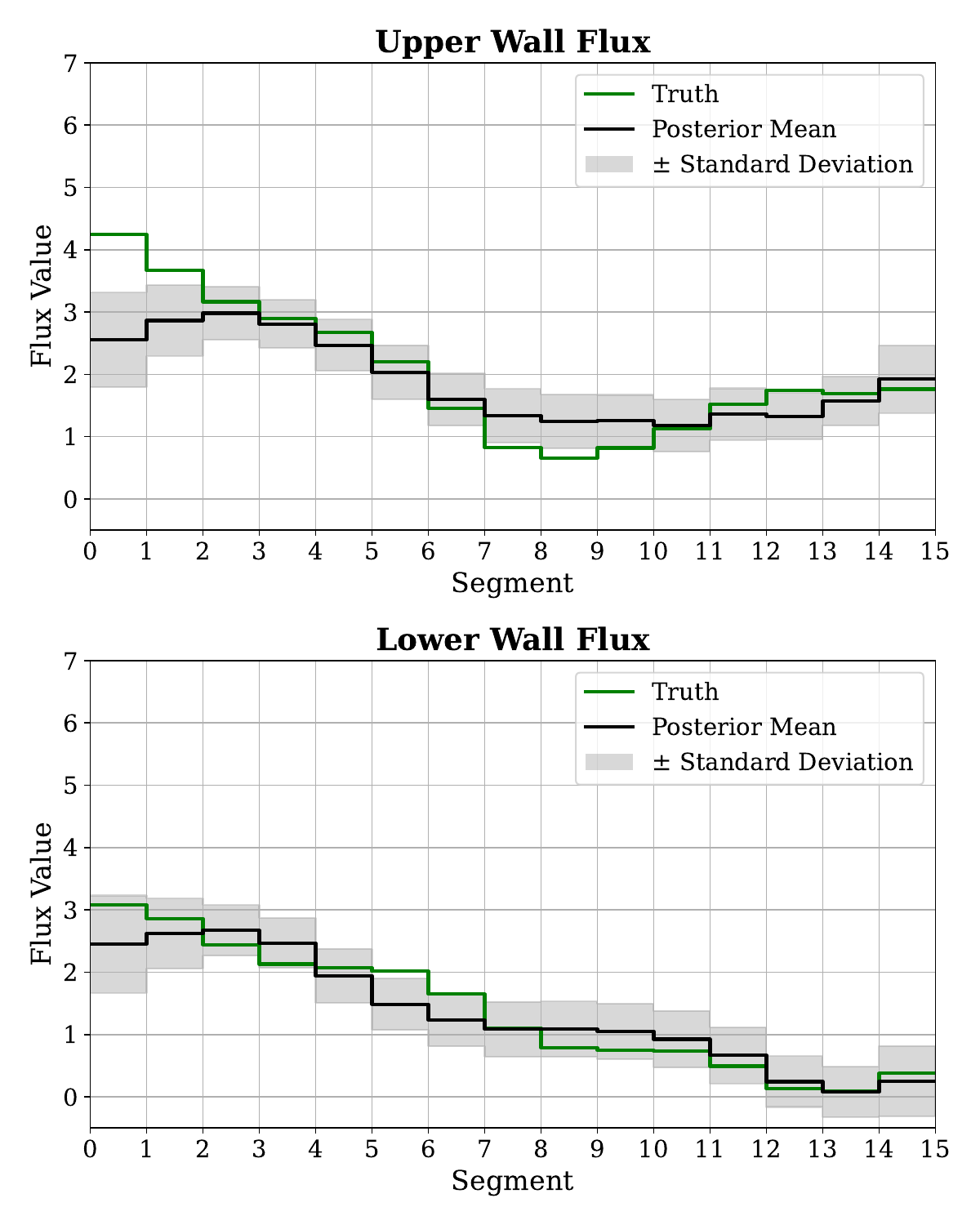}
    \caption{Posterior mean of boundary flux (black line) estimated using the variance exploding formulation with the ODE sampler for each segment in the upper and lower walls for three different measurements in the test dataset, corresponding true flux values (green line), and one standard deviation range around the posterior mean (gray shade). These results correspond to the nonlinear advection diffusion reaction problem}
    \label{fig:nonlinear_VE_mean_vs_true}
\end{figure}

In \Cref{fig:nonlinear_VE_mean_vs_true}, we present results for the variance exploding formulation with measurement noise $\sigma_{\epsilon} = 0.02$ for three test cases. For each case, we plot the true flux distribution, the empirical mean generated by the conditional diffusion model (considered to be the best guess), and the one standard deviation range around the mean. Similar to the linear problem, we observe that the mean is close to the true value, and that the true value typically lies within one standard deviation of the mean. 

\begin{figure}[h]
    \centering
    \includegraphics[width=\linewidth]{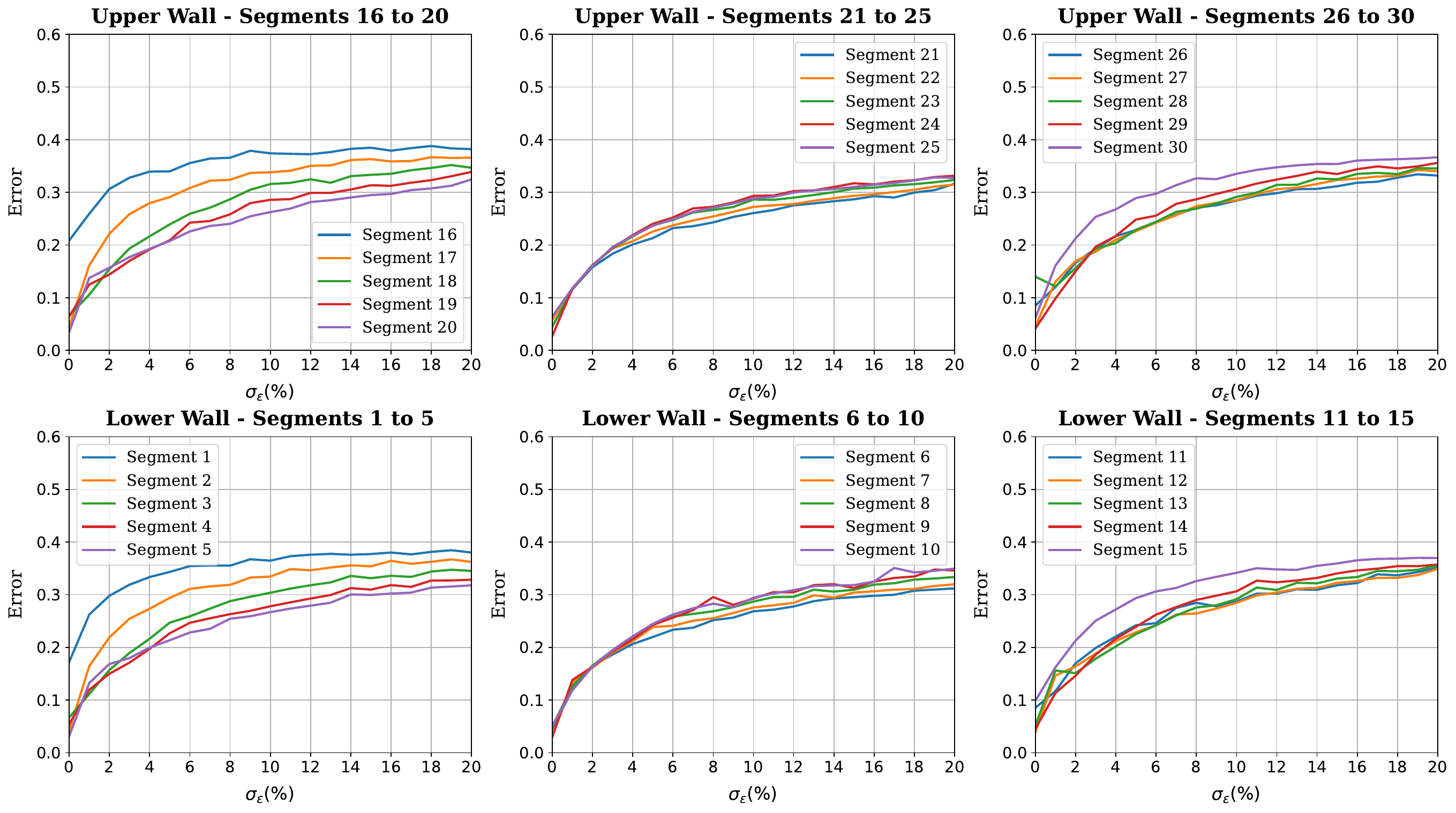}
    \caption{Sample-averaged absolute error of the generated posterior mean flux at each segment of the lower and the upper wall, normalized by the average value of flux, for all segments and all samples in the test set, for different levels of measurement noise in the nonlinear advection diffusion reaction problem. These results were obtained using the variance exploding formulation and ODE sampler and correspond to the nonlinear advection diffusion reaction problem}
    \label{fig:nonlinear_VE_error_vs_noise}
\end{figure}

A quantitative measure of the difference between the predicted mean and the true flux is shown in \Cref{fig:nonlinear_VE_error_vs_noise}. Here, for each segment, we plot the average absolute value of the difference between the mean predicted flux and the true flux across all test samples. Further, this value is normalized by the average value of flux for all segments and all samples in the test set. The trends observed in the linear problem in \Cref{fig:VE_error_vs_noise} are also observed in these plots. These include increasing error with increasing observation noise that saturates at higher values of noise, significant non-zero error even with zero measurement noise, and higher error for the first segment in both the upper and lower walls. 

\begin{figure}[h]
    \centering
    \includegraphics[width=\linewidth]{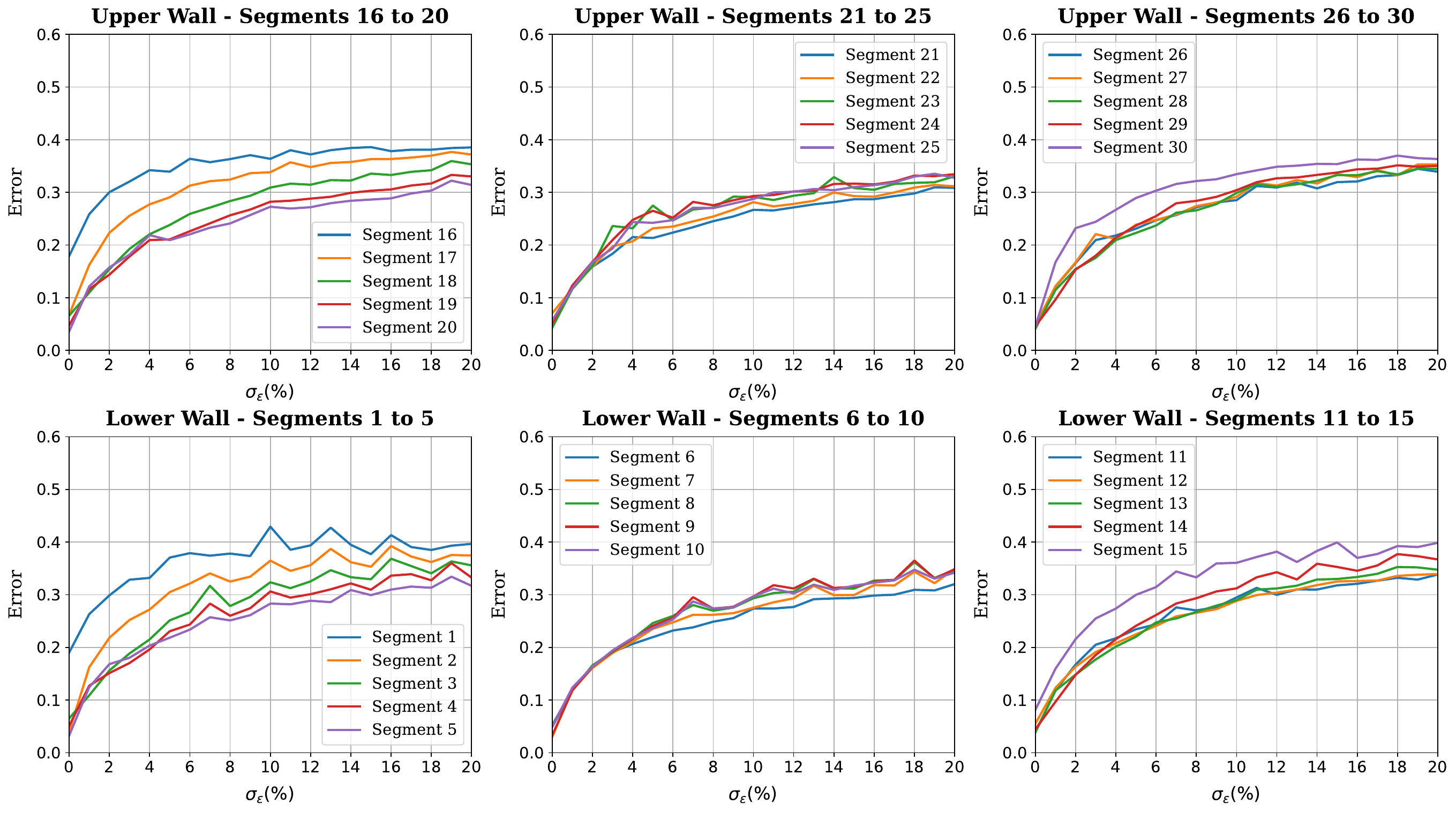}
    \caption{Sample-averaged absolute error of the generated posterior mean flux at each segment of the lower and the upper wall, normalized by the average value of flux, for all segments and all samples in the test set, for different levels of measurement noise in the nonlinear advection diffusion reaction problem. These results were obtained using the variance preserving formulation and ODE sampler}
    \label{fig:nonlinear_VP_error_vs_noise}
\end{figure}

In \Cref{fig:nonlinear_VP_error_vs_noise}, we plot similar results for the variance preserving formulation. By comparing \Cref{fig:nonlinear_VP_error_vs_noise} with \Cref{fig:nonlinear_VE_error_vs_noise}, we note that the two formulations incur similar errors.

\begin{figure}[h]
    \centering
    \includegraphics[width=\linewidth]{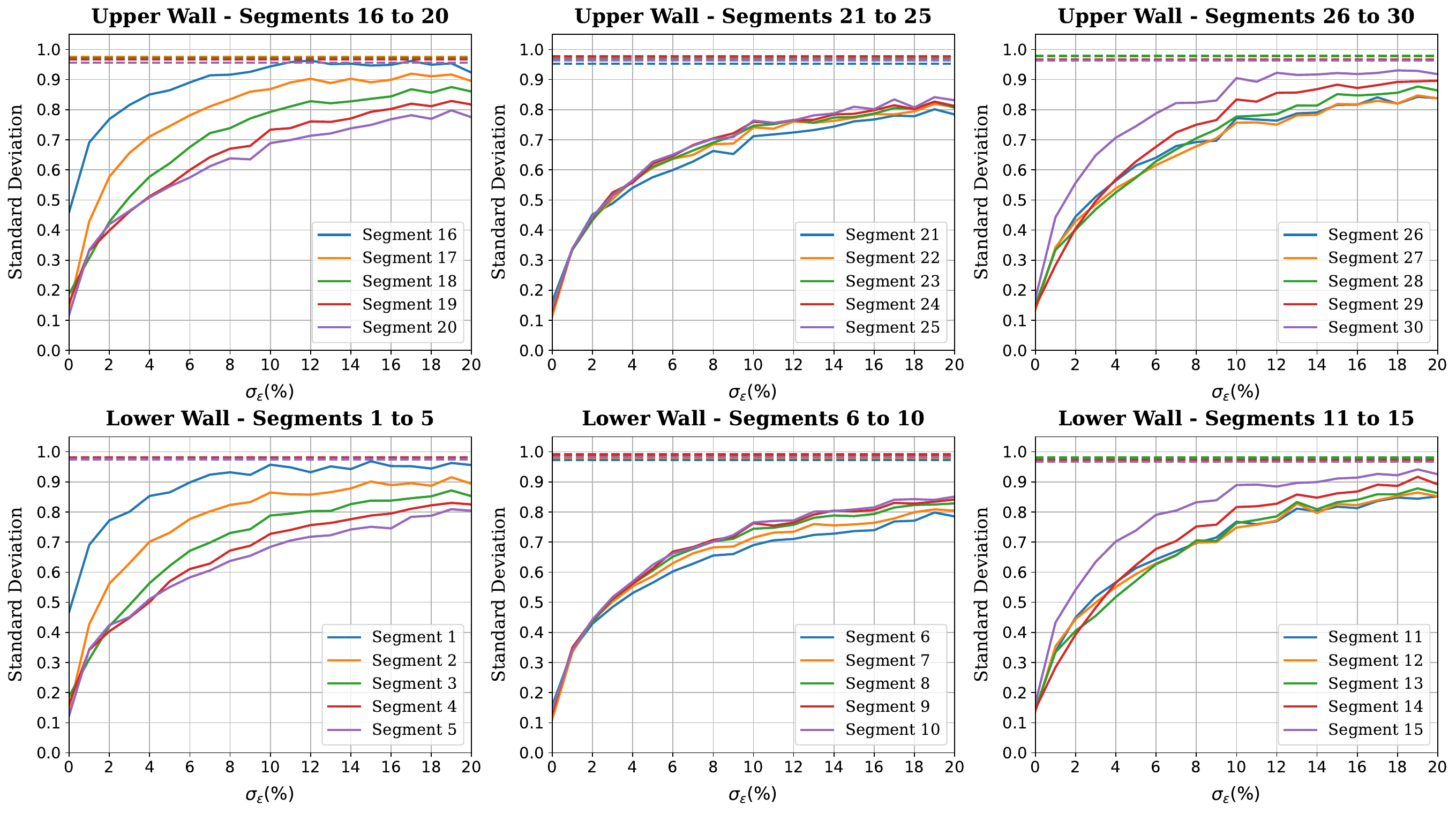}
    \caption{Sample-averaged posterior standard deviation of the flux for each segment of the lower and the upper wall for different levels of measurement noise obtained using the variance exploding formulation and ODE sampler in the nonlinear advection diffusion reaction problem}
    \label{fig:nonlinear_VE_std_vs_noise}
\end{figure}

\begin{figure}[h]
    \centering
    \includegraphics[width=\linewidth]{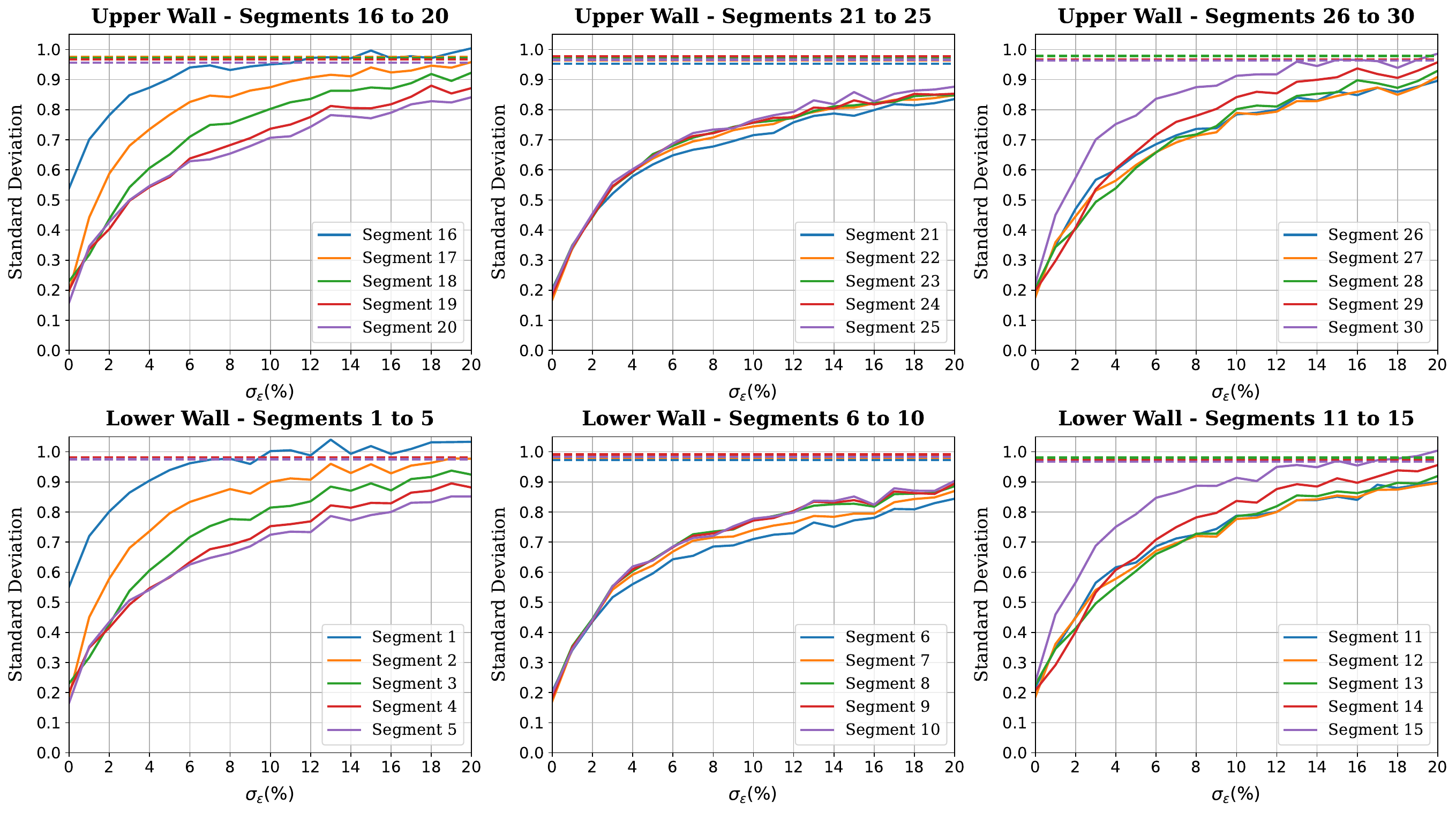}
    \caption{Sample-averaged posterior standard deviation of the flux for each segment of the lower and the upper wall for different levels of measurement noise obtained using the variance preserving formulation and ODE sampler in the nonlinear advection diffusion reaction problem}
    \label{fig:nonlinear_VP_std_vs_noise}
\end{figure}

In \Cref{fig:nonlinear_VE_std_vs_noise,fig:nonlinear_VP_std_vs_noise}, we plot the average standard deviation of the predicted flux obtained using the variance exploding and variance preserving formulations, respectively. The standard deviation is higher compared to the linear problem, particularly in the final segments along the horizontal axis (last column of the plots), resembling the behavior observed in the error plots. In these figures, the dashed horizontal lines indicate the standard deviation of the flux across all test samples, which approximates the standard deviation of the prior distribution for each segment. As the measurement noise increases, we observe that the posterior standard deviation approaches the standard deviation of the prior. This trend reinforces the notion that, at higher noise levels, the measurements become uninformative, and the posterior distribution reverts to the prior.


\section{Conclusions}\label{sec:conclusion}
In this paper, we have introduced several novel concepts in the development and application of diffusion models for solving probabilistic inverse problems. By adopting a pdf-centric view of the diffusion process, we provide a constructive derivation of the reverse process that maps a reference Gaussian distribution to a complex pdf. This includes deriving the existing variance exploding and variance preserving formulations as special cases, and also deriving a new family of variance preserving formulations. Further, it identifies a family of sampling algorithms that can be used to transform samples from the reference Gaussian distribution to the underlying data distribution of which the probability flow ODE and the stochastic ODE described in \cite{song2020score} are special cases. 

We also consider the application of diffusion models to conditional estimation and the solution of physics-driven probabilistic inverse problems. For several low-dimensional cases, we use diffusion models to generate samples from conditional distributions and quantify the error in the generated samples. In doing so, we establish a useful benchmark for these algorithms. Thereafter, we apply the conditional diffusion model to determine the boundary flux of a chemical species from sparse and noisy measurements of the interior concentration in an advection-diffusion and advection-diffusion-reaction problem. In the context of this problem, we describe how a single diffusion model can be trained to effectively solve the inverse problem associated with multiple observation operators. 

Taken together, the developments of this study provide a deeper understanding of the derivation of diffusion models and their application to physics-driven probabilistic inverse problems. 
	
\section{Acknowledgments}
The authors acknowledge support from ARO grant W911NF2410401. AMDC was supported in part by the Coordenacao de Aperfeicoamento de Pessoal de Nivel Superior – Brasil (CAPES) – Finance Code 001. The authors also acknowledge the Center for Advanced Research Computing (CARC, \href{https://carc.usc.edu}{carc.usc.edu}) at the University of Southern California for providing computing resources that have contributed to the research results reported within this publication.

\appendix
\setcounter{section}{0}
\renewcommand{\appendixname}{Appendix}
\renewcommand{\thesection}{Appendix~\Alph{section}}
\renewcommand{\thesubsection}{\Alph{section}\arabic{subsection}}
\renewcommand{\thefigure}{\Alph{section}\arabic{figure}}
\setcounter{figure}{0}
\renewcommand{\thetable}{\Alph{section}\arabic{table}}
\setcounter{table}{0}

\section{Family of variance preserving diffusion models}
\label{sec:var_pres}
Using the definition of $\xi(t)$ from \Cref{eq:xi_def} and of $\sigma(t)$ in \Cref{eq:sigma-t}, it is easy to show that 
\begin{equation}
\label{eq:temp1}
    \frac{\dot{\xi}(t)}{\xi(t)} = \frac{\gamma(t)}{2 \xi^2(t)} (1 -\xi^{-\mu}(t))^{\frac{2}{\mu} -1}.
\end{equation}
It is convenient to rewrite this formulation in terms of a user-defined function $\beta$ defined as 
\begin{equation}
\label{eq:defbeta}
    \beta(t) = 2 \frac{\dot{\xi}(t)}{\xi(t)}.
\end{equation}
This yields the solution 
\begin{equation}
\label{eq:defxi}
   \xi(t) = \exp \left(\frac{1}{2} \int_0^t \beta(s) \mathrm{d}s \right),  
\end{equation}
and given that $m(t) = \xi^{-1}(t)$,
\begin{equation} \label{eq:defm}
    m(t) =  \exp \left( - \frac{1}{2} \int_0^t \beta(s) \mathrm{d}s \right). 
\end{equation}
This along with $\hat{\sigma}(t) = m(t) \sigma(t)$ and \Cref{eq:xi_def}, yields
\begin{equation}
    \hat{\sigma}(t) = (1 - m(t)^{\mu} )^{\frac{1}{\mu}}.
\end{equation}
These values of $m(t)$ and $\hat{\sigma}(t)$ define the transformed kernel that appears in \Cref{eq:varpres-green}.

The next task is to rewrite the PDE satisfied by the kernel (\ref{eq:pde-var-pres01}) in terms of $\beta$ and $\mu$. For this we utilize \Cref{eq:temp1,eq:defbeta,eq:defxi} in \Cref{eq:pde-var-pres01} to arrive at
\begin{equation}
    \frac{\partial \probh{t}(\q| \x') }{\partial t} -\frac{\beta(t)}{2}  \nabla_{\q} \cdot(\q \probh{t}(\q| \x')) -\frac{\beta(t)}{2 } \left( 1 - \exp \left(- \frac{\mu}{2} \int_0^t \beta(s) \mathrm{d}s\right) \right)^{1 - \frac{2}{\mu}} \Delta_{\q}  \probh{t}(\q| \x')  = 0. 
\end{equation}

\section{Experimental settings}\label{appsubsec:models}

\paragraph{Score network} In all our experiments, we model the score network using a deep neural network, and encode the time-dependence  using the transformation:
\begin{equation}
    f(t) = [t - 0.5, \cos (2\pi t), \sin(2 \pi t), -\cos(4 \pi t)].
\end{equation}
The time embedding $f$ is concatenated with the spatial inputs before the first hidden layer. For the advection-diffusion problem studied in \Cref{subsec:flux-problem1}, the measurement operator, \ie the vector $\Mm$, is also concatenated with the other inputs. The missing measurements are set to $-1$. For the conditional density estimation problems in \Cref{subsec:cond-dens-toy-examples}, the score network consists of 2 hidden layers with \texttt{ReLU} activation. For the advection-diffusion problem in \Cref{subsec:flux-problem1}, the score network consists of 4 hidden layers with \texttt{ReLU} activation. We optimize the score networks using Adam with a constant learning rate of $\pwr{1}{-3}$ for 10,000 epochs and batch size equal to 1000. 

\paragraph{Variance Schedules} We set $T = 1$ in all experiments. For the variance exploding schedule, we adopt $\gamma(t) = \sigma_{\mathrm{max}}^{2t}$ such that $\sigma^2(t) = \left( \sigma_{\mathrm{max}}^{2t} - 1\right)  / \log \sigma^2_{\mathrm{max}}$. For the variance preserving schedule, we choose $\beta(t) = \beta_{\mathrm{min}} + t (\beta_{\mathrm{max}} - \beta_{\mathrm{min}})$ and $\mu = 2$. Both these schedules have been commonly adopted in the literature~\cite{baptista2025memorization,karras2022elucidating,song2020score}. 
\begin{table}[H]
    \centering
    \caption{Variance scheduling hyper-parameters for the numerical experiments in \Cref{sec:results}}
    \begin{tabular}{lccc}
        \toprule
        Hyper-parameter & \makecell{Conditional estimation \\ \Cref{subsec:cond-dens-toy-examples}} & \makecell{Advection-diffusion \\ \Cref{subsec:flux-problem1}}\\
        \midrule
        $\sigma_{\mathrm{max}}$ & 12 & 5\\
        $\beta_{\mathrm{min}}$ & 0.001 & 0.001\\
        $\beta_{\mathrm{max}}$ & 15 & 15\\
        \bottomrule
    \end{tabular}
    \label{tab:exp-settings}
\end{table}

\paragraph{Integrating the reverse process} We use an adaptive Explicit Runge-Kutta method of order 5(4), available through \texttt{SciPy}'s~\cite{2020SciPy-NMeth} \texttt{solve\textunderscore ivp} routine to integrate \Cref{eq:euler_maruyama-cond} when $\alpha = 0$. We integrate \Cref{eq:euler_maruyama-cond} using an explicit Euler-Maruyama method with $\Delta \tau = 0.002$~\cite{kloeden2012numerical} when $\alpha=1$.

\bibliographystyle{elsarticle-num-names} 
\bibliography{references}
	
\end{document}

%% file: macros.tex
\usepackage{lipsum}
\usepackage[labelsep=period,singlelinecheck=off,textformat=period]{caption}
\usepackage{subcaption}
\usepackage{fancyhdr} 
\usepackage{epsfig}
\usepackage{timesmt}
\usepackage{mathtools,nccmath}
\usepackage{mathrsfs}
\usepackage{wrapfig}
\usepackage[noabbrev]{cleveref}
\creflabelformat{equation}{#2\textup{#1}#3}

\usepackage{pdflscape}
\usepackage{afterpage}
\usepackage{float}
\usepackage{url}
\usepackage{enumitem}
\usepackage{accents}
\usepackage{cuted}
\usepackage{dirtytalk}
\usepackage{multirow}
\usepackage{multicol}
\usepackage{booktabs}
\usepackage{array}
\usepackage{makecell}
\usepackage{colortbl}
\usepackage{soul}
\newcolumntype{n}{>{\columncolor{blue!5}}c}

\usepackage{pbox}
\usepackage[ruled,commentsnumbered,linesnumbered,boxed]{algorithm2e}
\SetArgSty{textnormal}
\SetKwComment{Comment}{/* }{ */}

\usepackage[flushleft]{threeparttable}
\usepackage{algpseudocode}
\usepackage{setspace}
\usepackage{courier}
\usepackage{psfrag,pstool}

\newcommand{\comment}[1]{}

\Crefname{equation}{Eq.}{Eqs.}
\Crefname{figure}{Fig.}{Figs.}
\Crefname{tabular}{Tab.}{Tabs.}
\crefname{algocf}{alg.}{algs.}
\Crefname{algocf}{Algorithm}{Algorithms}
\Crefname{fct}{Fact}{Facts }

\usepackage{amsmath}
\usepackage{amssymb}
\usepackage{amsthm}
\usepackage{mathrsfs}

\usepackage{bm}
\usepackage{mathtools}
\usepackage{nicefrac}

\theoremstyle{plain}

\theoremstyle{definition}

\theoremstyle{remark}

\newcommand{\probb}[2]{\mathrm{p}_{#1}\!\left({#2}\right)}
\newcommand{\prob}[1]{\mathrm{p}_{#1}}
\newcommand{\probt}[1]{\tilde{\mathrm{p}}_{#1}}
\newcommand{\probh}[1]{\hat{\mathrm{p}}_{#1}}

\newcommand{\Mm}{{\mathbf M}}

\newcommand{\mm}{{\mathbf m}}

\newcommand{\q}{{\bm{q}}}

\newcommand{\w}{{\bm{w}}}

\newcommand{\x}{{\mathbf x }}

\newcommand{\X}{{\mathbf X}}

\newcommand{\y}{{\mathbf y}}
\newcommand{\Y}{{\mathbf Y}}

\newcommand{\z}{{\mathbf z}}

\newcommand{\thetaa}{\bm{\theta}}

\newcommand{\Nx}{{n_{\mathcal{X}}}}
\newcommand{\Ny}{{n_{\mathcal{Y}}}}

\newcommand{\ie}{i.e., }
\newcommand{\eg}{e.g., }
\newcommand{\supth}[1]{\ensuremath{{#1}^{\text{th}}}}

\newcommand{\pwr}[2]{\ensuremath{{#1} \times 10^{#2}}}

\usepackage{pgfplots}
\pgfplotsset{compat=1.16}
\usepackage{tikz}

\usetikzlibrary{shapes.misc}
\usetikzlibrary{arrows,arrows.meta,calc,backgrounds}

\tikzstyle{block} = [draw,rectangle,thick,minimum height=2em,minimum width=2em]
\tikzstyle{sum} = [draw,circle,inner sep=0mm,minimum size=2mm]
\tikzstyle{connector} = [->,thick]
\tikzstyle{line} = [thick]
\tikzstyle{branch} = [circle,inner sep=0pt,minimum size=1mm,fill=black,draw=black]
\tikzstyle{guide} = []
\tikzset{>=latex}

\makeatletter
\newcommand{\hathat}[1]{%
	\begingroup%
	\let\macc@kerna\z@%
	\let\macc@kernb\z@%
	\let\macc@nucleus\@empty%
	\hat{\raisebox{.35ex}{\vphantom{\ensuremath{#1}}}\smash{\hat{#1}}}%
	\endgroup%
}
\makeatother

\creflabelformat{equation}{#2(#1)#3}
\crefrangelabelformat{equation}{#3(#1)#4 to #5(#2)#6}

\labelcrefformat{subequation}{#2(#1)#3}
\labelcrefrangeformat{subequation}{#3(#1)#4 to #5(#2)#6}